\documentclass[sn-apa]{sn-jnl}

\usepackage{soul}
\usepackage[utf8]{inputenc}
\usepackage[small]{caption}
\usepackage{subcaption}
\usepackage{amsmath}
\usepackage{amsthm}
\usepackage{amssymb}
\usepackage{mathtools}
\usepackage{enumerate}
\usepackage{algpseudocode}

\jyear{2021}%

\theoremstyle{thmstyleone}%
\theoremstyle{thmstyletwo}%

\theoremstyle{thmstylethree}%

\newcommand{\dtw}{d}   
\algrenewcommand\algorithmicrequire{\textbf{Input:}}  
\algrenewcommand\algorithmicensure{\textbf{Output:}}  
\begin{document}

\title[Autonomous Agents and Multi-Agent Systems]{An Information-Theoretic Approach for Estimating Scenario Generalization in Crowd Motion Prediction}


\author*[1]{\fnm{Gang} \sur{Qiao}}\email{gq19@scarletmail.rutgers.edu}
\author[1]{\fnm{Kaidong} \sur{Hu}}\email{kaidong.hu@rutgers.edu}
\author[1]{\fnm{Seonghyeon} \sur{Moon}}\email{sm2062@cs.rutgers.edu}
\author[1]{\fnm{Samuel S.} \sur{Sohn}}\email{sss286@cs.rutgers.edu}
\author[2]{\fnm{Sejong} \sur{Yoon}}\email{yoons@tcnj.edu}
\author[1]{\fnm{Mubbasir} \sur{Kapadia}}\email{mk1353@cs.rutgers.edu}
\author[1]{\fnm{Vladimir} \sur{Pavlovic}}\email{vladimir@cs.rutgers.edu}

\affil*[1]{\orgdiv{Computer Science}, \orgname{Rutgers University}, \state{NJ}, \country{USA}}
\affil[2]{\orgdiv{Computer Science}, \orgname{The College of New Jersey},  \state{NJ}, \country{USA}}


\abstract{Learning-based approaches to modeling crowd motion have become increasingly successful but require training and evaluation on large datasets, coupled with complex model selection and parameter tuning. To circumvent this tremendously time-consuming process, we propose a novel scoring method, which characterizes generalization of models trained on source crowd scenarios and applied to target crowd scenarios using a training-free, model-agnostic Interaction + Diversity Quantification score, ISDQ.  The Interaction component aims to characterize the difficulty of scenario domains, while the diversity of a scenario domain is captured in the Diversity score.  Both scores can be computed in a computation tractable manner.  Our experimental results validate the efficacy of the proposed method on several simulated and real-world (source,target) generalization tasks, demonstrating its potential to select optimal domain pairs before training and testing a model.}  




\maketitle

\section{Introduction}\label{sec:Introduction}
Crowd movement studies the motion behavior of many agents in obstacle-configured environments. In decentralized crowd systems, we assume that an agent knows the configuration of obstacles, its own task, and its own policy. However, an agent does not know other agents' tasks and policies in the same scenario but can sense their movement. Additionally, we assume that the agents do not explicitly communicate with each other during their movement. As the setting is decentralized, a rational agent would independently plan and follow its shortest path, along which interactions with other agents may occur. In this work, a \emph{scenario} refers to the configuration of obstacles in an environment and all agents' tasks. An agent's \emph{task} refers to its initial position and destination, the start time when the agent is introduced into the environment, the maximal number of movement steps allowed, and the radius (size) of the agent.  

\begin{figure*}[h!]
 \centering
 \hspace{-0.2cm}\begin{subfigure}[b]{0.412\textwidth}
    \includegraphics[width=\textwidth]{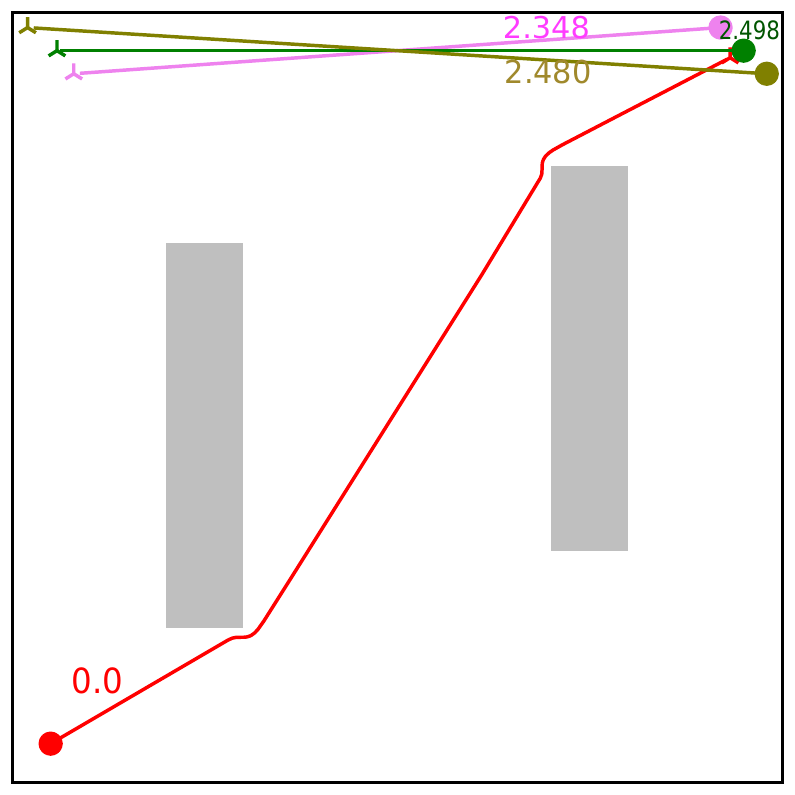}
    \caption{task level inter-agent interaction}
    \label{fig:task_level_inter_agent_interaction_a}
  \end{subfigure}
  \begin{subfigure}[b]{0.44\textwidth}
    \hspace{0.2cm}\vspace{0.08cm}\includegraphics[width=\textwidth]{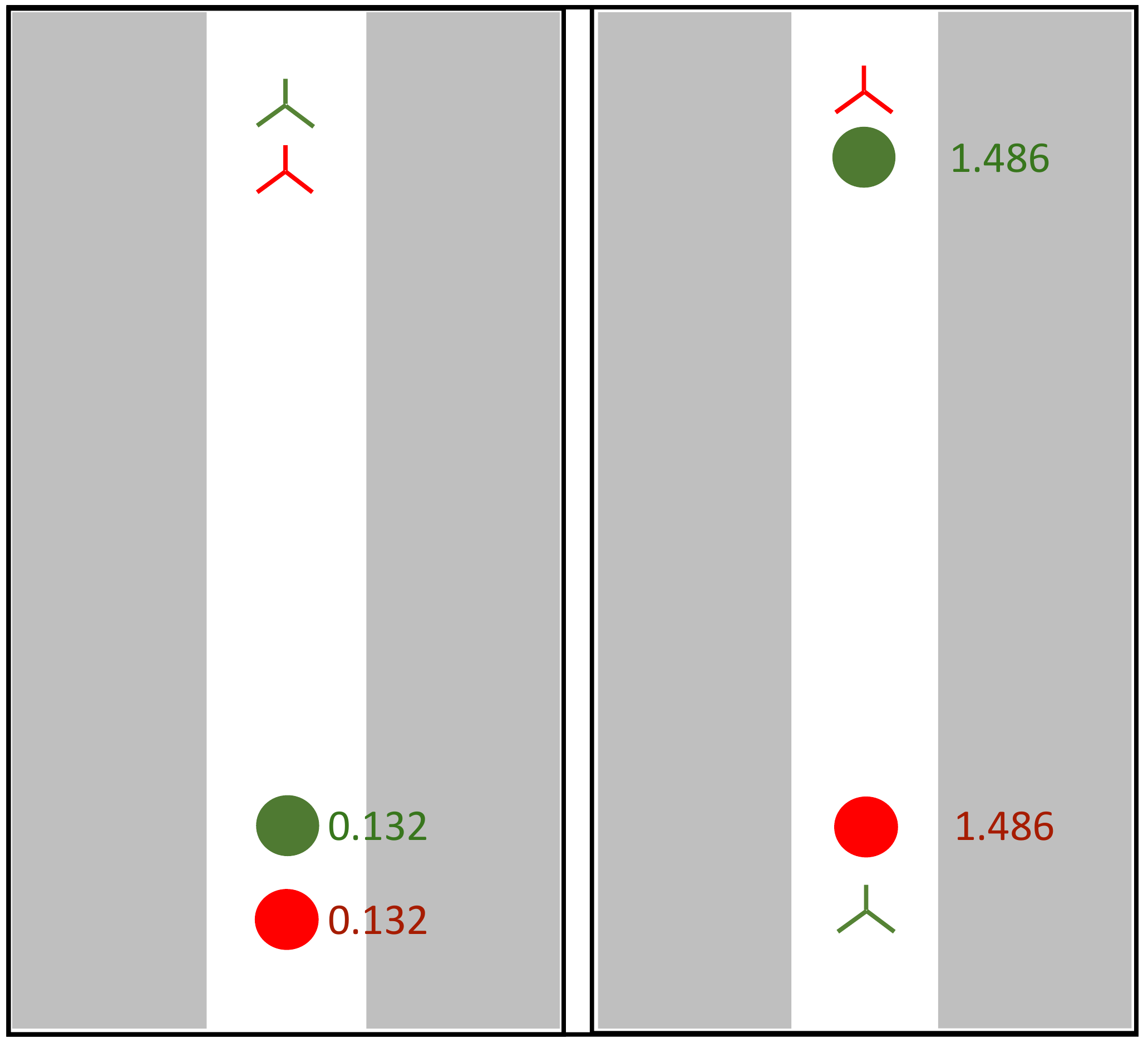}
    \caption{influence factor: navigation tasks}
    \label{fig:task_level_inter_agent_interaction_b}
  \end{subfigure}
  \begin{subfigure}[b]{0.70\textwidth}
    \includegraphics[width=\textwidth]{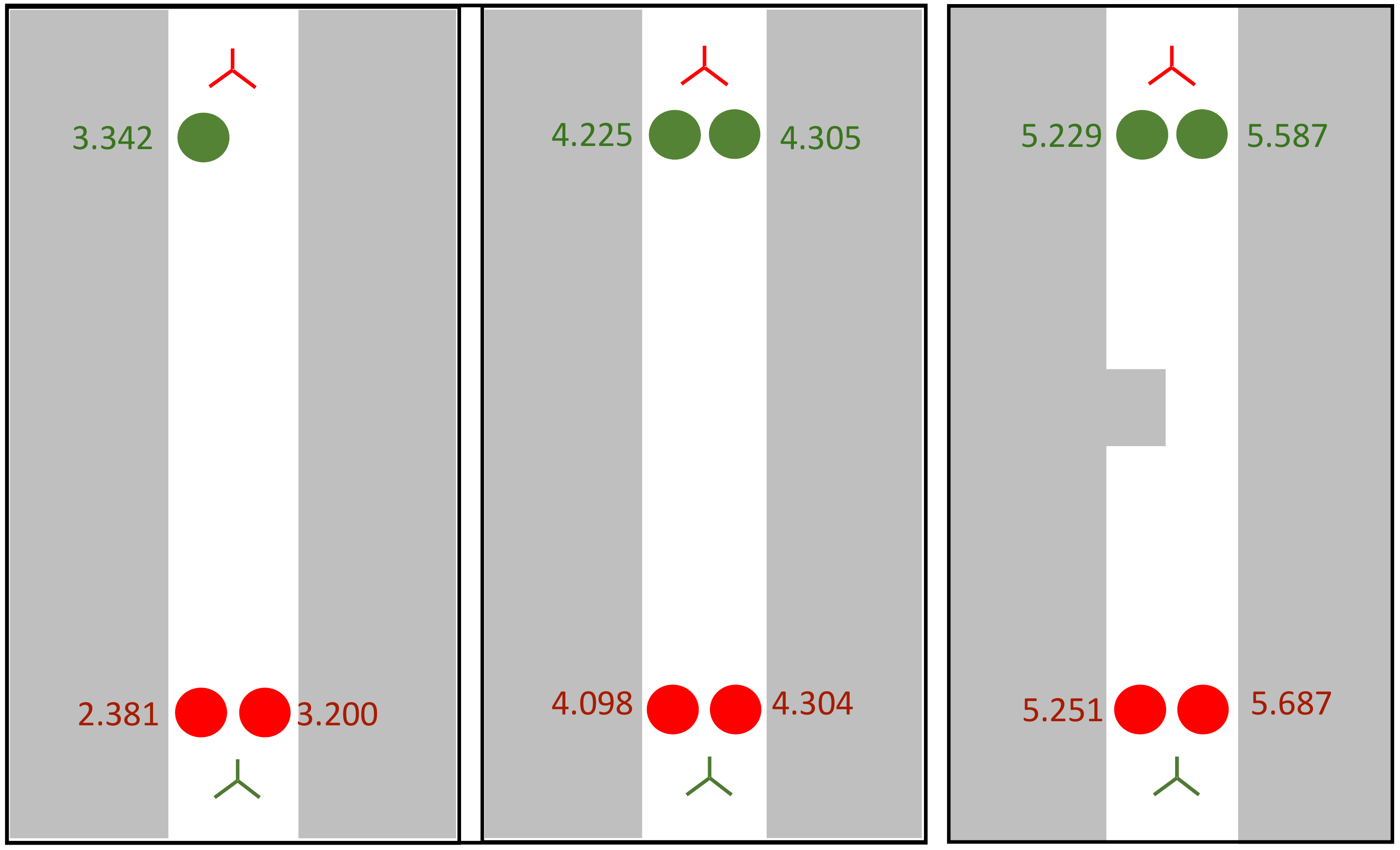}
    \caption{influence factors: task density and obstacle configuration}
    \label{fig:task_level_inter_agent_interaction_c}
   \end{subfigure}
 \caption{Illustration of the notion of task-level inter-agent interaction difficulty and its common influencing factors. The gray rectangles represent environmental obstacles. A filled circle and clover-like shape represents the initial position and destination of an agent of the same color. In \autoref{fig:task_level_inter_agent_interaction_a}, assuming the four agents enter the scenario and start their movement at the same time, intuitively, the red agent would have little interaction with the other three agents because when it moves to the top-right area of the scenario, probably, the other three agents would have already moved to the top-left area of the scenario. However, among the three other agents (pink, green, and brown), there are noticeable interactions among them in that they have to come across each other to reach their own destination. The interaction difficulty between an agent and all other agents, quantified by a value of the same color shown in the figure, being predictable without simulation, is at task level, inherent from the navigation task, no matter which steering model will be exploited. \autoref{fig:task_level_inter_agent_interaction_b} comparatively illustrates the influence of the assignment of tasks on task-level inter-agent interaction difficulty. Other common influencing factors such as the number of tasks (or task density) and obstacle configuration, and their superposition, are depicted in \autoref{fig:task_level_inter_agent_interaction_c}.}
\label{fig:task_level_inter_agent_interaction}
\end{figure*}

Under this setting, in a scenario, some tasks may involve more frequent agent ``encounters," potential inter-agent collisions, making them more difficult than other tasks, independent of the chosen steering policy/model. Such \textbf{task-level} \textbf{inter-agent interaction difficulty} reflects how much an agent \emph{has to} interact with other agents, contributing to the inherent difficulty of the task. The notion of task-level inter-agent interaction difficulty is important in this study, illustrated in \autoref{fig:task_level_inter_agent_interaction_a}. It is affected by multiple factors including the assignment of tasks, task density, obstacle configuration, and their superposition, shown in \autoref{fig:task_level_inter_agent_interaction_b} and \autoref{fig:task_level_inter_agent_interaction_c}. If accumulated or averaged over all tasks in a scenario, this measure can further characterize the interaction difficulty of the scenario or a collection of scenarios, which we refer to as the domain. This difficulty is inherently related to the complexity of solving many multi-agent problems, including multi-agent pathfinding and multi-agent trajectory prediction \cite{DBLP:conf/mig/SohnLMQ0YPK21}.  

\begin{figure*}[t!]
 \centering
    \begin{subfigure}[b]{0.80\textwidth}
        \includegraphics[width=\textwidth]{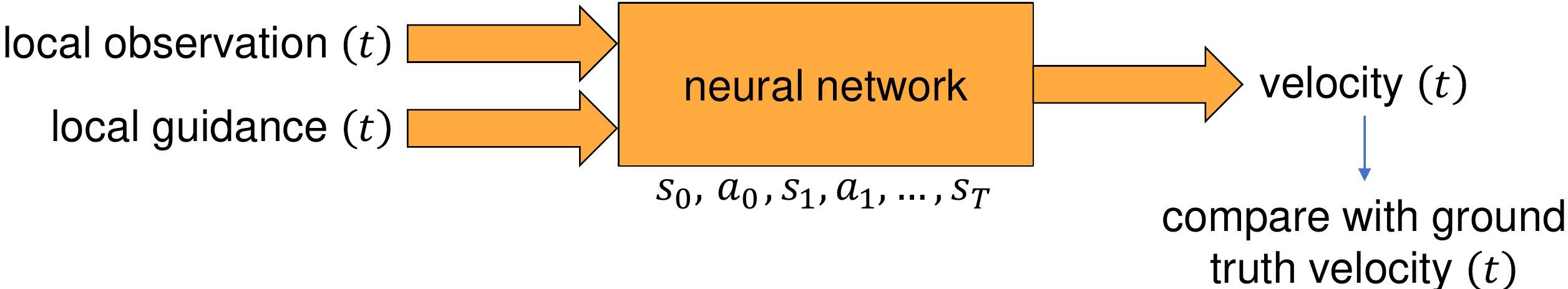}
        \caption{Standard generalization}
        \label{fig:RG_2_SG_a}
    \end{subfigure}
    \begin{subfigure}[b]{0.80\textwidth}
        \hspace{1cm}\includegraphics[width=\textwidth]{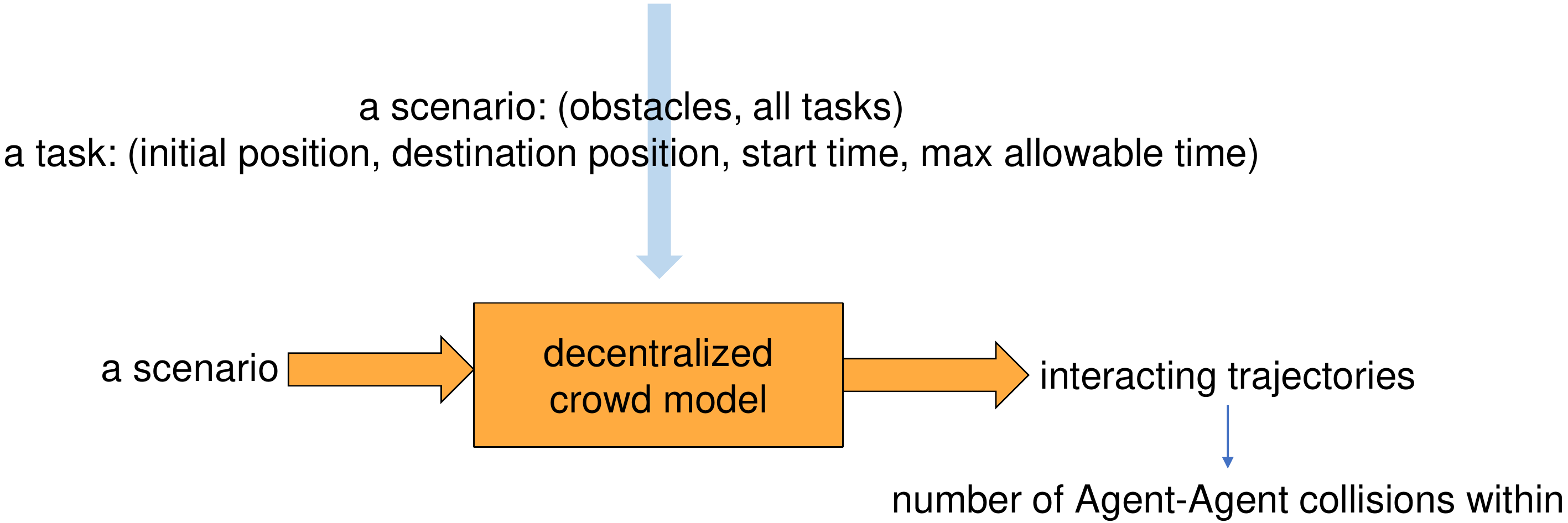}
        \caption{Scenario generalization}
        \label{fig:RG_2_SG_b}
    \end{subfigure}
\caption{The notion shift from standard generalization to scenario generalization. Standard generalization shown in \autoref{fig:RG_2_SG_a} is a microscopic view of crowd simulation for a concrete neural network model, which takes in the state (local observation and local guidance signal) at a time step and outputs an action (velocity) to be executed in the next step. This neural network model needs to be applied to each of the agents and applied repeatedly to yield the interacting trajectories of all agents. The error of standard generalization is computed by comparing the predicted output (velocity to be executed in the next step) with a ground truth output (expert velocity at the 
168
 same state). By contrast, with the notion of scenario, scenario generalization shown in \autoref{fig:RG_2_SG_b} is a macroscopic view of crowd simulation for an abstract decentralized crowd model, which takes in a scenario as an input sample and outputs the interacting trajectories of all agents in that scenario. The error of scenario generalization is computed within the predicted output itself, e.g., considering the number of agent--agent collisions.}
\label{fig:RG_to_SG}
\end{figure*}  

With the notion of ``scenario'', a crowd simulation could be viewed, macroscopically, as providing an abstract decentralized crowd model with a scenario as the input sample. The abstract decentralized crowd model outputs a set of interacting trajectories of all agents in that scenario. According to this view, a scenario is a sample drawn from either a source scenario domain or a target scenario domain. Therefore, a data-driven decentralized crowd modeling could be viewed as we first draw many scenarios from a source scenario domain to train an abstract decentralized crowd model and then apply the model to an unseen target scenario domain. Scenario generalization answers the question of whether the decentralized crowd model will seriously deteriorate if the target scenario domain significantly differs from the source scenario domain. An illustration of the shift of notion from standard generalization to scenario generalization is given in \autoref{fig:RG_to_SG}.  

In the context of crowds \cite{qiao2019scenario}, the scenario generalization of a crowd model to new scenarios is subtly but essentially different from the standard generalization of a model. For a standard generalization, the predicted output is compared with a ground truth output, while for scenario generalization, the error is measured within the output itself (the interacting trajectories), with metrics such as the number of agent--agent collisions and the number of agent--obstacle collisions. For this reason, scenario generalization surprisingly shares a similar formulation with the generalization of unsupervised machine learning models \cite{hansen1996unsupervised}. However, the objective of unsupervised learning is, in general, to recover the underlying true data distribution from data samples, while the crowd model aims to build a steering model that behaves well in an unseen target scenario domain. Furthermore, for a standard generalization, in order to calculate the error between the predicted output from a learned crowd model and the ground truth output from an expert model, the two models need to take the same input, which implies that at least one model can not make sequential decisions: either the learned crowed model or the expert model has to follow the trajectories of the other model, so that they have a common input basis. By contrast, scenario generalization only involves a crowd model that makes sequential decisions. 

Scenario generalization could be evaluated, empirically, by training a family of models on sufficiently many examples from the source domain using reinforcement learning (RL) methods or imitation learning methods, and then applying them to scenarios from the target domain to explicitly evaluate their performance. However, this can rapidly become a computationally intractable task, sensitive to many factors, such as the sizes of the source and target datasets, the configuration of input features, the model structures, and the learning strategies (e.g., optimization algorithms, hyperparameters, and possibly reward functions). we seek to circumvent this challenge by providing a pre-training characterization of (i) the impact of a source domain on the scenario generalization of a machine learning crowd model and (ii) the inherent task-level inter-agent interaction difficulty of a target domain. This approach would make it possible to not only assess the scenario generalization for a fixed pair of source-target domains, avoiding the need for collecting large crowd-movement data (by simulation or real data collection) and training complex models, but also enable the rapid elucidation of the best source-target domain pairs among multiple possible choices \emph{prior to} the training and testing of the learning model. 

\begin{figure}[t]
  \centering
  \includegraphics[width=0.95\linewidth]{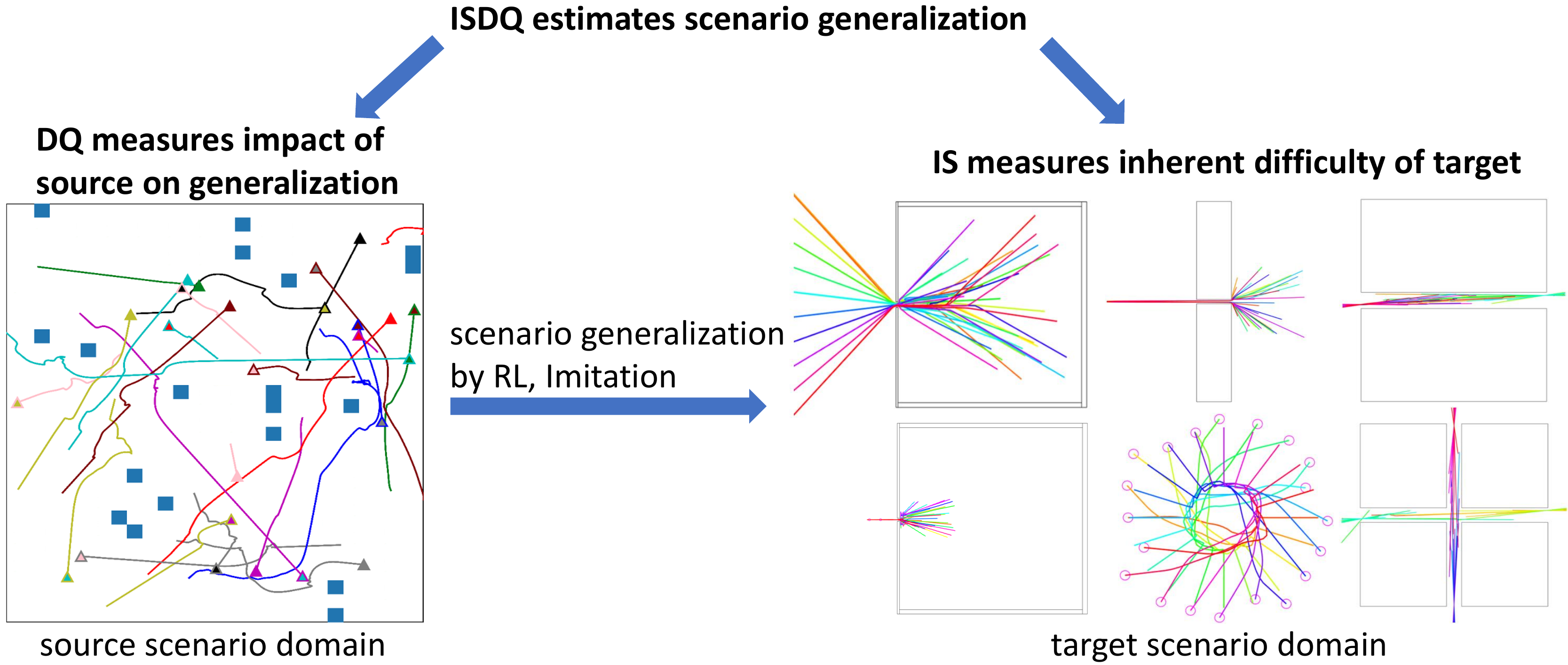}
  \caption{The utility of ISDQ is to estimate scenario generalization. It consists of (i) DQ that measures the influence of a source scenario domain on the generalization performance of a crowd learning model, and (ii) IS that measures the inter-agent interaction difficulty of a target domain for a crowd model.}
  \label{fig:ISDQ_utility}
\end{figure}

In this work, we propose an information theoretic approach that enables a training-free, model-agnostic evaluation of scenario generalization of imitation learning models and model-free reinforcement learning models. As shown in \autoref{fig:ISDQ_utility}, it estimates an intermediate proxy for scenario difficulty, the \textbf{Interaction Score (IS)}, which characterizes the task-level interaction difficulty of crowd movement in a target domain. When augmented with the \textbf{Diversity Quantification (DQ)} on the source domain, the combined \textbf{ISDQ} score is shown to offer an effective means to select the source-target domain pair likely to result in the best learning model source-to-target generalization, prior to the training of the actual imitation learning or model-free reinforcement learning models. We provide empirical evidence to demonstrate the effectiveness of the proposed \textbf{ISDQ} score.\\ 

\noindent Our main contributions are summarized in the following:
\begin{enumerate}
\item As far as we know, we are the first to propose a score (IS) that measures task-level inter-agent interaction difficulty in decentralized multi-agent navigation scenarios.
\item We propose a score (DQ) that measures the diversity of a scenario domain.
\item We analyze that the combined score (ISDQ) is able to estimate the generalization ability of learning agents, from a source scenario domain to a target scenario domain. Thus ISDQ enables rapid selection of the best source-target domain pairs among multiple possible choices prior to training and testing of the learning models.
\item We conduct systematic empirical evaluations on imitation learning models and model-free reinforcement learning models, to support the aforementioned claims, and show that ISDQ outperforms a baseline \cite{berseth2013steerplex} and a state of the art (SOTA) method \cite{corneanu2020computing}. The results are recapitulated in \autoref{tab:main_results}.\\
\end{enumerate} 

\begin{sidewaystable}
\sidewaystablefn
\caption{Summary of main empirical results in this work}
\label{tab:main_results}
\fontsize{8.5}{8.5} \selectfont
\begin{tabular}{|c|c|ccc|}
\hline
Purpose                                                                                                                                                                             & Method                                                                                                                                                                                                                                                                                                         & \multicolumn{3}{c|}{Result}                                                                                                                                                                                                                                                                                                                                                                                                                                                                                                                                                                                                                                                                                                                                                                                                                                                                          \\ \hline
\begin{tabular}[c]{@{}c@{}}efficacy tests of \\ task-level \\ inter-agent \\ interaction \\ difficulty \\ scoring \\ methods in \\ three \\ dynamic \\ characteristics\end{tabular} & \begin{tabular}[c]{@{}c@{}}for each characteristic, \\ two comparative \\ scenarios are designed. \\ A scoring method is \\ applied to them to \\ compare their \\ scores. Then we \\ check the \\ comparison results\end{tabular}                                                                             & \multicolumn{1}{c|}{\begin{tabular}[c]{@{}c@{}}temporal accumulation\\ \\ subjective assess: \\ scenario 1 $<$  scenario 2\\ \\ baseline: \\ scenario 1 $\gg$ scenario 2\\ \\ IS: \\ scenario1 $<$ scenario2\end{tabular}} & \multicolumn{1}{c|}{\begin{tabular}[c]{@{}c@{}}consequent interaction\\ \\ subjective assess: \\ scenario 1 $<$ scenario 2\\ \\ baseline: \\ scenario 1 $>$ scenario 2\\ \\ IS: \\ scenario 1 $<$ scenario 2\end{tabular}}                                                                         & \begin{tabular}[c]{@{}c@{}}movement direction\\ \\ subjective assess: \\ scenario 1 $<$ scenario 2\\ \\ baseline: \\ scenario 1 $>$ scenario 2\\ \\ IS: \\ scenario 1 $<$ scenario 2\end{tabular}                                                                                       \\ \hline
\begin{tabular}[c]{@{}c@{}}efficacy tests of \\ task-level \\ inter-agent \\ interaction\\ difficulty \\ scoring methods \\ on three \\ scenario \\ domains\end{tabular}            & \begin{tabular}[c]{@{}c@{}}in each domain, \\ comparative \\ relationships \\ among either \\ benchmarks within \\ a domain, or tasks \\ in a scenario, are \\ studied\end{tabular}                                                                                                                            & \multicolumn{1}{c|}{\begin{tabular}[c]{@{}c@{}}ExSD\\ \\ the ranking of the six \\ benchmarks by IS \\ accords with the ranking \\ by subjective assess, \\ while baseline violates \\ this ranking\end{tabular}}                                           & \multicolumn{1}{c|}{\begin{tabular}[c]{@{}c@{}}EgRD\\ \\ per-agent analysis \\ in a scenario reveals that: \\ (i) IS identifies tasks that \\ are spatially isolated from \\ other tasks; (ii) IS captures \\ temporally isolated tasks; \\ (iii) baseline results are not \\ as reasonable as IS results\end{tabular}} & \begin{tabular}[c]{@{}c@{}}SRD\\ \\ (i) IS reflects contributory \\ factors, their superpositions \\ and the dynamic \\ characteristics of \\ interactivity of tasks; \\ (ii) baseline results has a \\ large variance, cannot \\ reveal contributory factors \\ nor their superpositions\end{tabular} \\ \hline
\begin{tabular}[c]{@{}c@{}}efficacy tests of \\ estimations of \\ scenario \\ generalization\end{tabular}                                                                           & \begin{tabular}[c]{@{}c@{}}first rank ground \\ truth scenario \\ generalizations by \\ actually training and\\ testing models.\\ Then rank estimated \\ scenario generalizations. \\ Finally check whether \\ there is a consistency \\ between ground \\ truth ranking and \\ estimated ranking\end{tabular} & \multicolumn{3}{l|}{\begin{tabular}[c]{@{}l@{}}ground truth ranking: EgRD = SDD $\succ$ ExSD $\succ$ SRD\\ \\ ISDQ ranking: EgRD $\succ$ SDD $\succ$ ExSD $\succ$ SRD\\ \\ baseline ranking: EgRD $\succ$ ExSD $\succ$ SRD $\succ$ SDD\\ \\ another SOTA ranking: SDD $\succ$ ExSD = EgRD $\succ$ SRD or EgRD = SRD = SDD $\succ$ ExSD\end{tabular}}                                                                                                                                                                                                                                                                                                                                                                                                                                                                                          \\ \hline
\end{tabular}
\normalsize
\end{sidewaystable} 

The rest of the paper is arranged as the follows: in Section \ref{sec:Prior_Work}, prior works and the relationship between them and this work are discussed. In Section \ref{sec:Proposed_Method}, we introduce the proposed algorithms; the proposed framework for computing task-level inter-agent interaction is presented in Section \ref{sec:Interaction_Score}, followed by the discussion of the algorithmic strategies for concretizing the proposed framework in Section \ref{sec:Algorithmic_Strategy}. The properties of IS is discussed in Section \ref{sec:IS_properties}. The diversity quantification is introduced in Section \ref{sec:Diversity_Quantification}. In Section \ref{sec:Scenario_Generation_and_Approximation}, we apply the proposed task-level inter-agent interaction measurement and the diversity quantification measurement to estimate scenario generalization. For empirical evaluation, three sets of experiments are conducted. The details of the experimental setup and dataset are described in Section \ref{sec:Experiment_Setup}. The first set of experiments is discussed in Section \ref{sec:Efficacy_Test_in_Three_Hypotheses}. It verifies the efficacy of IS in measuring the interaction difficulty in hypothesized scenarios. The second set of experiments mentioned in Section \ref{sec:Efficacy_Tests_on_Three_Domains} tests the efficacy of IS in measuring the interaction difficulty on three data domains. The third set of experiments discussed in Section \ref{sec:Verify_Estimation_of_Scenario_Generalization} verifies the utility of ISDQ in estimating scenario generalization. Further discussions about the influence of internal simulator, trajectory clustering method, and obstacle configuration on IS, and the influence of start and destination diversities on scenario generalization are addressed in Section \ref{sec:Further_Discussion}. We conclude this paper and provide a possible future direction in Section \ref{sec:Conclusion}. 

\section{Prior Work} \label{sec:Prior_Work}
In the context of crowd movement, most works encode inter-agent interaction rules into policy \cite{helbing2000simulating,van2011reciprocal,curtis2013right,knob2018visualization,mohamed2020social}. They may also arrive at policy about interaction through trial and error \cite{schulman2017proximal,lee2018crowd,jaques2019social}, or by imitation learning \cite{ho2016generative,long2017deep}. These methods preset interaction rules or learn interaction policies to overcome the inter-agent interaction difficulty within tasks.  

Some works measure inter-agent interaction based on the agents' actual movement. These methods calculate geometric or kinematic quantities including spatial-temporal closeness \cite{knob2018visualization}, the influence of relative velocity on relative position \cite{olivier2013collision}, acceleration \cite{amirian2020opentraj}, and the time till collision \cite{karamouzas2014universal}. They rely on agents' actual trajectories, and this could be viewed as a preliminary step toward task-level inter-agent interaction. In task-level inter-agent interaction measurement, \cite{berseth2013steerplex} detects interactions by specifying the timesteps to waypoints along planned paths and checking waypoints' spatial-temporal proximity. Despite its efficiency, without simulation, it cannot capture tasks' dynamic nature.  

For generalization, in addition to exact measurement \cite{qiao2019scenario}, recently there have been attempts to predict test error without the test set, given the architecture of a model and the train set \cite{corneanu2019does,corneanu2020computing}, based on the topological structure of correlations between each pair of neurons during training. Other works explicitly or implicitly exploit interaction for trajectory interpolation, or extrapolation \cite{xu2018encoding,qiao2018role,li2019interaction}. They are the applications of interactivity. In terms of crowd analysis, there are also works related to measuring simulation quality, comparative crowd analysis, and scenario specification \cite{daniel2021perceptually,he2020informative,karamouzas2018crowd,kapadia2016scenario}.

This work is in the spirit of~\cite{berseth2013steerplex}, in that it does not require the actual trajectories of agents as input for geometric or kinematic analysis. Our method incorporates the rule-based approach~\cite{helbing2000simulating} as an internal steering model. For estimating generalization, unlike~\cite{corneanu2019does,corneanu2020computing}, our method does not need to train a model. Moreover, our estimated scenario generalizations are verified by true scenario generalizations~\cite{qiao2019scenario}. 

\begin{figure*}[h!]
\centering
\includegraphics[width=1.0\textwidth]{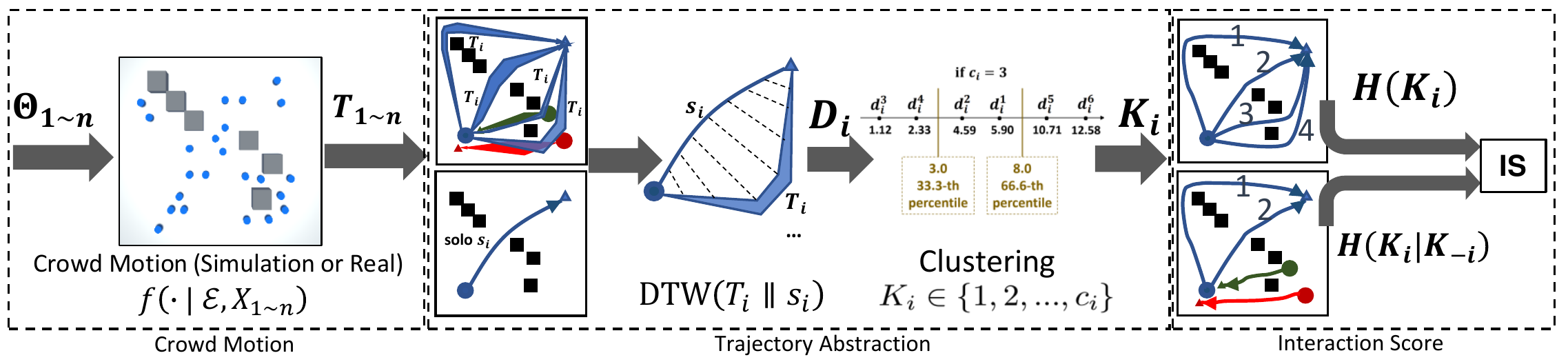}
\caption{Diagram for measuring task-level inter-agent interaction (in this example, between the agent in blue and other agents). It consists of a crowd motion block, trajectory abstraction block, and an interaction score (IS) block. The crowd motion block takes in a random parameter and transforms it into the random interactive trajectories of all agents via crowd simulation. The trajectory abstraction block compares the trajectory of an agent (e.g., the blue agent) and its solo trajectory (obtained by running the simulation of this agent within the obstacle configuration, but without other agents) to yield a DTW difference, which goes through a clustering process (we use the percentile method) and outputs an abstracted trajectory index. The IS block measures the reduced uncertainty in predicting the blue agent's abstracted trajectory modes resulting from the interaction with other agents. An agent's mode is shown with a curve with the abstracted index, representing a collection of similar trajectories for the agent. The destination of an agent is indicated by a triangle of the same color. The black rectangles in the maps denote obstacles.}  
\label{fig:System_Block_Diagram}
\end{figure*}

\section{Proposed Method} \label{sec:Proposed_Method}
The overall method of IS is introduced in Section \ref{sec:Interaction_Score} and illustrated in \autoref{fig:System_Block_Diagram}. The main idea behind IS is to assess the difference between the effort an agent makes when moving in an environment alone compared with that when it moves in the presence of all other agents. The computational procedure of IS is explained in detail in Section \ref{sec:Algorithmic_Strategy}, followed by the discussion of its properties in Section \ref{sec:IS_properties}. In Section \ref{sec:Diversity_Quantification}, we briefly introduce the method of quantifying the diversity of a scenario domain.  

\subsection{Interaction Score} \label{sec:Interaction_Score}
Assume that a scenario with obstacle configuration $\mathcal{E}$ is occupied by $n$ agents, which are collectively denoted as $X_{1 \sim n} \vcentcolon= (X_{1}, X_{2}, ..., X_{n})$, where the $i$-th agent is $X_{i}$ and the remaining agents are $X_{-i} \vcentcolon= (X_{1}, ..., X_{i-1}, X_{i+1}, ..., X_{n})$. Each agent $X_{i}$, in response to its steering model $\Theta_{i}$ within $\mathcal{E}$ and in interaction with other agents, produces a trajectory $T_{i}$. We denote this inter-agent dependent process by  $f(\Theta_{1 \sim n} \mid \mathcal{E}, X_{1 \sim n}) = T_{1 \sim n}$\footnote{We use $\Theta_{1 \sim n} \vcentcolon= (\Theta_{1}, \Theta_{2}, ..., \Theta_{n})$ and $T_{1 \sim n} \vcentcolon= (T_{1}, T_{2}, ..., T_{n})$.}. The agent $X_{i}$ can also act alone in the environment in the absence of other agents. We represent this case as $f(\Theta_{i} \mid \mathcal{E}, X_{i}) = s_{i}$, where $s_i$ is the so-called \textit{solo} trajectory of the agent. 

To compute the IS efficiently, we propose to first abstract the movement of all agents into abstraction modes and then evaluate the discrepancy using an information-theoretic measure. We discuss this process in the following section. 

\subsubsection{Trajectory Abstraction} \label{sec:Trajectory_Abstraction}
To improve representational efficiency, we aim to abstract $T_{i}$ to a finite number of modes. Each mode contains similar realizations of $T_{i}$, relative to the solo trajectories $s_i$.   We use dynamic time warping (DTW) \cite{salvador2007toward} to capture the difference: $D_{i} = \text{DTW}(T_{i} \, \| \, s_{i})$.  

$D_{i}$ serves to create the trajectory abstraction index $K_{i} \in \{1, 2, \ldots, c_i\}$ through a clustering process. In our specific case, we aim to obtain bins with a uniform distribution of trajectory instances, as illustrated in \autoref{fig:System_Block_Diagram}, leading to modes of $T_{i}$. The trajectory abstraction block abstracts $T_{i}$ of the blue agent into four modes and then indexes them with the mode set $\{1, 2, 3, 4\}$. The block outputs abstracted index tuple $K_{1 \sim n} \vcentcolon= (K_{1}, K_{2}, ..., K_{n} )$. The joint probability $P(K_{1 \sim n})$ is estimated based on the occurrence frequencies of $K_{1 \sim n}$ arising from different instances of $T_{1 \sim n}$.  

\subsubsection{Computing Interaction Score} \label{sec:Computing_Interaction_Score_1} 
We use mutual information to quantify the co-occurrence of abstracted trajectories among agents and call the result the Interaction Score (IS):
\begin{equation} \label{equ:MI_formulation_random}
    \text{IS}(K_{i}; K_{-i}) = H(K_{i}) - H(K_{i} \mid K_{-i}) = \mathop{\mathbb{E}}_{(K_{i},K_{-i})}\left[\log\frac{P(K_{i},K_{-i})}{P(K_{i})P(K_{-i})}\right]
\end{equation} 

The intuition for Eq.\eqref{equ:MI_formulation_random} is depicted in the IS block of \autoref{fig:System_Block_Diagram}. The two maps show how the blue agent’s trajectory modes are influenced by other agents’ trajectory modes. In the map above, the solo abstraction modes of the blue agent are more uncertain (four modes), leading to high $H(K_i)$. In contrast, in the bottom map, there are only two modes because of the influence of the green and red agents $K_{-\text{blue}} = K_{\text{green},\text{red}}$, encompassed in $H(K_i \mid K_{-i})$. This score naturally characterizes interactions in a scenario. For instance, when there are no interactions, $H(K_i \mid K_{-i}) = H(K_i)$, IS becomes null. On the other extreme, when the motion of one agent is completely constrained by other agents, $H(K_i \mid K_{-i}) = 0$, and IS is maximized. 

Note that the above description is the conceptual explanation of the algorithm. For a mode of trajectories, instead of being present/absent, its actual occurring probability is continuous, influenced by the movements of other agents. 

\subsection{Algorithmic Strategy of Interaction Score} \label{sec:Algorithmic_Strategy}
In Section \ref{sec:Interaction_Score}, a general framework of the IS measurement is described. This section details the computational procedure, along with a few simplification tricks. 

\subsubsection{Sampling Parameters and Trajectories} \label{sec:Sampling_Parameters_and_Trajectories}
With an empirical distribution of $\Theta_{1 \sim n}$, we sample $m$ different parameters $\{\theta_{1 \sim n}^{j}\}_{j=1}^{m}$, and run one simulation involving all agents at each sampled parameter $\theta_{1 \sim n}^{j}$,\footnote{For a parameter $\theta_{1 \sim n}^{j}$, assign $\theta_{1}^{j}$, $\theta_{2}^{j}$, ..., $\theta_{n}^{j}$ to $n$ agents correspondingly and run the simulation in the decentralized setting.} which leads to a tuple of interactive trajectories $t_{1 \sim n}^{j} \vcentcolon=  ( t_{1}^{j}, t_{2}^{j}, ..., t_{n}^{j} )$, $j$=1, 2, ..., $m$. All the crowd trajectories are organized in \autoref{tab:traj_simulation}. In the table, each row contains a tuple of interactive trajectories obtained according to one simulation parameter. Each column contains $m$ trajectories belonging to an agent by running $m$ crowd simulations. For trajectory $t_{i}^{j}$, the subscript and superscript denote the agent index and the sampled parameter index respectively. 

\begin{table}[h]
\begin{center}
\caption{Trajectories organized in a table by running simulations at $m$ sampled parameters. A row represents $n$ interactive trajectories belonging to $n$ agents by running a simulation at one sampled parameter. A column represents a tuple of $m$ trajectories of one agent by running $m$ crowd simulations. For later discussion, assume $\theta_{1 \sim n}^{j_{1}} \neq \theta_{1 \sim n}^{j_{2}}, \forall j_{1} \neq j_{2}$ in the table.}
\label{tab:traj_simulation}
\begin{tabular}{ccccc} \toprule
                                      &    agent 1    &   agent 2     &   $\cdots$   &   agent $n$   \\ \midrule
simulation at $\theta_{1 \sim n}^{1}$ &  $t_{1}^{1}$  &  $t_{2}^{1}$  &   $\cdots$   &  $t_{n}^{1}$  \\ 
simulation at $\theta_{1 \sim n}^{2}$ &  $t_{1}^{2}$  &  $t_{2}^{2}$  &   $\cdots$   &  $t_{n}^{2}$  \\ 
simulation at $\theta_{1 \sim n}^{3}$ &  $t_{1}^{3}$  &  $t_{2}^{3}$  &   $\cdots$   &  $t_{n}^{3}$  \\ 
                $\vdots$              &  $\vdots$     &  $\vdots$     &   $\vdots$   &  $\vdots$     \\ 
simulation at $\theta_{1 \sim n}^{m}$ &  $t_{1}^{m}$  &  $t_{2}^{m}$  &   $\cdots$   &  $t_{n}^{m}$  \\ \bottomrule
\end{tabular}
\end{center}
\end{table}  

\subsubsection{The Clustering Procedure} \label{sec:The_Clustering_Procedure}
For agent $X_{i}$, DTW is applied to each sampled trajectory of $( t_{i}^{1}, t_{i}^{2}, ..., t_{i}^{m})$ and its solo trajectory $s_{i}$, which results in a difference tuple $( \dtw_{i}^{1}, \dtw_{i}^{2}, ..., \dtw_{i}^{m})$, $i$=1, 2, ..., $n$. 
 
The number of modes $c_{i}$ is set to be proportional to the mean value of the difference tuple $\{ \dtw_{i}^{1}, \dtw_{i}^{2}, ..., \dtw_{i}^{m} \}$\footnote{Different agents may have different $c_{i}$, for $i$=1, 2, ..., $n$.} to capture the intuition that a larger average detour implies a higher number of modes in $( t_{i}^{1}, t_{i}^{2}, ..., t_{i}^{m})$. After that, the DTW values in $( \dtw_{i}^{1}, \dtw_{i}^{2}, ..., \dtw_{i}^{m} )$ are first sorted from least to greatest and then separated by $c_{i}-1$ percentiles (the $1 \cdot 100/c_{i}$-th percentile, the $2 \cdot 100/c_{i}$-th percentile, ..., the ($c_{i}-1) \cdot 100/c_{i}$-th percentile), which define the bin boundaries along the DTW dimension. Thus, trajectories are correspondingly grouped into $c_{i}$ modes, each containing a subset of trajectories of similar DTW values. 

\begin{figure}[htb]
 \centering
 \includegraphics[width=0.80\textwidth]{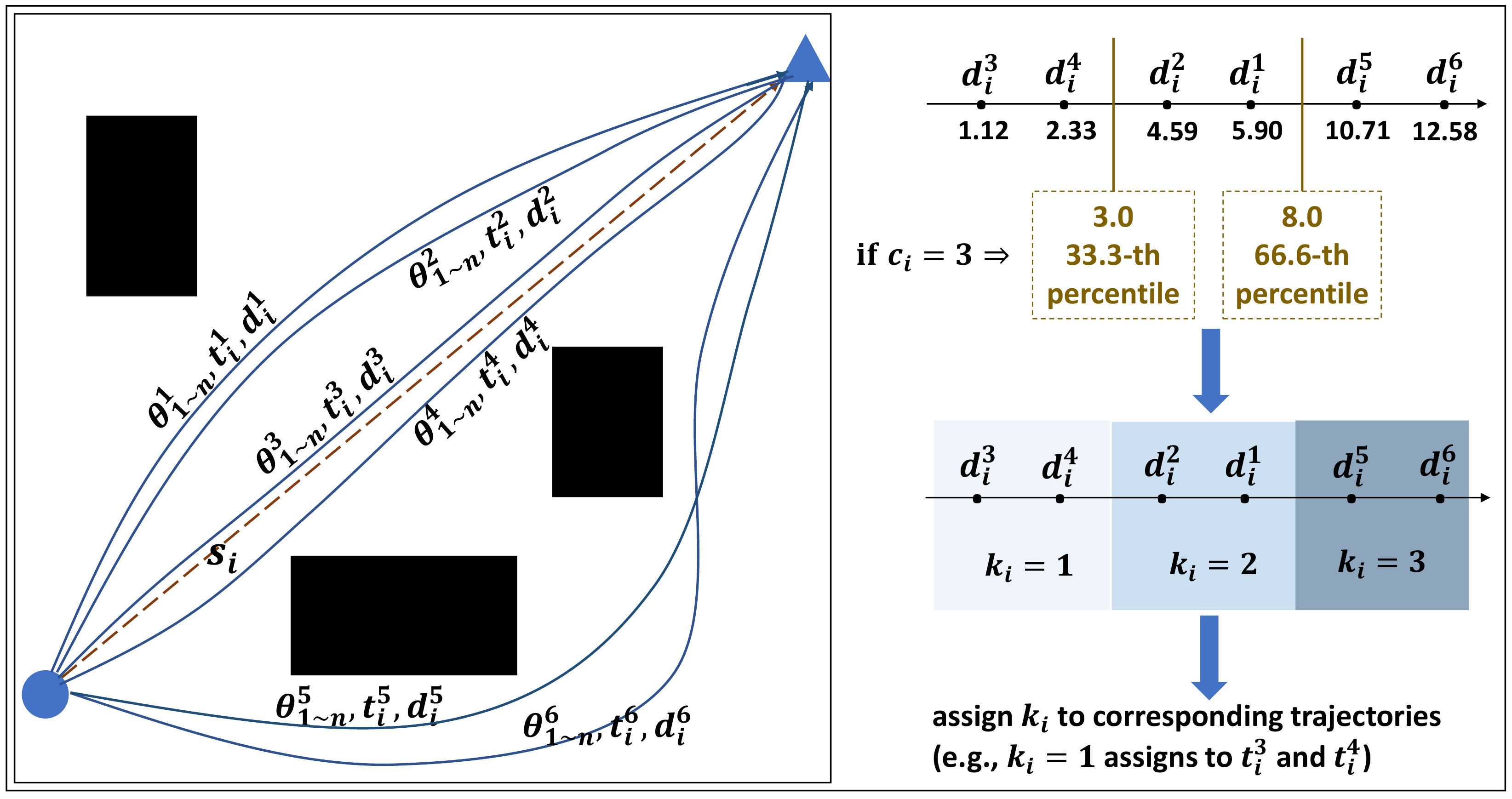}
\caption{The trajectory abstraction procedure in the DTW block. If we have six sampled parameters $\theta_{1 \sim n}^{j}$, then we have six sampled trajectories (represented by the blue solid lines) of the agent $X_{i}$, denoted as $t_{i}^{j}$, where $j = 1, 2, ..., 6$. Each trajectory is compared with the solo trajectory $s_{i}$ (represented by the red dashed line) of agent $X_{i}$, yielding a DTW value $d_{i}^{j}$. All six DTW values are sorted from least to greatest and grouped into $c_{i}$ modes by $c_{i}-1$ percentiles. An abstracted mode index $k_{i}$ is assigned to each trajectory within a mode. Thus, all trajectories of agent $X_{i}$ are assigned with a sample of $K_{i}$.}
\label{fig:MI_construct_random_variable}
\end{figure}

For agent $X_{i}$, each mode and all trajectories within are assigned with an index $k_{i} \in \{1, 2, ..., c_{i}\}$, a sample of the abstracted trajectory index $K_{i}$, $i$=1, 2, ..., $n$. We distinguish the index of a specific trajectory $t_{i}^{j}$ by $k_{i}^{j}$. The trajectory abstraction procedure of our choice to construct $K_{i}$ for agent $X_{i}$ is depicted in \autoref{fig:MI_construct_random_variable}. 

The gist of the method is that even though the random trajectory index $K_{i}$ obeys the uniform distribution due to the usage of the percentiles on the $i$-th \emph{column} in \autoref{tab:traj_simulation}, $i = 1, 2, ...n$, (meaning all marginal distributions are uniform), the joint distribution of the random trajectory indexes of all agents, $K_{1 \sim n} \vcentcolon= (K_{1}, K_{2}, ..., K_{n} )$, is in general not uniform, encoding the co-occurrences of all entities (all movement modes) in a \emph{row} of \autoref{tab:traj_simulation}. 

\subsubsection{Computing the Interaction Score} \label{sec:Computing_Interaction_Score_2}
For a concise representation, we choose the uniform distribution with $m$ different outcomes for the empirical distribution of $\Theta_{1 \sim n}$, viz. $P(\Theta_{1 \sim n} = \theta_{1 \sim n}^{j}) = \frac{1}{m}$, for $j$ = 1, 2, ..., $m$. This suggests that the $m$ different parameters in \autoref{tab:traj_simulation} constitute the support of $\Theta_{1 \sim n}$. Therefore, the distribution of the abstracted index tuple $K_{1 \sim n}$ is given by
\begin{equation} \label{equ:joint_probability_general}
\begin{split}
    P(K_{1 \sim n} = k_{1 \sim n}) = \sum_{\theta_{1 \sim n}^{j} \in \mathbb{S}} P(\Theta_{1 \sim n} = \theta_{1 \sim n}^{j}) = \frac{ \mid \mathbb{S} \mid }{m},
\end{split}
\end{equation} 
with $k_{1 \sim n}$ being an abstracted index tuple \emph{value} mapped from any parameter in \autoref{tab:traj_simulation}. In addition, $\mathbb{S} = { \{\theta_{1 \sim n}^{j} : g \circ f(\theta_{1 \sim n}^{j}) = k_{1 \sim n}\} }$, with $f$ being the simulation function, $g$ being the clustering function, and $\mid \cdot \mid$ being the cardinality of a set. In other words, $\mathbb{S}$ = $\{\theta^{1}_{1 \sim n}, \theta^{2}_{1 \sim n}, \theta^{3}_{1 \sim n}, ...\}$ is the set of parameters of the internal simulator for all agents in the scenario. These parameters have a common property: after simulation and clustering, their joint modes lead to the same \emph{value} $k_{1 \sim n}$. $\mid \mathbb{S} \mid$ counts the number of elements in the set $\mathbb{S}$. In general, this count number is not equal to the total number of parameters $m$.

Taking Eq.\eqref{equ:joint_probability_general} and \autoref{tab:traj_simulation} together, we have:
\begin{equation} \label{equ:joint_probability_table}
    P(K_{1 \sim n} = k_{1 \sim n}) = \frac{\# \text{ rows in \autoref{tab:traj_simulation}} \text{ that lead to } k_{1 \sim n} }{m}
\end{equation} 

Since $k_{i}$ represents a collection of trajectories of similar DTW values for agent $X_{i}$, $k_{1 \sim n}$ represents a particular crowd movement pattern relative to the solo trajectories, and $P(K_{1 \sim n} = k_{1 \sim n})$ reflects how easily the crowd movement pattern occurs with a simulation at a sampled parameter. 

Given the joint distribution of the full index tuple $K_{1 \sim n}$, the distribution of a partial tuple is marginalized by
\begin{equation} \label{equ:marginal_probability_xi}
\begin{split}
    P(K_{i} = k_{i}) & = \frac{\# \text{ rows in \autoref{tab:traj_simulation}} \text{ that lead to } k_{(i, \thicksim)} }{m}\\
    P(K_{-i} = k_{-i}) & = \frac{\# \text{ rows in \autoref{tab:traj_simulation}} \text{ that lead to } k_{(\thicksim, -i)} }{m},
\end{split}
\end{equation}
where $k_{(i,  \thicksim)}$ denotes any value of $k_{1 \sim n}$ with the $i$-th digit being $k_{i}$ while ignoring the values of the remaining digits. Similarly, $k_{( \thicksim, -i)}$ stands for any value of $k_{1 \sim n}$ with all the remaining digits being $k_{-i}$, regardless of the value of the $i$-th digit. 

Thus, the IS of agent $X_{i}$ in Eq.\eqref{equ:MI_formulation_random} is estimated with samples
\begin{equation} \label{equ:MI_formulation_estimation}
    \text{IS}(K_{i}; K_{-i}) \approx \frac{1}{m} \sum_{j=1}^{m} \log\frac{ P(K_{i}=k_{i}^{j},K_{-i}=k_{-i}^{j}) }{ P(K_{i}=k_{i}^{j})P(K_{-i}=k_{-i}^{j}) },
\end{equation}
where $P(K_{i}=k_{i}^{j},K_{-i}=k_{-i}^{j})=P(K_{1 \sim n}=k_{1 \sim n}^{j})$. $k_{-i}^{j}$ denotes the abstracted trajectory index tuple of agents $X_{-i}$ in the $j$-th row of \autoref{tab:traj_simulation}. $k_{1 \sim n}^{j}$ is the abstracted index tuple of all agents in the $j$-th row. 

The method is summarized in Algorithm \autoref{alg:MI_algorithm}.
\begin{algorithm}[h]
\begin{algorithmic}[1]
\small
\caption{Measure Task-Level Inter-Agent Interactivity}
\label{alg:MI_algorithm}
    \Require a scenario (obstacle $\mathcal{E}$ and tasks $X_{1 \sim n}$ of $n$ agents);
    \Statex \hspace{0.50cm} the number of sampled parameters $m$;
    \Statex \hspace{0.50cm} the parameter $\theta^{\ast}$ for solo simulation;
    \Statex \hspace{0.50cm} the proportional coefficient $\alpha$ for the number of abstracted modes
    \Ensure task-level interaction score between every task and all the remaining tasks

    \For {$i \gets 1$ to $n$}
        \State \text{path planning for task $X_{i}$;}
        \State \text{run solo simulation for $X_{i}$ at $\theta^{\ast}$, yielding $s_{i}$;}
    \EndFor
    \Statex
    
    \For {$j \gets 1$ to $m$}
        \State \text{sample parameter $\theta_{1 \sim n}^{j} \sim \Theta_{1 \sim n}$;}
        \State \text{run simulation at $\theta_{1 \sim n}^{j}$ in decentralized setting;} 
    \EndFor
    \Statex
    
    \For {$i \gets 1$ to $n$} 
        \State \text{aggregate task $X_{i}$'s trajectories, yielding $\{ t_{i}^{j}\}_{j=1}^{m}$;}
        
        \For {$j \gets 1$ to $m$}
            \State \text{compute DTW value $\dtw_{i}^{j}$ between $t_{i}^{j}$ and $s_{i}$;}
        \EndFor
        \Statex
    
        \State \text{compute mean value of $\{ \dtw_{i}^{j} \}_{j=1}^{m}$ and $c_{i}=\alpha \cdot \text{mean}$;}
        \State \text{group $\{ \dtw_{i}^{j} \}_{j=1}^{m}$ by ($c_{i}-1$) percentiles;}
        \State \text{assign each group with an index $k_{i} \in \{1, 2, ..., c_{i}\}$;}
    \EndFor
    \Statex
    
    \State \text{compute joint distribution via Eq.\eqref{equ:joint_probability_table};}
    \For {$i \gets 1$ to $n$}
        \State \text{compute marginal distributions via Eq.\eqref{equ:marginal_probability_xi};}
        \State \text{compute IS via Eq.\eqref{equ:MI_formulation_estimation}, yielding $\text{IS}(K_{i}; K_{-i})$;}
    \EndFor
    \Statex
    
    \State \textbf{Return} \text{$\text{IS}(K_{i}; K_{-i})$, $i = 1, 2, ..., n$.}
\normalsize
\end{algorithmic}
\end{algorithm}  

\subsection{Properties of Interaction Score} \label{sec:IS_properties}
Based on the properties of mutual information and probability, IS accords with the semantic meanings of the task-level inter-agent interaction difficulty. For instance,
\begin{enumerate}
  \item IS is zero in the case that agent $X_{i}$ and agents $X_{-i}$ have spatially or temporally independent tasks ($K_{i}$ and $K_{-i}$ are independent).
  
  \item If there are only two agents in the scenario (say $X_{1}$ and $X_{2}$), both will obtain the same IS score. This can be seen from the fact that mutual information is symmetric: $\text{IS}(K_{1}; K_{2}) = \text{IS}(K_{2}; K_{1})$.
  
  \item For a given scenario, if a new task (say $X_{n+1}$) is added incrementally in $X_{-i}$ (the task of agent $X_{i}$ and the existing tasks of agents $X_{-i} \vcentcolon= (X_{1}, ..., X_{i-1}, X_{i+1}, ..., X_{n})$ do not change), IS is monotonically non-decreasing. This can be seen in the following relationships: $\text{IS}(K_{i}; K_{-i})=H(K_{i})-H(K_{i} \mid K_{1}, ..., K_{i-1}, K_{i+1}, ..., K_{n}) \leqslant H(K_{i})- H(K_{i} \mid K_{1}, ..., K_{i-1}, K_{i+1}, ..., K_{n}, K_{n+1}) = \text{IS}(K_{i}; K_{-i}, K_{n+1})$.
  
  \item If the obstacle configuration $\mathcal{E}$ varies, in general, the IS measurement also changes even if the task of agent $X_{i}$ and the tasks of agents $X_{-i}$ are fixed. Thus, IS is implicitly dependent on $\mathcal{E}$.
  
  \item In IS, the task of agent $X_{i}$ and the tasks of agents $X_{-i}$ may have different start times, stemming from the property of probability that the joint distribution $P(K_{i}, K_{-i})$ does not require the events $K_{i}$ and $K_{-i}$ to occur within the same time interval (or start at the same time). One event could occur before, at the same time, or after the other. This property enables the application of IS measurement to scenarios where agents enter and/or exit at different timesteps.
\end{enumerate} 

\subsection{Diversity Quantification} \label{sec:Diversity_Quantification}
Given a scenario, its obstacle configuration is not a measurement of a physical quantity on obstacles in that scenario. Instead, its obstacle configuration embodies the way all obstacles present in the scenario, including all obstacles' information, such as obstacles' positions, shapes, boundaries, sizes, and their spatial relationships, thus called \emph{configuration}. Therefore, given a scenario, there is a specific configuration of obstacles that \emph{corresponds to} the scenario.
Precisely, the meaning of obstacle configuration are within the context of scenario \emph{domain}. For example, in \texttt{ExSD} domain, there are six different types of obstacle configurations (Evacuation 1, Evacuation 2, Bottleneck squeeze, Concentric circle, Hallway two-way, and Hallway four-way). From each type, denoted as $e_{l}$, $l=1, 2, ..., 6$, tasks can be sampled by varying the initial and/or destination positions of agents, yielding scenarios of the same type of obstacle configuration $e_{l}$. (Recall a scenario consists of both (i) obstacle configuration and (ii) tasks. If either is changed, the scenario is changed). In this view, $e_{l}$ is a label of obstacle configuration, from which numerous scenarios can be sampled. Since there are no constraints on how many scenarios are sampled from each $e_{l}$, given scenario samples drawn from a domain, one can count the number of occurrences of scenarios of the same obstacle configuration label $e_{l}$. In short, $\mathcal{E}$ is a random variable about obstacle configuration for a scenario domain, while $e_{l}$ is one of its instantiations (specific obstacle configuration label), with different number of occurrences. 

Further more, variations in the obstacle configuration $\mathcal{E}$ and tasks $X_{1\sim n}$ result in the diversity of scenarios within a domain. Characterizing this variation is essential to understanding scenario generalization. To this end, we suggest to use the negative of the joint entropy $DQ = -H(X_{1 \sim n}, \mathcal{E})$ to characterize the diversity in scenarios sampled from a (source) domain. It is further decomposed into (i) the entropy of the environment $\mathcal{E}$ and (ii) the conditional entropy of tasks $X_{1\sim n}$ under the environment $\mathcal{E}$ for this purpose. Both can be easily estimated from scenario samples. 

First, the obstacle configuration $\mathcal{E}$ may vary within the samples. Assume that among $m$ sampled scenarios from a (source) domain, there are $L$ different types of obstacle configurations. One may estimate the probability (frequency) of the occurrence for each type of obstacle configuration and then use the entropy to measure the diversity of $\mathcal{E}$:
\begin{equation} \label{equ:diversity_of_Env}
H(\mathcal{E}) = -\sum_{l=1}^{L}p(e_{l}) \log p(e_{l})
\end{equation} 

Second, the initial position of an agent, $I$, may vary within the samples. One may divide the spatial space of the environment into, say $N$ cells, and estimate the probability (frequency) for each cell where the initial position of an agent falls into, in the viewing of all agents from the (source) domain as independent and identically distributed. The same method applies to the destination of an agent, denoted as $D$. Thus one may further estimate the diversity of the paired initial position $I$ and destination $D$ of an agent with the joint entropy:
\begin{equation} \label{equ:diversity_of_initial_destination}
    H(I, D) = - \sum_{i=1}^{N}\sum_{j=1}^{N}p(i_{i}, d_{j}) \log p(i_{i}, d_{j})
\end{equation} 

Third, the diversities in the initial position $I$, the destination $D$, and the obstacle configuration $\mathcal{E}$ can be joined:
\begin{equation} \label{equ:diversity_of_scenarios}
    H(I, D, \mathcal{E}) = H(\mathcal{E}) + H(I, D \mid \mathcal{E}) = H(\mathcal{E}) + \sum_{l=1}^{L} p(e_{l}) H(I, D \mid \mathcal{E} = e_{l})
\end{equation} 

Lastly, the higher the diversity in the scenarios of a source domain, the lower the estimated target cost would be:
\begin{equation} \label{equ:Q(S)}
\text{DQ}(\mathbb{S}) = - H(I, D, \mathcal{E}),
\end{equation} 
where $\mathbb{S}$ stands for the source scenario domain. 

Note that scaling the spatial space of environments of different (source) domains may be necessary to compare them. In addition, the start time when an agent is presented into the environment could be incorporated in the formulation, depending on the amount of scenario samples. 

\section{Scenario Generalization \& Approximation} \label{sec:Scenario_Generation_and_Approximation}
\subsection{Scenario Generalization} \label{sec:Scenario_Generalization}
\cite{qiao2019scenario} proposed the notion of scenario generalization for crowd modeling -- a macroscopic view on the performance of an abstract decentralized crowd model that receives a scenario as input and outputs predicted interacting trajectories. A good scenario generalization implies a concrete steering model that behaves well in an unseen target scenario domain, measured by application-oriented metrics such as the number of agent--agent collisions:
\begin{equation} \label{equ:direct_SG}
\begin{split}
    \epsilon_{\mathbb{T}}(h) = \int_{\mathbb{T}}p_{T}(t)\cdot \text{cost}(h(t \mid \mathbb{S})) \,dt \approx \frac{1}{m} \sum_{i=1}^{m}\text{cost}(h(t_{i} \mid \mathbb{S})),
\end{split}
\end{equation}
where $\mathbb{T}$ stands for a target scenario domain with distribution $p_{T}(t)$, and $h$ is a model trained on the source domain $\mathbb{S}$. $\{t_{i}\}_{i=1}^{m}$ are $m$ target scenarios drawn from the target domain. In our study, the cost is the number of agent--agent collisions along model trajectories. 

\subsection{Approximating Scenario Generalization}\label{sec:Approximating_Scenario_Generalization}
The explicit evaluation of scenario generalization of a model $h$, Eq.\eqref{equ:direct_SG}, involves training the model on $\mathbb{S}$ and then testing the trained model on $\mathbb{T}$. However, training a model using any SOTA learning paradigm, such as reinforcement learning or imitation learning (Behavior Cloning - BC \cite{long2017deep} \& Generative Adversarial Imitation Learning - GAIL \cite{ho2016generative}), is exceptionally time-consuming, requiring repeated trials for configurations of input features, hyper-parameter turnings, and reward presetting. 

To approximate scenario generalization, our fundamental assumption is that the true cost in Eq.\eqref{equ:direct_SG} can be decomposed into two independent terms. The first term $\text{DQ}(\mathbb{S})$ encodes the influence of the source domain on the performance of a crowd model, regardless of what target scenario the model will be applied to. The second term $\text{IS}(t_{i})$ represents the interaction difficulty of the target scenario itself (how difficult the test would be), regardless of what model will be applied to test, with $\lambda$ being a trade-off factor between the two terms: 
\begin{equation} \label{equ:approximate_SG}
    \epsilon_{\mathbb{T}}(h) 
    \approx \frac{1}{m} \sum_{i=1}^{m} [\, \text{IS}(t_{i}) + \lambda \text{DQ}(\mathbb{S}) \,]
    = \left[\, \frac{1}{m} \sum_{i=1}^{m} \text{IS}(t_{i}) \,\right]  + \lambda \text{DQ}(\mathbb{S}) 
\end{equation} 

By averaging over $m$ target scenarios, the first component measures the interaction difficulty of the target domain. The second component is still $\text{DQ}(\mathbb{S})$ because it is independent of $\mathbb{T}$. We quantify $\text{DQ}(\mathbb{S})$ based on the diversity of scenarios from a source domain, according to the intuition that a model exposed to a greater diversity of scenarios during training would be more adaptive to unseen test scenarios. 

A practical use of approximating scenario generalization is that it provides a support in selecting the domain among multiple candidate domains. In the case of a single target domain and multiple source domains, with the goal to select the best source domain to train a model, according to Eq.\eqref{equ:approximate_SG}, one only needs to select the source domain with the smallest $\text{DQ}(\mathbb{S})$ term since the IS term of the target domain remains the same for all source domains. Similarly, in the case of a single source domain and multiple target domains, having the aim of selecting the best target domain for a trained model, one only needs to select the target domain with the smallest IS term. 

\section{Experiment Setup} \label{sec:Experiment_Setup}
\subsection{IS Measurement Setup} \label{sec:MI_Measurement_Setup}
To compute the IS measurement, we exploit social force \cite{helbing2000simulating} as the internal steering model family/architecture since it is computational efficient, easy to control, and one of the most widely studied and extensively validated simulation models. 

In general, the abstract crowd model does not assume homogeneous agents, since we need to represent as many steering policies for all tasks in the scenario as possible. Thus, ideally, each task should sample its own internal simulation parameter, notated by $\theta^{j}_{1 \sim n}$ ($\theta^{j}_{1}$ for agent 1, $\theta^{j}_{2}$ for agent 2, ..., $\theta^{j}_{n}$ for agent $n$) instead of $\theta^{j}$ for all $n$ agents. However, in the experiment setup, in order to simplify the crowd simulation and reduce the simulation time (with an appropriately smaller $m$ value), we experimentally set them to be identical ($\theta^{j}_{1} = \theta^{j}_{2} = ... = \theta^{j}_{n}$), and find that even in this simplified setup, IS suffices to capture the task-level inter-agent interaction difficulties and to differentiate tasks in a scenario. In other words, for simplifying the simulation in our experiment, the steering models of all interactive agents in one simulation are the same, although an agent does not know other agents' models. 
    
Each agent plans independently with A-star algorithm, leading to a sequence of planned waypoints. During the movement, an agent targets the current farthest visible waypoint (the one that is not occluded along the line of sight of the agent by obstacles or other agents) as its local destination and removes some ``outdated" waypoints.
    
The solo trajectory $s_{i}$ of an agent is obtained by (i) planning the shortest path for the agent with A-star and (ii) simulating the single agent in the environment with the simulator $\theta^{\ast}$ without other agents. During the simulation, at each step, the agent finds the current farthest planned waypoint visible and treats it as the temporary destination that yields the attractive force. This way, the ideal smooth version of the shortest path for the agent $s_{i}$ is obtained.

\begin{figure}[h!]
\small
\begin{tabular}{c c}
\hspace{-0.1cm}\includegraphics[width=0.50\linewidth]{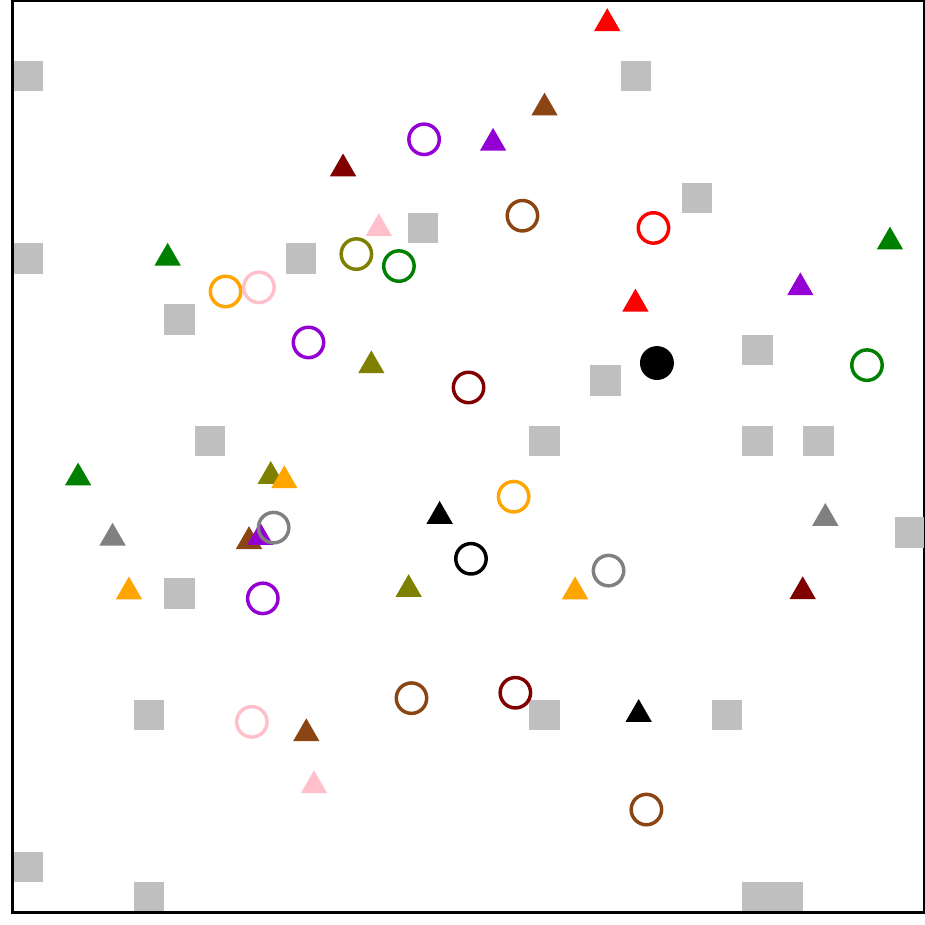}
\includegraphics[width=0.50\linewidth]{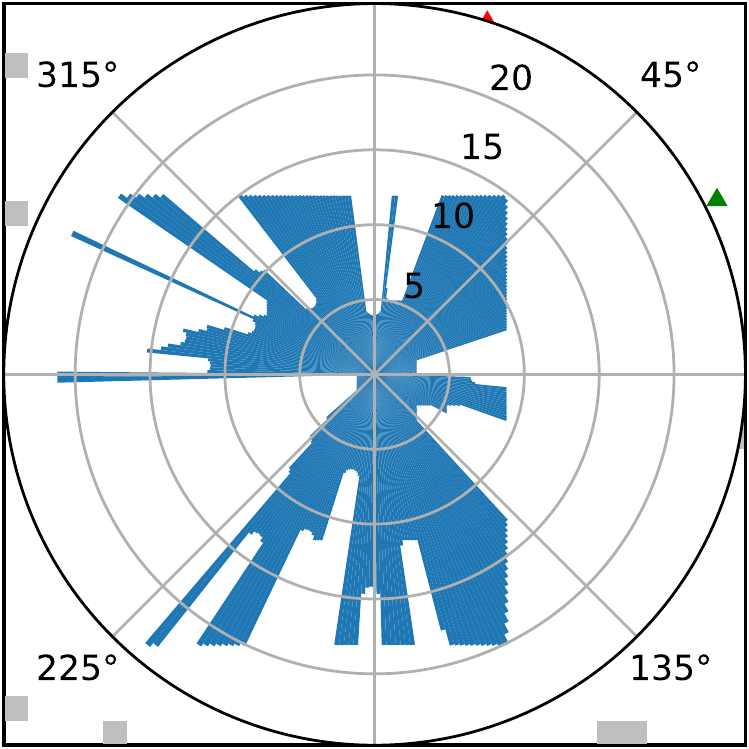}
\end{tabular}
\caption{A range map is directional distances from the center of an ego agent (the filled black agent in the left figure) to the surfaces of other agents and the surfaces of obstacles. The ego agent's range map shown in the right figure has a resolution of 360 degrees. The map is further aligned to the local coordinate system of the agent, to make a movement decision for the next step by a learning model. When computing solo trajectory, there are no other agents but obstacles in the range map. Unfilled circles are other agents. The triangles are destinations.}
\label{fig:illustration_range_map}
\end{figure}

$s_{i}$ is the solo trajectory of agent $i$ from its initial position to its destination. It is obtained by (i) planning the shortest path for agent $i$ with a deterministic planning model (in our experiment A-star algorithm is used) and (ii) simulating the single agent $i$ with the obstacles but without any other agents, using social force model set to be $\theta^{*}$. During the simulation phase, at each step, agent $i$ finds the \emph{current farthest visible} planned waypoint, which is not occluded along the line of sight of the agent by obstacles and treats it as the temporary destination that yields the attractive force (the repulsive forces come from obstacles). This way, we have the composite force and thus acceleration for the agent at that step. By accumulating steps, we obtain the solo trajectory $s_{i}$ as a smooth version of the shortest path for agent $i$.
Note: to compute the line of sight, we calculate the range map for each agent at each step, illustrated in \autoref{fig:illustration_range_map}. If there are other agents in the scenario, the range map in local coordinate system of an agent is part of the input of BC/GAIL/PPO learning model of that agent, described in \autoref{sec:Groundtruth_Scenario_Generalization_Ranking}. When calculating solo trajectory $s_{i}$, there are no other agents but obstacles.

$\theta^{\ast}$ is used to run the solo simulation. We set $\theta^{\ast}$ as $\theta^{50}$ simply to denote that we have a parameter in between the two extreme case parameters for the simulator. In fact, the algorithm is designed not to be sensitive to the choice of $\theta^{\ast}$: when running the solo simulation, there exist no other agents and thus no agent--agent repulsion ($\theta_{1}$ $\sim$ $\theta_{300}$ only serves in varying the agent--agent repulsion).

To determine the support of $\Theta$ for the uniform distribution $P(\Theta = \theta^{j}) = \frac{1}{m}$, $j$ = 1, 2, ..., $m$, \footnote{Here, the agent index is omitted in the notation since the steering models of all agents have the same parameter in a simulation.} we linearly and evenly interpolate $m = 300$ points between the two end points, $\theta^{1}$ and $\theta^{m}$, in the social force model's agent repulsion importance, repulsion-agent-A, and repulsion-agent-B parameters, shown in \autoref{tab:param_configuration} (we aim to capture only the influence of other agents on an agent; thus, we eliminate the change of other parameters like the repulsion from walls). $\theta^{1}$ imposes small agent--repulsion factors, while $\theta^{m}$ imposes significant agent--repulsion factors. Simulations on these uniformly distributed parameters form the ``max-entropy" interactive trajectory, hence the diversity of $T_{i}$ for agent $X_{i}$ in the scenario.
    
For setting $c_{i}$, we consider that although a set of interactive trajectories of an agent forming a mode might fall into a new homotopy class, modes are not identical to homotopy classes, due to the fact that in a decentralized setting (Section \ref{sec:Introduction}), modes are not planned but formed by interaction. Therefore, $c_{i}$ is not set in advance through the number of homotopy classes, but to be determined by the deviation of the sampled trajectories from the ideal solo trajectory. In our experiment, for each agent $X_{i}$, $i$ = 1, 2, ..., $n$, its number of abstracted modes $c_{i}$ is set to be proportional to the mean value of the DTWs of agent $X_{i}$, expressed by $c_{i} = \alpha \cdot \text{mean of } \{ \dtw_{i}^{1}, \dtw_{i}^{2}, ..., \dtw_{i}^{m} \}$, where the proportional coefficient $\alpha$ is set to be 0.5.
    
Regarding $\lambda$, $\lambda$ is used to adjust the relative weights of IS and DQ. It only influences the final ISDQ score. When it comes to selecting a source or a target domain, the linearization of the true target error in Eq.\eqref{equ:direct_SG} ensures that the ranking is not sensitive to the choice of $\lambda$. This is summarized in the following: (i) to select among source domains given a target domain, the IS$(\mathbb{T})$ term is fixed for all candidate source domains; one only needs to select a source domain with the lowest DQ$(\mathbb{S})$ value. (ii) To select among target domains, given a source domain, DQ$(\mathbb{S})$ is a constant for all candidate target domains; one only needs to select a target domain with the lowest IS$(\mathbb{T})$ value. Though irrelevant to the ranking, we set $\lambda=0.1$ instead of $1$ to make the range of ISDQ values close to the range of true target errors.

Last but not the least, we believe that a good ground truth metric for a learning model should be easy to understand and objective (event-based, non-parameterized, without the need to use another algorithm to describe it). In addition, considering that the main challenge for an agent equipped with a path planner and sensors in a decentralized crowd is to avoid collisions with other agents, we choose to use the number of agent-agent collisions as the ground truth metric for the exact scenario generalization of a learning model.

\begin{table}[h]
\centering
\caption{The ending parameters $\theta^{1}$ and $\theta^{m}$ of the IS measurement for the internal social force simulator. All $m=300$ parameters are linearly interpolated between them. Assume the mass of an agent is 80kg.}
\begin{tabular}{ccc} 
\toprule
                                                               &  $\theta^{1}$     &  $\theta^{m}$      \\
\midrule
Max speed (m/s)                                                &  2.6              &  2.6               \\ 
Acceleration (m/$\text{s}^{2}$)                                &  0.5              &  0.5               \\ 
Agent repulsion importance                                     &  0.0              &  10.0              \\ 
Agent body force / mass ($\text{s}^{-2}$)                      &  1500.0           &  1500.0            \\ 
Wall body force / mass ($\text{s}^{-2}$)                       &  1500.0           &  1500.0            \\ 
Sliding friction force / mass ($\text{m}^{-1}\text{s}^{-1}$)   &  3000.0           &  3000.0            \\ 
Repulsion-agent-B (m)                                          &  0.01             &  0.28              \\ 
Repulsion-agent-A / mass (N/Kg)                                &  5.0              &  60.0              \\ 
Repulsion-wall-B (m)                                           &  0.20             &  0.20              \\ 
Repulsion-wall-A / mass (N/Kg)                                 &  63.33            &  63.33             \\ 
\bottomrule
\label{tab:param_configuration}
\end{tabular}
\end{table} 

\subsection{Baseline Setup} \label{sec:Baseline_Setup}
There are only a few existing methods that directly solve the same problem: measure task-level inter-agent interaction difficulty. Thus, we extract part of the computational procedure in \cite{berseth2013steerplex} and treat it as the baseline (denoted as \textbf{BL}). The computation is as follows: (i) Plan the shortest path for each agent in a scenario using A-star, (ii) empirically assign a timestamp to each planned waypoint according to the start time of the agent and the grid size in A-star planning, and (iii) test each pair of waypoints (agents) for proximity in terms of space and time. If the distance of two waypoints is below a threshold, they are considered to have an anticipated interaction. The final baseline score would be the number of such anticipated interactions averaged over scenarios and over agents from a domain. The method measures inter-agent interaction at task level and is compared with the IS measurement. 

Particularly, an anticipated interaction is detected if $\| p_{i} - p_{j} \| < \epsilon_{d}$ and $\mid t_{i} - t_{j}\mid \, < \epsilon_{t}$, where $p_{i}$ is the spatial position of a waypoint belonging to agent $X_{i}$ at timestep $t_{i}$, and $p_{j}$ is the spatial position of a waypoint of agent $X_{j}$ at timestep $t_{j}$. $\epsilon_{d}$ and $\epsilon_{t}$ are the spatial and temporal closeness threshold respectively. In the experiment, we adjust $\epsilon_{d}=1.6m$ and $\epsilon_{t}=1s$ for all tests. 

\subsection{Another Comparison Method to Predict Generalization} \label{sec:Another_Comparison_Method_to_Predict_Generalization}
Given a point cloud and distance metric, a general pipeline of persistent homology in topological data analysis (TDA) for analyzing the structure of the point cloud first includes conducting Vietoris--Rips filtration by gradually increasing the distance threshold between two points that determines their connectivity and then computing the creation and destruction of all the point loops based on the connectivities, which are represented in the form of a persistent diagram \cite{munch2017user}. The persistent diagram characterizes the main shape of the point cloud and is robust against small perturbations on the points. 

In recent years, there have been attempts to combine the merits of topological data analysis (TDA) and machine learning. \cite{corneanu2020computing} applies the persistent homology in TDA to predict the generalization of a learning model for image recognition, facial action unit recognition, and semantic segmentation. In their method, a point is defined as the activation of a neuron of the learning model during the training procedure, and the distance metric between two points is defined as the correlation of the activations between the two neurons. The topological summary (feature) extracted from the persistent diagram is used to train a linear regressor that directly predicts the gap between the train error and the target error, assuming that the source domain and target domain obey the same distribution. Although the method makes a direct prediction of the error gap, it requires not only actually training the model but also keeping track of the correlations between activations of pairs of neurons of the model at every training epoch. 

We modify the method in \cite{corneanu2020computing} to apply it in the decentralized crowd setting. We actually train the proximal policy optimization (PPO) model \cite{schulman2017proximal} on a source domain and calculate the correlations of activations for each pair of 500 sampled neurons of the PPO model at every training epoch, based on which we compute the topological summary (feature) of the persistent diagram on the source domain to fit a linear regressor with the error being the number of agent--agent collisions. Once the PPO model is trained, we apply it to the target domain to further characterize the topological summary (feature) of the target domain. Feeding the feature of the target domain to the linear regressor, we get the predicted error gap between the source and the target and thus the target error prediction in terms of the number of agent--agent collisions. We treat the modified method as another scenario generalization estimator and denote it as PPO-TDA for comparison. 

\begin{figure}[h!]
\begin{tabular}{c c c c c c}
   \includegraphics[width=0.17\linewidth]{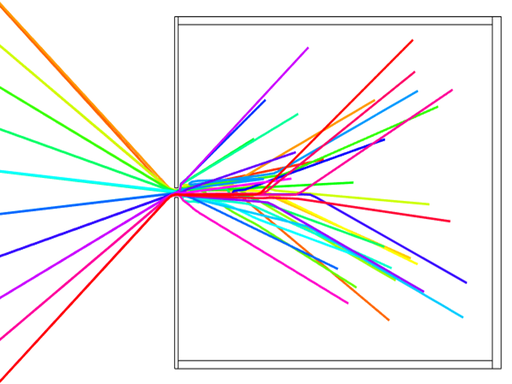}
 & \includegraphics[width=0.125\linewidth]{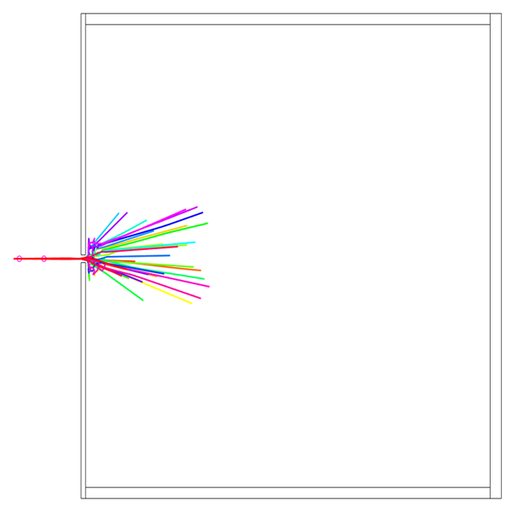}
 & \includegraphics[width=0.13\linewidth]{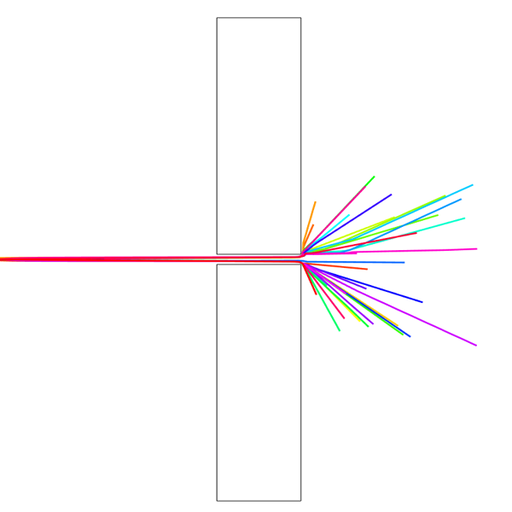}
 & \includegraphics[width=0.13\linewidth]{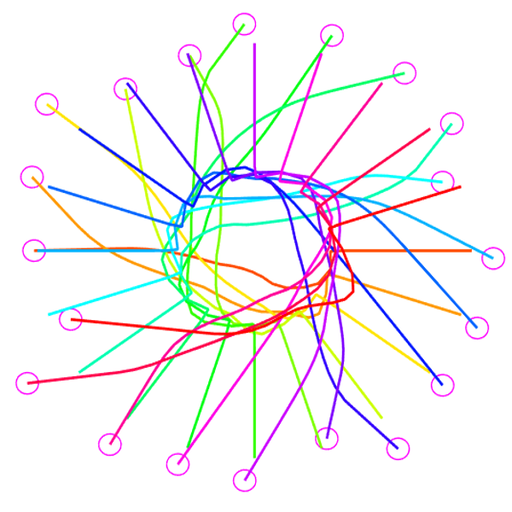}
 & \includegraphics[width=0.12\linewidth]{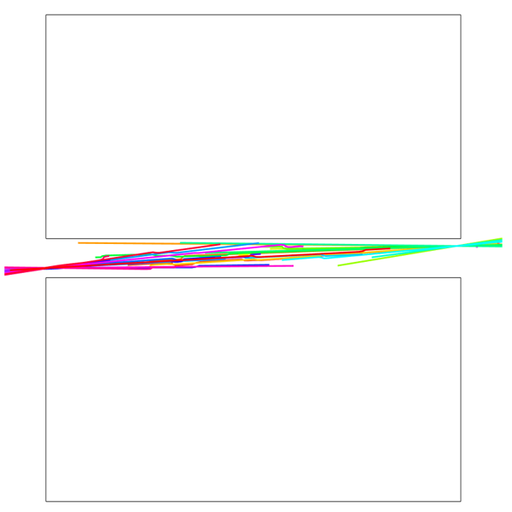}
 & \includegraphics[width=0.125\linewidth]{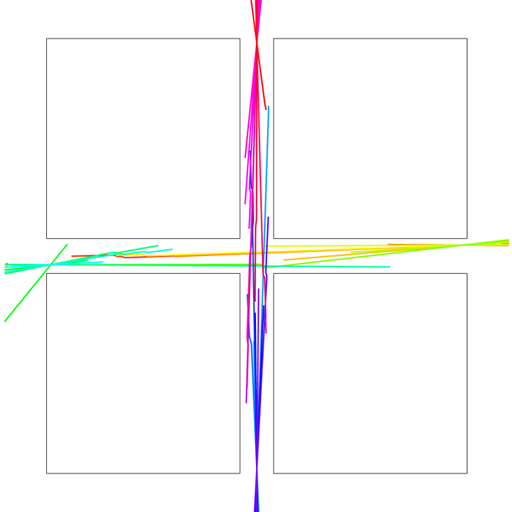} \\ 
 (a) & (b) & (c) & (d) & (e) & (f) \\ 
\end{tabular}
\caption{Six benchmarks in exocentric standard domain (\texttt{ExSD}). (a) Evacuation 1, (b) Evacuation 2, (c) Bottleneck squeeze, (d) Concentric circle, (e) Hallway two-way, and (f) Hallway four-way. Image source: \cite{qiao2019scenario}.}
\label{fig:X_domain}
\end{figure} 

\begin{figure}[h!]
\centering
\begin{tabular}{c}
   \includegraphics[width=0.48\linewidth]{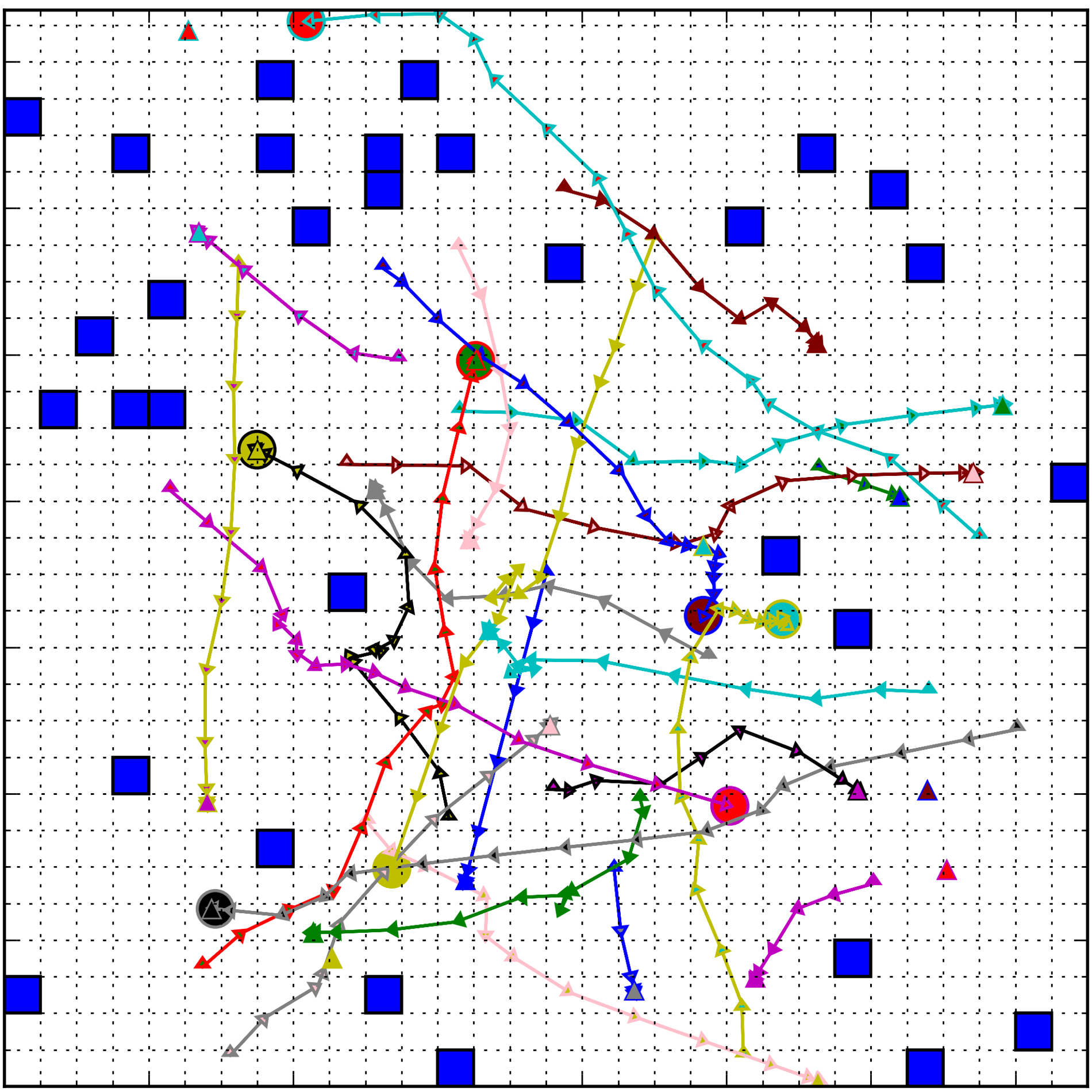}
\end{tabular}
\caption{Example scenarios sampled from egocentric representative domain (\texttt{EgRD}), shown with expert trajectories. Each agent (denoted as a circle) aims to reach its destination (a triangle of the same color) while avoiding other agents and obstacles (blue squares). Image source: \cite{qiao2019scenario}.}
\label{fig:G_domain}
\end{figure} 

\begin{figure*}[h!]
 \centering
 \includegraphics[width = \textwidth]{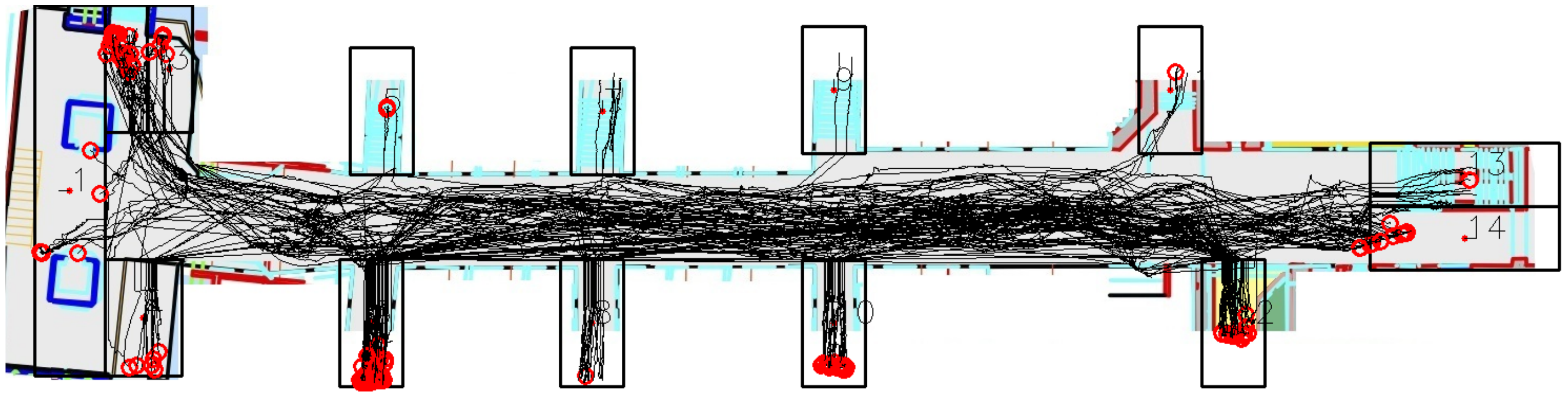}
 \caption{Visualization of Stanford real domain (\texttt{SRD}), which consists of a high density of real pedestrian trajectories collected at a train station, shown with black curves that start from the initial positions denoted by red circles. The pedestrians enter into or exit from the scenario asynchronously. Positions in a trajectory contain noise due to detection, tracking, and localization errors.}
\label{fig:R_domain}
\end{figure*} 

\begin{figure*}[h!]
 \centering
 \includegraphics[width=\textwidth]{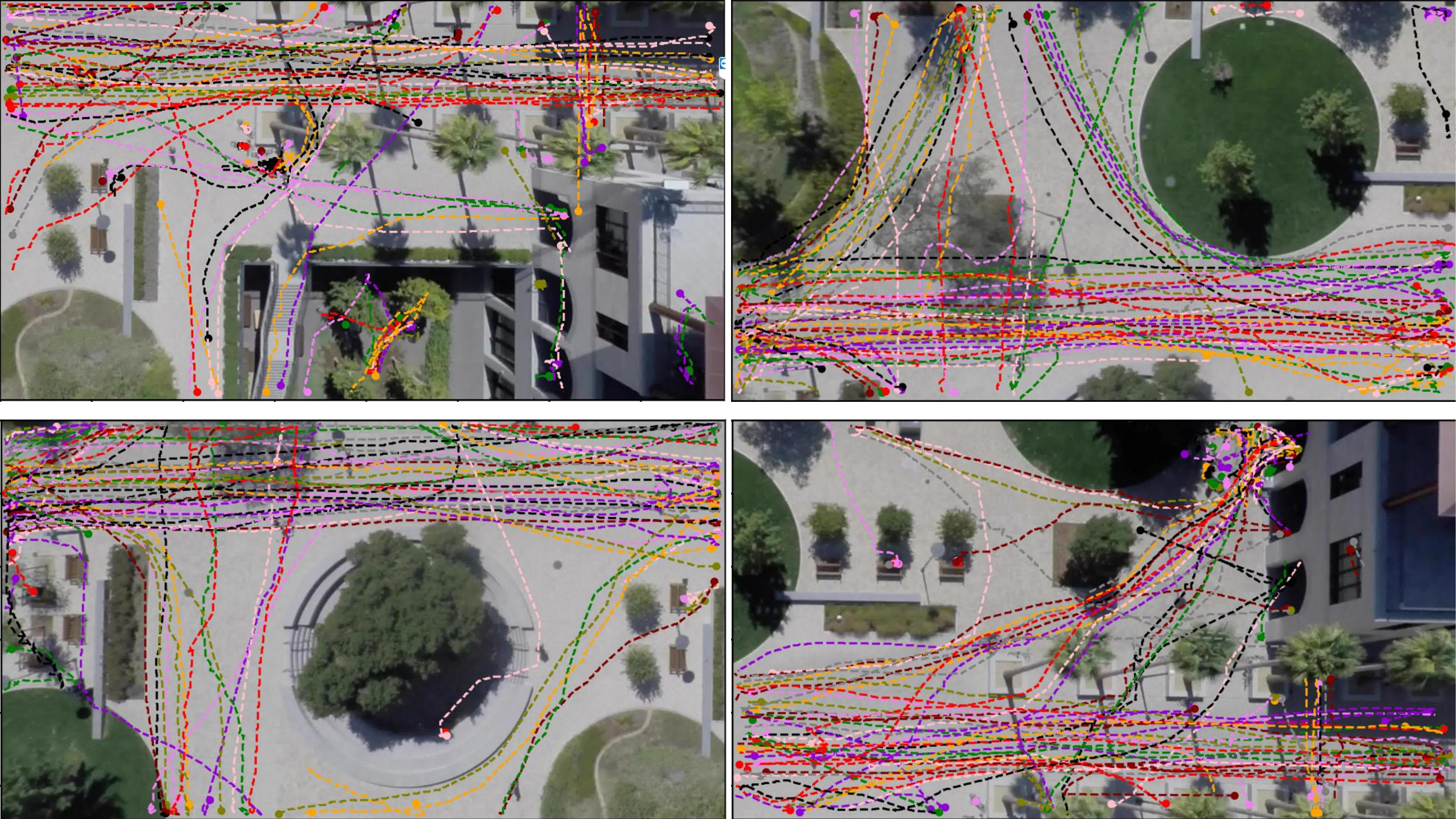}
 \caption{Visualization of the Stanford drone domain (\texttt{SDD}), which consists of four environment configurations including non-rectangular obstacles captured by a drone. Pedestrians enter or exit the scenarios sporadically and asynchronously, illustrated by dashed color curves.}
\label{fig:D_domain}
\end{figure*} 

\subsection{Descriptions of Scenario Domains} \label{sec:Descriptions_of_Scenario_Domains}
The scenario domains exocentric standard domain (\texttt{ExSD}), egocentric representative domain (\texttt{EgRD}), and Stanford real domain (\texttt{SRD}) are studied in \cite{qiao2019scenario}. In this work, we include an extra domain -- Stanford drone domain (\texttt{SDD}) -- for the estimation of scenario generalizations. 

\paragraph{Exocentric Standard Domain (\texttt{ExSD})} 
\texttt{ExSD} provides six benchmarks visualized in \autoref{fig:X_domain}. They include (i) Evacuation 1: 30 agents evacuate through a doorway of width 2.4m. (ii) Evacuation 2: Similar to Evacuation 1 but with the doorway width narrowed to 1.4m. (iii) Bottleneck squeeze: 30 agents enter and traverse a hallway of width 4.2m. (iv) Concentric circle: 20 agents symmetrically placed along a circle aim to reach antipodal positions. (v) Hallway two-way: 30 agents travel in either direction through a hallway. (vi) Hallway four-way: 32 agents arrive from and travel to any of the four cardinal directions. The benchmarks create scenarios by varying the initial positions and destinations of agents within certain regions. 
    
\paragraph{Egocentric Representative Domain (\texttt{EgRD})} 
\texttt{EgRD} provides a scenario space from where scenarios are sampled to embody small-scale but general inter-agent interactions. Both the positions of obstacles and the initial positions/destinations of agents obey a uniform distribution in the 2D spatial space, visualized in \autoref{fig:G_domain}. A scenario contains about 25 agents. Therefore, this domain contains highly diverse scenarios. 

\paragraph{Stanford Real Domain (\texttt{SRD})} 
\texttt{SRD} ~\cite{alahi2016social}, visualized in \autoref{fig:R_domain}, consists of a large set of real pedestrian trajectories collected at a train station of size $25 \text{m} \times 100 \text{m}$ for $12 \times 2$ hours by a set of distributed cameras. Identity numbers and position histories with timestamps of the pedestrians are extracted from the image sequences with detection and tracking algorithms. Creating the dataset is challenging due to the following reasons: (1) The agent density is quite high. In a time duration of 4 minutes, there are about 500 pedestrians moving in the train station. (2) Pedestrians are highly asynchronous. They enter into and exit from the train station at different timestamps, without a unified time controller. (3) The data is noisy due to the detection, tracking, and localization error and the difficulty to measure the accurate positions of the obstacles (infeasible areas). 

\paragraph{Stanford Drone Domain (\texttt{SDD})} 
\texttt{SDD} ~\cite{robicquet2016learning} contains four different obstacle configurations due to the aerial displacement of a drone that takes videos of the coupa area of Stanford campus from a bird's-eye view, shown in \autoref{fig:D_domain}. There are four videos in this domain, each containing 87 $\sim$ 123 pedestrians entering and exiting the scenario sporadically and asynchronously. Areas like lawns are also considered obstacle areas even though occasionally some pedestrians traverse across them. The original dataset is in the image coordinate system. With the Google Earth tool that measures distance, the world coordinates of obstacles can be annotated, and the pedestrian positions are consequently converted into the world coordinate system. 

In \texttt{SDD}, four scenarios are extracted from four videos and used as target scenarios. In \texttt{SRD}, 10 scenarios are extracted from a large dataset and used as target scenarios. In \texttt{ExSD}, 100 scenarios are sampled as source scenarios, while 18 scenarios are sampled as target scenarios (for each benchmark in \texttt{ExSD}, we sample scenarios by varying the initial positions and destinations of agents). Finally, in \texttt{EgRD}, 4000 scenarios are sampled as source scenarios, and 100 scenarios are sampled as target scenarios by varying both the tasks of agents and the obstacle configurations. The number of scenarios in a domain are determined according to the availability of data and the time consumption in running a scenario. 

\section{Efficacy Test in Three Hypotheses} \label{sec:Efficacy_Test_in_Three_Hypotheses}
\autoref{fig:three_tests} visualizes scenarios of the three tests to compare the measurement and the baseline based on the corresponding three hypotheses. 

\subsection{Temporal Accumulation} \label{sec:Temporal_Accumulation}
The test checks whether the bias can be restrained along the temporal accumulation of trajectories (paths). To this end, two comparative scenarios are designed, shown in \autoref{fig:three_tests_a}. In scenario 1 (left), two agents move side by side maintaining close distance toward the paralleled triangles. In scenario 2 (right), two agents move toward each other and aim to reach the antipodal positions. Intuitively, the interaction in scenario 1 is less than that in scenario 2, although within a close distance, agents in scenario 1 do not need much effort to avoid each other since their movements are in parallel. In contrast, in scenario 2, agents have to make a reciprocal turn to avoid each other, requiring significant effort. 

In terms of the average interaction over the two agents, the baseline accumulates interactions along time steps that leads to 33 $ \gg$ 1 (scenario 1 vs scenario 2), which is a large bias. The IS measurement restrains the influence of temporal accumulation and results to 1.815 $<$ 1.965 (scenario 1 vs scenario 2). 

\subsection{Consequent Interaction} \label{sec:Consequent_Interaction}
The test checks whether the consequence of a preceding interaction is incorporated into the subsequent interactions in measuring the interactivity between the task $X_{i}$ and tasks $X_{-i}$. To this end, two comparative scenarios are designed, shown in \autoref{fig:three_tests_b}. In scenario 1, agent $X_{i}$ starts from the left and aims to reach the right triangle, while agents $X_{-i}$ queue in a line and aim to reach the left triangle. In scenario 2, the agent $X_{i}$ remains the same, whereas agents $X_{-i}$ start in a V-formation, with each agent targeting a left triangle at the same horizontal level. Intuitively, the task of agent $X_{i}$ in scenario 1 is less difficult than that of agent $X_{i}$ in scenario 2: if the agent $X_{i}$ in scenario 1 manages to avoid the first agent in $X_{-i}$, it does not need to make much effort to avoid the remaining agents in a line. In contrast, in scenario 2, even though the agent $X_{i}$ avoids the first agent, it would still confront another agent in the V-formation, which requires consequent movement adjustment. 

Owing to a higher degree of waypoints overlapping, the baseline measures a higher spatial-temporal proximity of waypoints between agent $X_{i}$ and agents $X_{-i}$ in scenario 1 than that in scenario 2, leading to 7 $>$ 4 (scenario 1 vs scenario 2). The IS measurement incorporates consequences of preceding interactions, resulting in 4.127 $<$ 4.716 (scenario 1 vs scenario 2). 

\subsection{Movement Direction} \label{sec:Movement_Direction}
\begin{figure}
  \begin{subfigure}[b]{0.48\textwidth}
        \includegraphics[width=\textwidth]{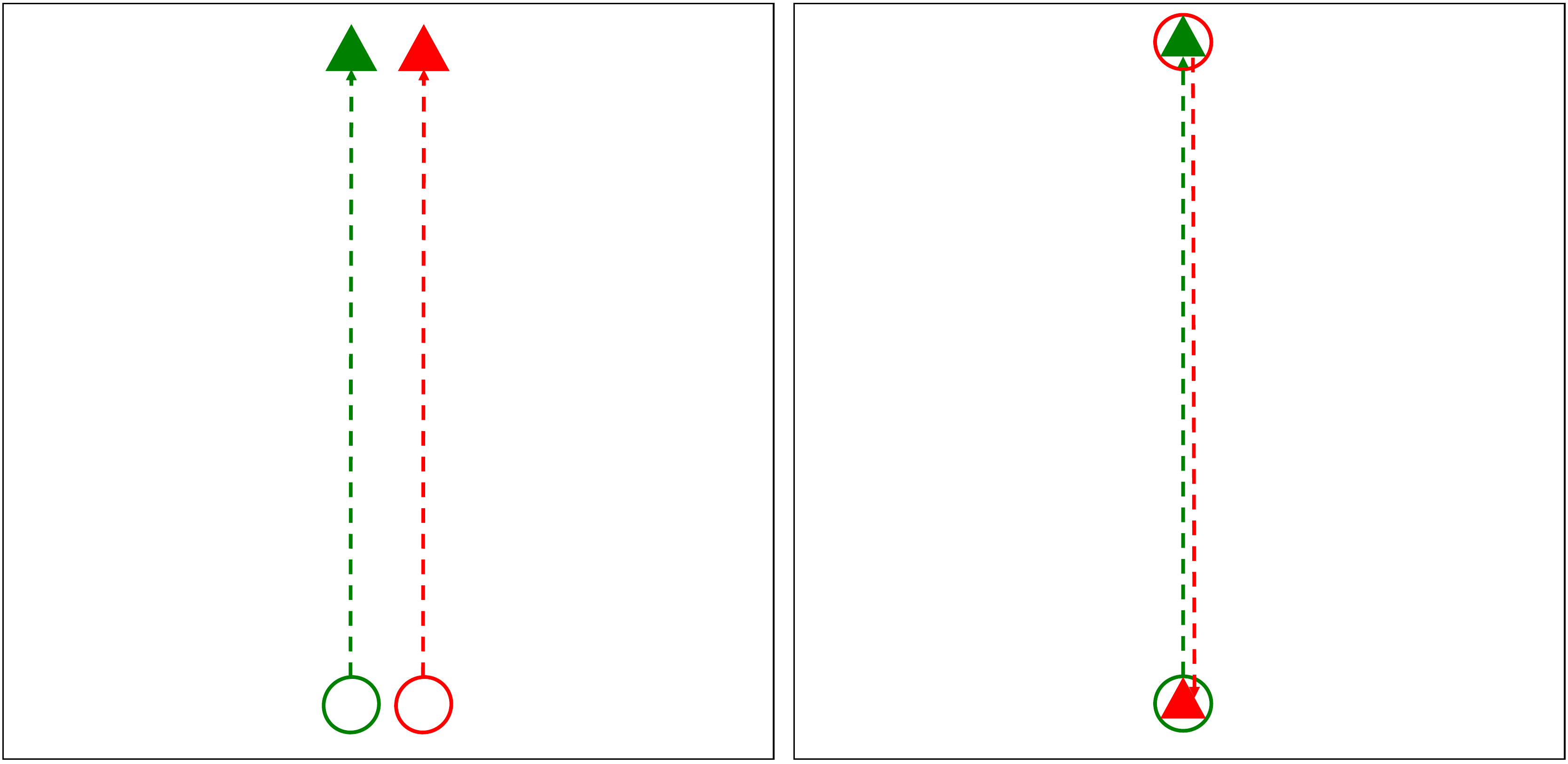}
        \caption{Temporal accumulation test}
        \label{fig:three_tests_a}
  \end{subfigure}
  \begin{subfigure}[b]{0.48\textwidth}
        \includegraphics[width=\textwidth]{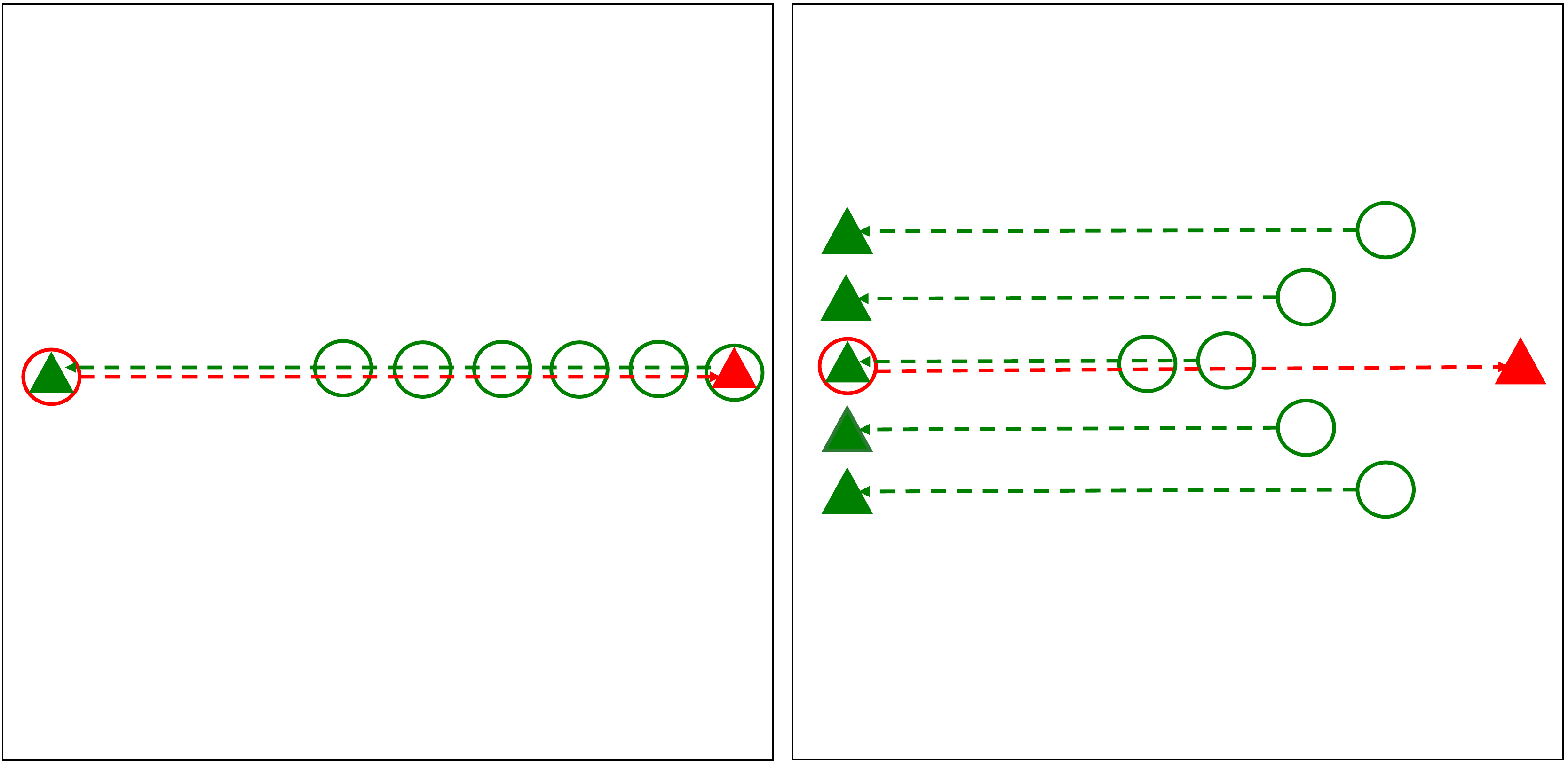}
        \caption{Consequent interaction test}
        \label{fig:three_tests_b}
  \end{subfigure}\\
  \begin{subfigure}[b]{0.48\textwidth}
        \includegraphics[width=\textwidth]{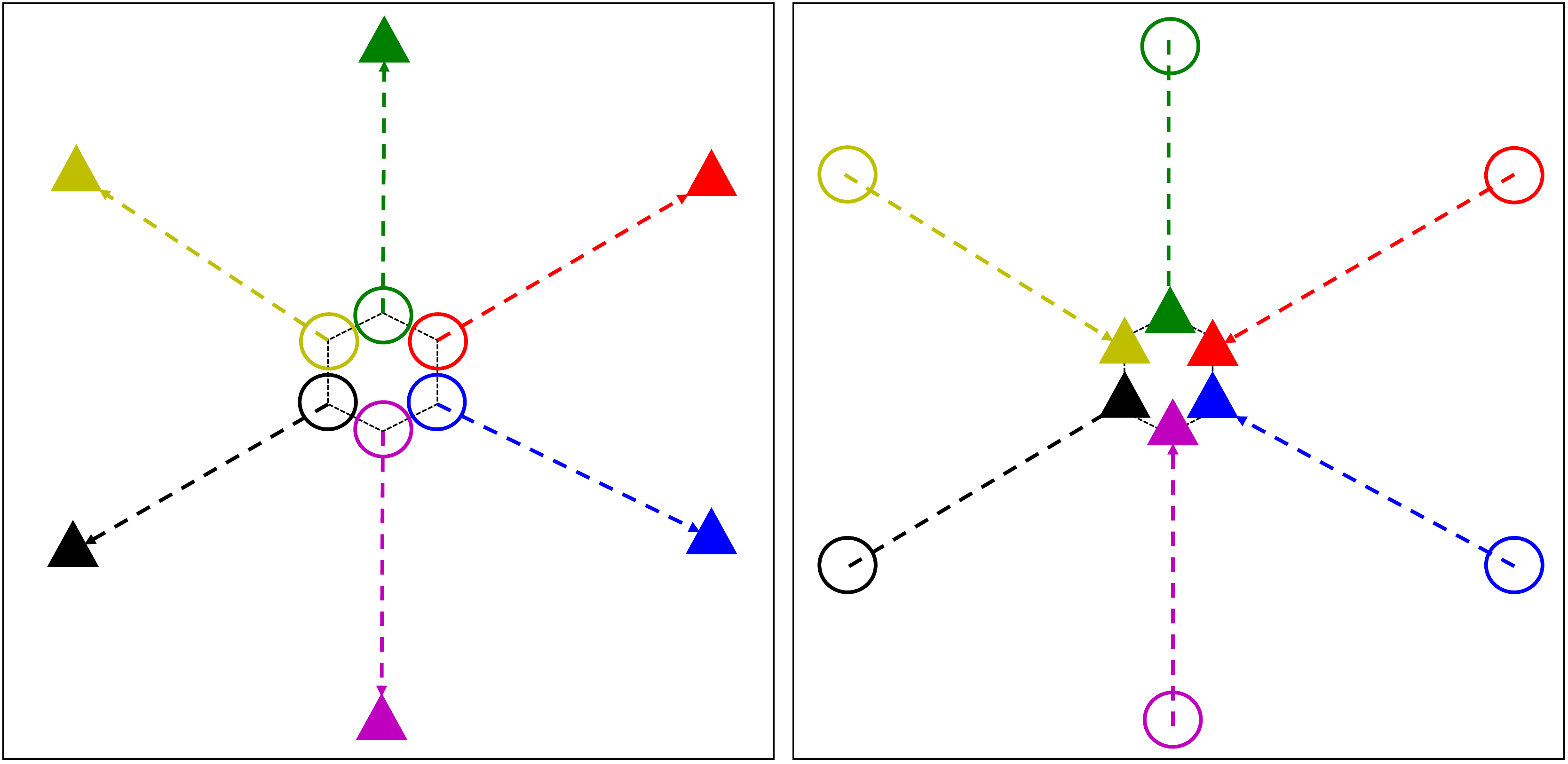}
        \caption{Movement direction test}
        \label{fig:three_tests_c}
  \end{subfigure}
\caption{Scenarios for three tests to compare the IS measurement and the baseline based on three hypotheses. Each hypothesis comprises two scenarios: scenario 1 (left) and scenario 2 (right). The initial position is represented by a circle, while the destination is represented by a triangle of the same color.}
\label{fig:three_tests}
\end{figure}

This test checks whether movement direction is reflected in the measurement. To this end, two comparative scenarios are designed, shown in \autoref{fig:three_tests_c}. In scenario 1, six agents placed tightly along a hexagonal area aim to move outward to their destinations. In scenario 2, six agents move inward toward the hexagonal area. Intuitively, the interaction in scenario 1 is lower than that in scenario 2: agents in the first scenario enjoy a clearer space during their movement. Agents in the second scenario have a reduced space when moving inwards. Without knowing other agents' destinations, agents in scenario 2 need to adapt to the situation that seems to be more and more urgent. 

For the average interaction over the six agents, the baseline is supposed to report equal interactions for the two scenarios, but planning diagonally introduces a discretization error, which leads to 9 $>$ 6 (scenario 1 vs scenario 2). The IS measurement reflects the movement direction and results in 1.109 $<$ 1.504 (scenario 1 vs scenario 2). 

In summary, the measurement is better than the baseline in terms of lowering the bias of temporal accumulation, reflecting consequent interactions and embodying the impact of movement direction on the task-level interactivity. 

\section{Efficacy Tests on Three Domains} \label{sec:Efficacy_Tests_on_Three_Domains}
\subsection{Results on Exocentric Standard Domain} \label{sec:Results_on_Exocentric_Standard_Domain}
\begin{figure}
 \centering
 \begin{subfigure}[b]{0.48\textwidth}
        \includegraphics[width=\textwidth]{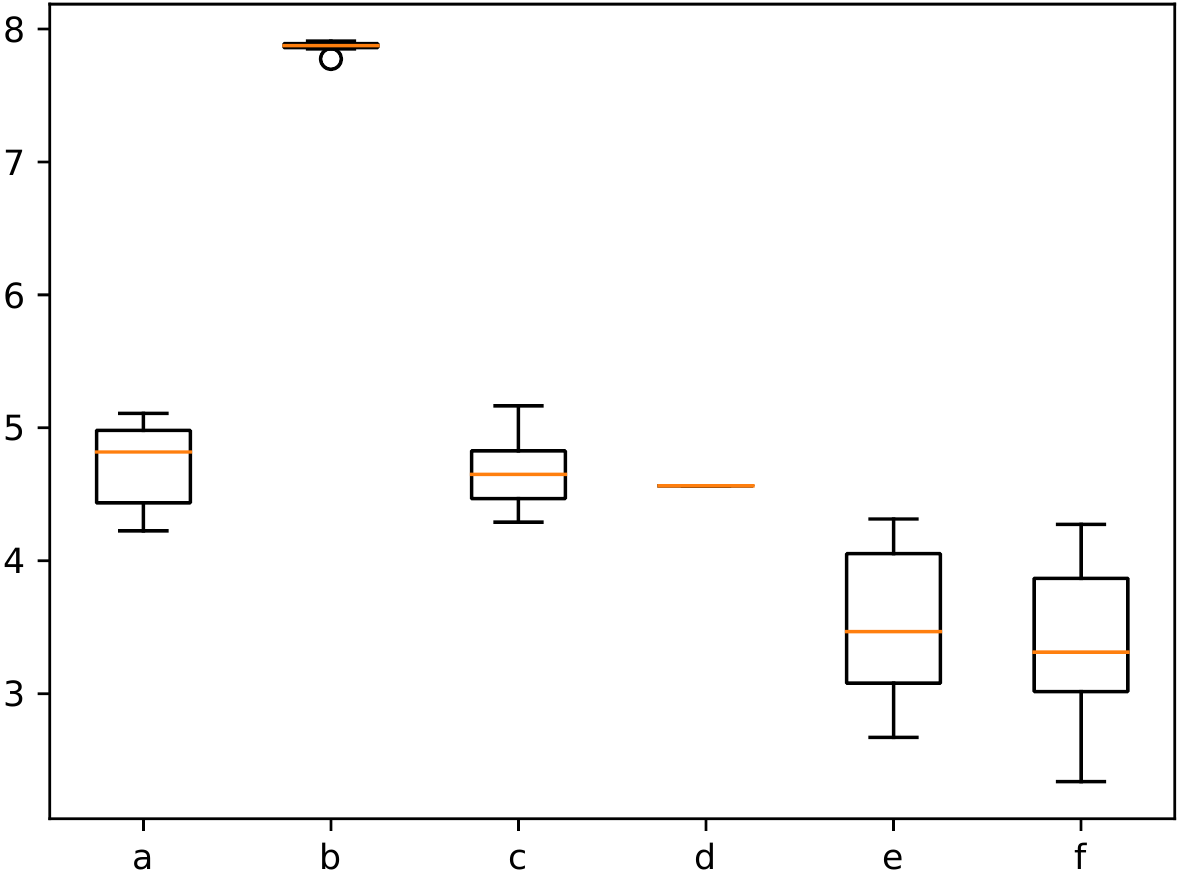}
        \caption{Boxplot of IS on \texttt{ExSD}}
        \label{fig:IS_X}
  \end{subfigure}
  \begin{subfigure}[b]{0.495\textwidth}
        \includegraphics[width=\textwidth]{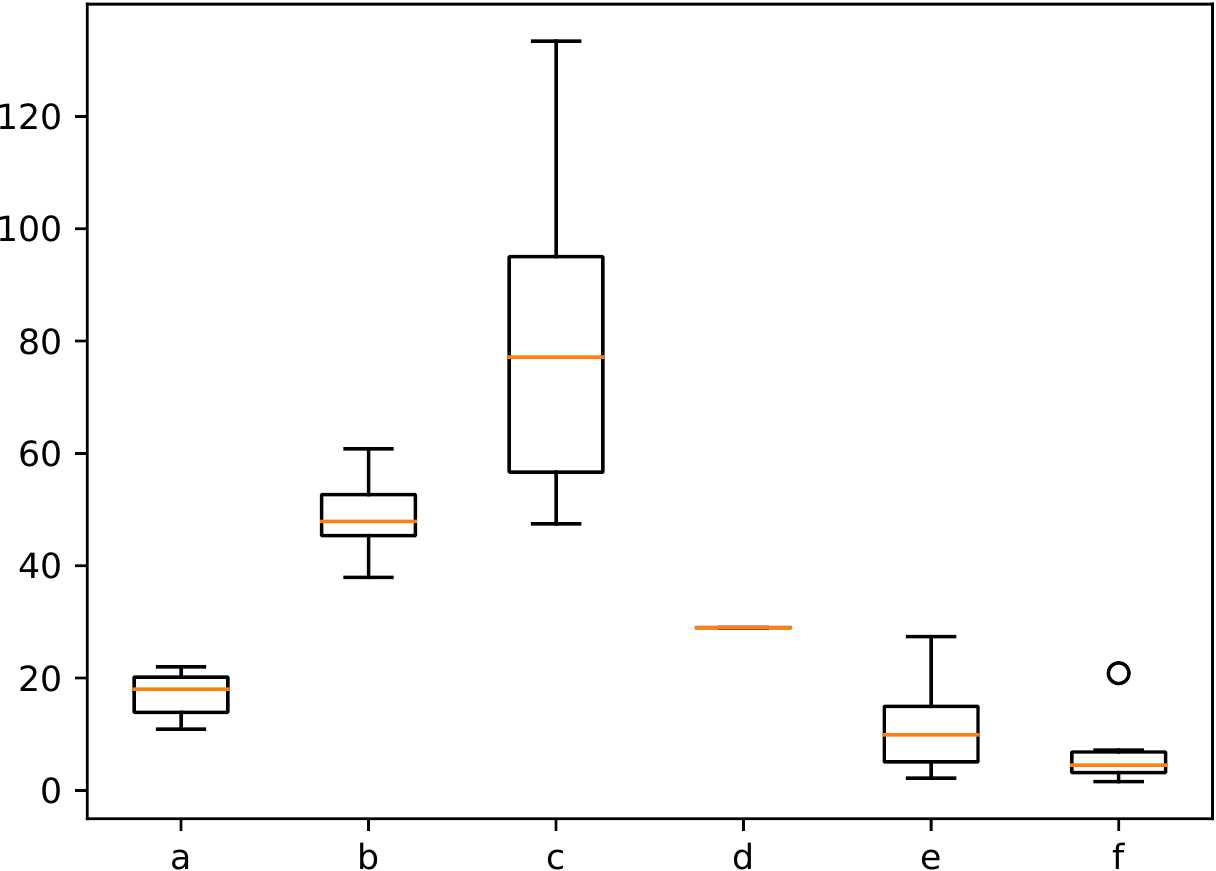}
        \caption{Boxplot of Baseline on \texttt{ExSD}}
        \label{fig:Baseline_X}
  \end{subfigure}\\
 \caption{The box plots of the IS measurement (\autoref{fig:IS_X}) and the baseline measurement (\autoref{fig:Baseline_X}) over each of the six benchmarks in \texttt{ExSD}. Along the horizontal axis, characters denote benchmarks with the order: (a) Evacuation 1, (b) Evacuation 2, (c) Bottleneck squeeze, (d) Concentric circle, (e) Hallway two-way, (f) Hallway four-way. The vertical axis is the quantification of the corresponding measurement.}
\label{fig:BoxPlot_X}
\end{figure}

\autoref{fig:X_domain} visualizes six benchmarks of \texttt{ExSD}~\cite{qiao2019scenario}. Experimental results obtained by using the IS measurement over each of the six benchmarks in \texttt{ExSD} are summarized in \autoref{fig:IS_X}. Three in-depth observations on the domain explain why the IS measure is reasonable. First, Evacuation 2 has the highest inter-agent interaction difficulty, followed by Evacuation 1 and Bottleneck squeeze. The three benchmarks have the same number of agents (30 agents). Although Bottleneck squeeze has a much longer corridor than the others, there is little interaction once agents enter the corridor. Therefore, the main contributory factor of the inter-agent interaction difficulty in these three benchmarks would be the width of the doorway. In Evacuation 2, the doorway is 1.4m wide. The doorway of Evacuation 1 is 2.4m wide, while the doorway of Bottleneck squeeze is 4.2m wide. Second, Concentric circle ranks lower than Bottleneck squeeze because it has only 20 agents. Besides, since its 20 agents are placed symmetrically along a circle, their movement forms a symmetric rotational pattern, suggesting fewer head-on interactions and reduced task interaction. Third, Hallway two-way and Hallway four-way, with 30 agents and 32 agents respectively, are relatively simpler tasks since the agents do not need to concentrate on a specific area. The hallways are much wider (16m) as well. This analysis reveals that the IS captures the contributory factors of task-level inter-agent interaction in \texttt{ExSD}. 

Experiment results with the baseline~\cite{berseth2013steerplex} over the six benchmarks of \texttt{ExSD} are summarized in \autoref{fig:Baseline_X}. Bottleneck squeeze ranks the highest, followed by Evacuation 2, Concentric circle, and Evacuation 1. This ranking does not disclose the contributory factors of task-level inter-agent interaction. 

\subsection{Results on Egocentric Representative Domain} \label{sec:Results_on_Egocentric_Representative_Domain}
\begin{figure*}[t!]
 \centering
    \begin{subfigure}[b]{0.48\textwidth}
        \includegraphics[width=\textwidth]{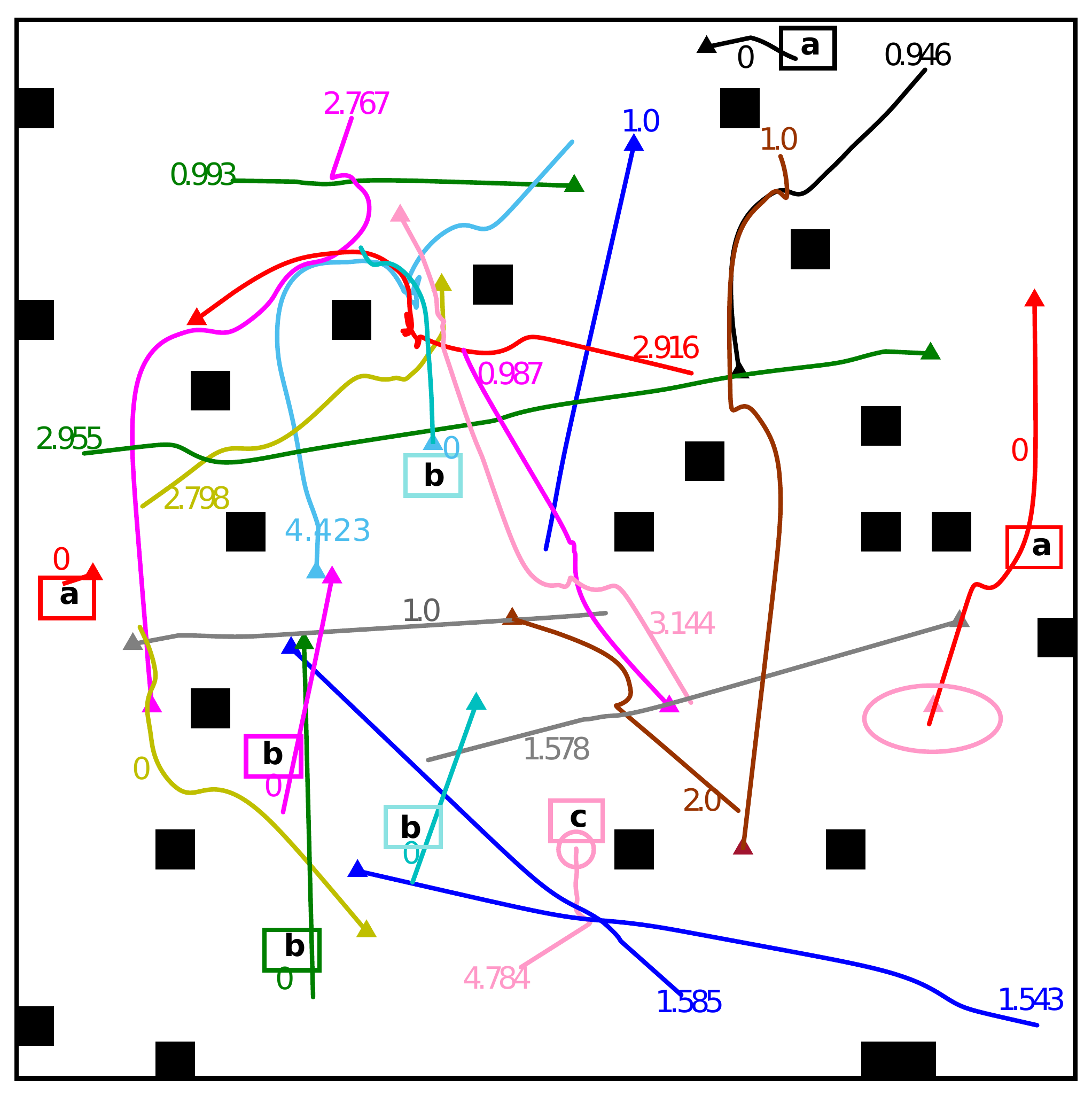}
        \caption{IS on a scenario of \texttt{EgRD}}
        \label{fig:IS_G}
    \end{subfigure}
    \begin{subfigure}[b]{0.48\textwidth}
        \includegraphics[width=\textwidth]{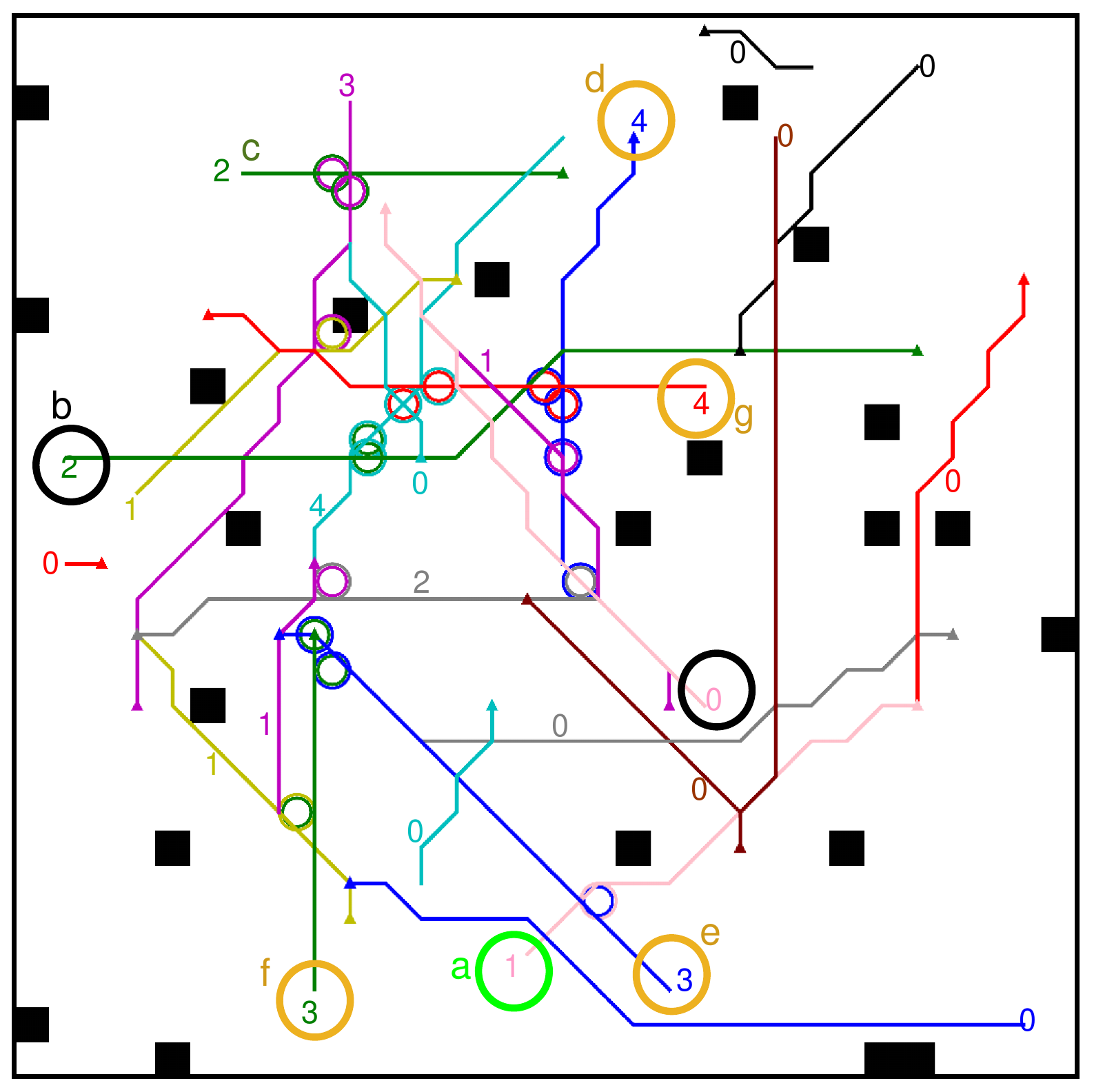}
        \caption{Baseline on a scenario of \texttt{EgRD}}
        \label{fig:Baseline_G}
    \end{subfigure} 
\caption{Visualizations of (a) the per-agent IS measurement and (b) the per-agent baseline in a scenario of \texttt{EgRD}. In both figures, filled black squares represent obstacles, and triangles represent destinations. A number labeled near a task of the same color denotes the inter-agent interactivity for that task. In (a), the trajectories are simulated at $\theta^{\ast}$. Text boxes with ``a" inside indicate that IS identifies tasks that are spatially isolated from other tasks. Text boxes with ``b" inside denote that IS captures tasks that are temporally isolated from other tasks. A text box with ``c" inside shows a case where the internal simulator fails to steer the agent to its intended destination, which is marked by an ellipse. In (b), a smaller circle, comprising two concentric rings of different colors of the involved tasks, denotes a detected interaction by the baseline, and large circles are for analysis purpose: the planned path marked with a big green circle demonstrates that the baseline method is free of the failing agent issue; the planned paths marked with big black circles have baseline results that are noticeably lower than those based on intuition; the planed paths marked with big yellow circles signify the baseline results that are higher than intuition.}
\label{fig:G_domain_measurement}
\end{figure*}

\autoref{fig:IS_G} and \autoref{fig:Baseline_G} visualize a scenario in \texttt{EgRD}. \autoref{fig:IS_G} illustrates the per-agent results of the IS measurement in the scenario. The marks on the agents emphasize the following observations: (i) our measurement identifies tasks that are spatially isolated from other tasks. These tasks, indicated by a text box with ``a", are either short or spatially far away from other tasks; and (ii) our measurement captures temporally isolated tasks, indicated by a text box with ``b". Even though these tasks seem to have intersections with other trajectories in the 2D spatial space, in the spatio-temporal space, such tasks allow an agent with regular speed to temporally avoid other agents by reaching the intersections earlier or later. Thus, our IS measurement can distinguish tasks in a scenario based on their interaction intensity. The average of the IS measurement over all agents and over 100 scenarios in \texttt{EgRD} is shown in~\autoref{tab:approximated_sg_on_target_domains}. Note that in rare cases when agents fail to reach the goal, the IS measure may introduce noise as indicated by a text box with ``c", with its intended destination depicted by an ellipse. This is a potential limitation of IS. 

The baseline result on the same setting is also summarized in~\autoref{tab:approximated_sg_on_target_domains}. \autoref{fig:Baseline_G} illustrates the per-agent baseline results on the same scenario, where a detected interaction is denoted by a small circle comprising two rings of different colors. These colors indicate the two interacting trajectories' colors. Since the baseline only uses a static A-star path, it is free of the failing agent issue (green ``a"). However, detected interactions and the measure are not as reasonable as the IS measurement. For example, planned paths marked with big black circles have baseline results that are noticeably lower than those by intuition if we compare one of them (``b") with another path (``c") in the top-most part of the figure with the same value. 

Moreover, agents marked by big yellow circles designate the baseline results that are higher than those by intuition. For instance, a blue planned path (``d") has a higher-than-intuition baseline value, resulting from the discretization when planning diagonally. The others (``e", ``f", and ``g") are affected by the fact that the baseline method cannot capture the potential changes in the subsequent interactions caused by preceding interactions if there are multiple detected interactions. For instance, if the blue path (``e") could detour a little to avoid the pink (``a") consequently, the corresponding blue agent may also prevent the later spatial-temporal encounter with the green (``f"). Besides, the red path (``g") could have had a path belonging to a different homotopy~\cite{bhattacharya2010search} under the influence of other agents. 

\subsection{Results on Stanford Real Domain} \label{sec:Results_on_Real_Domain}
\begin{figure*}[t!]
 \centering
    \begin{subfigure}[b]{0.24\textwidth}
        \includegraphics[width=\textwidth]{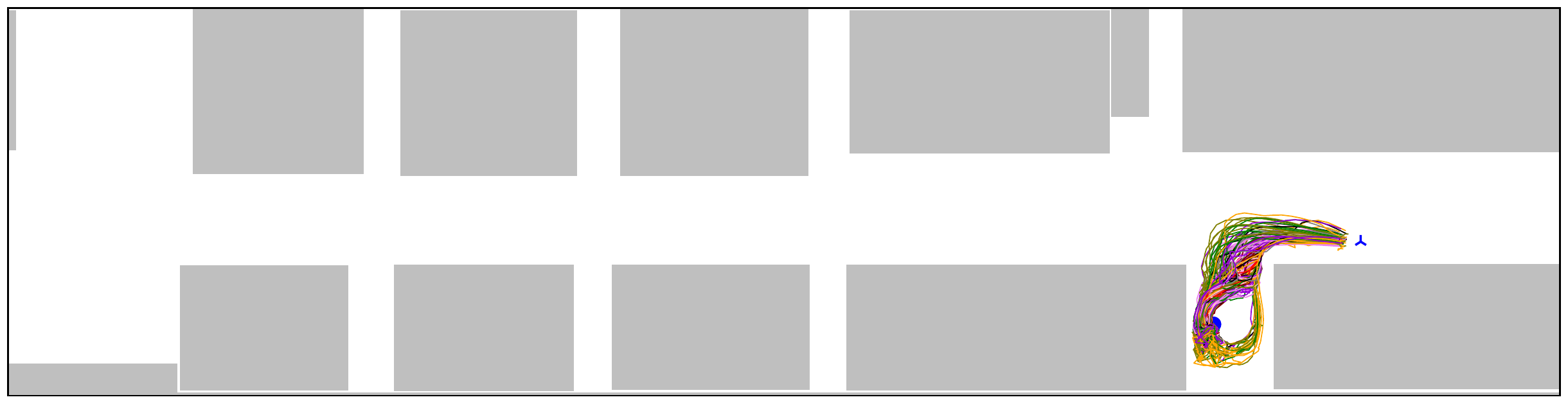}
        \caption{IS = 3.91}
        \label{fig:R_domain_measurement_a}
    \end{subfigure}
    \begin{subfigure}[b]{0.24\textwidth}
        \includegraphics[width=\textwidth]{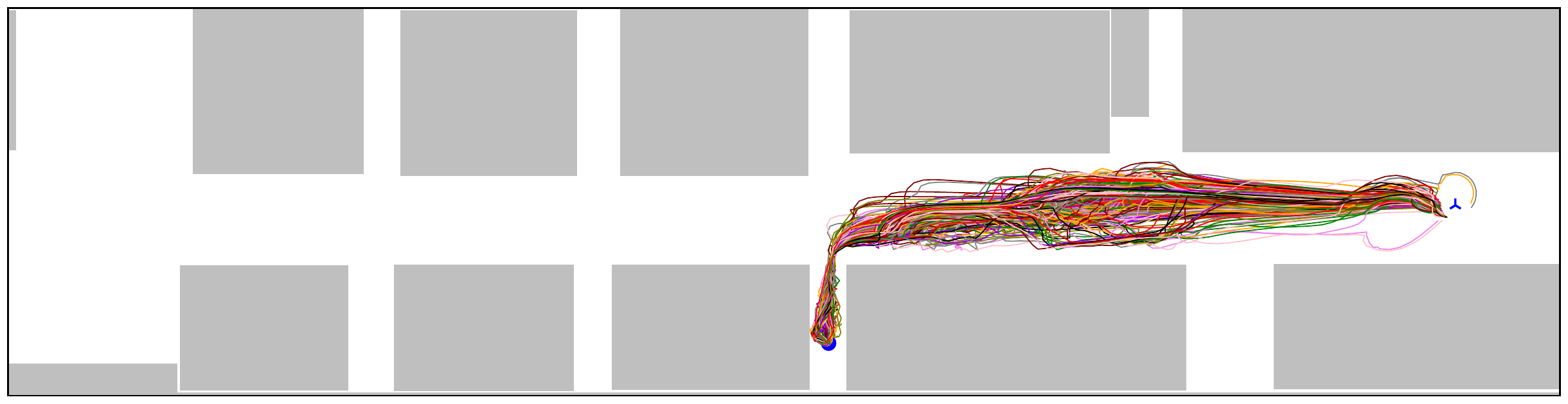}
        \caption{IS = 5.28}
        \label{fig:R_domain_measurement_b}
    \end{subfigure}
    \begin{subfigure}[b]{0.24\textwidth}
        \includegraphics[width=\textwidth]{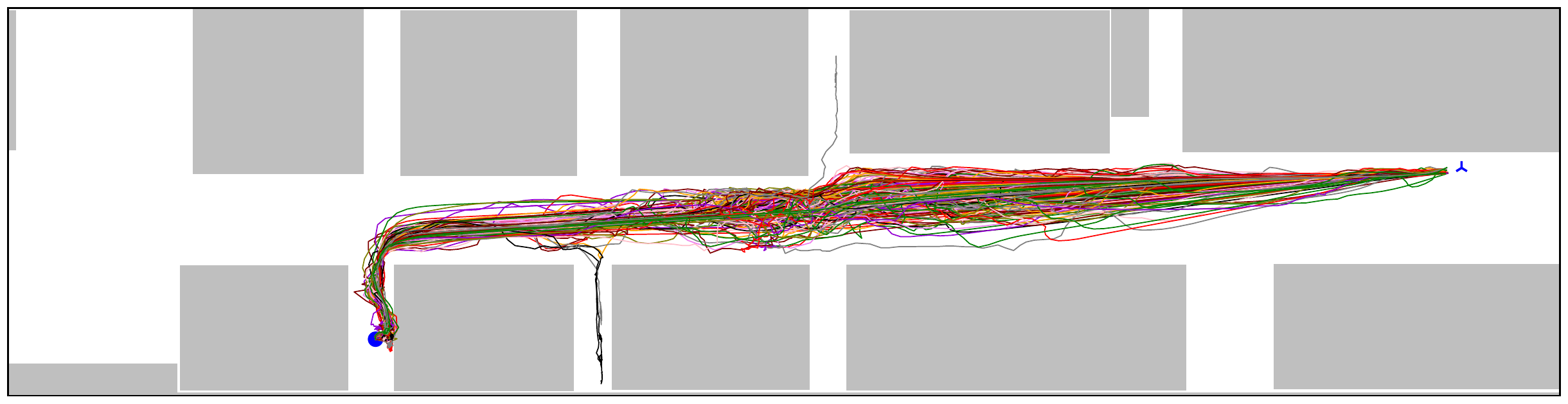}
        \caption{IS = 6.31}
        \label{fig:R_domain_measurement_c}
    \end{subfigure}
    \begin{subfigure}[b]{0.24\textwidth}
        \includegraphics[width=\textwidth]{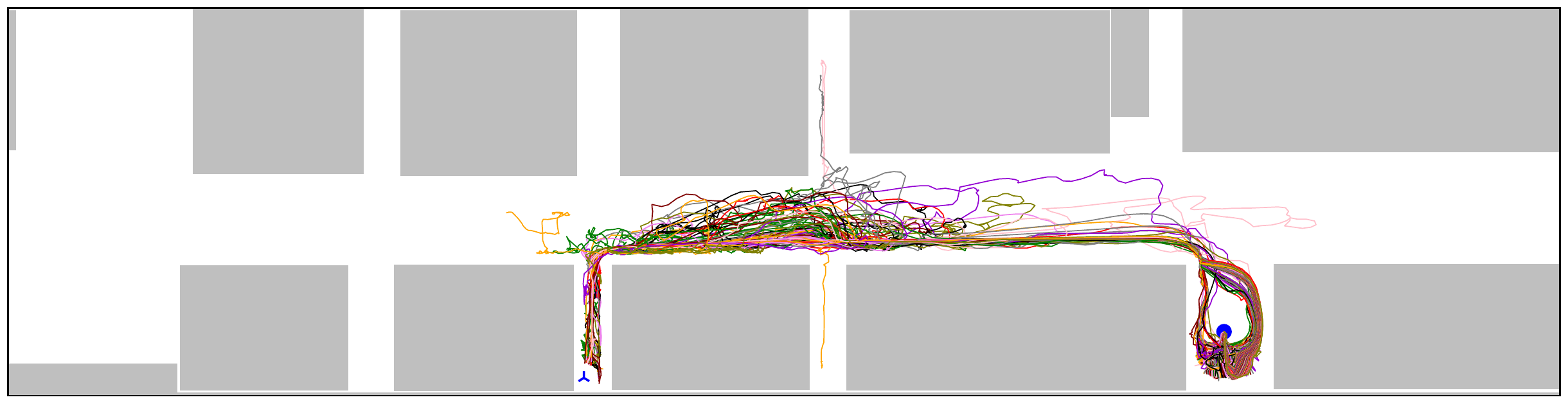}
        \caption{IS = 7.27}
        \label{fig:R_domain_measurement_d}
    \end{subfigure}
    \\
    \begin{subfigure}[b]{0.24\textwidth}
        \includegraphics[width=\textwidth]{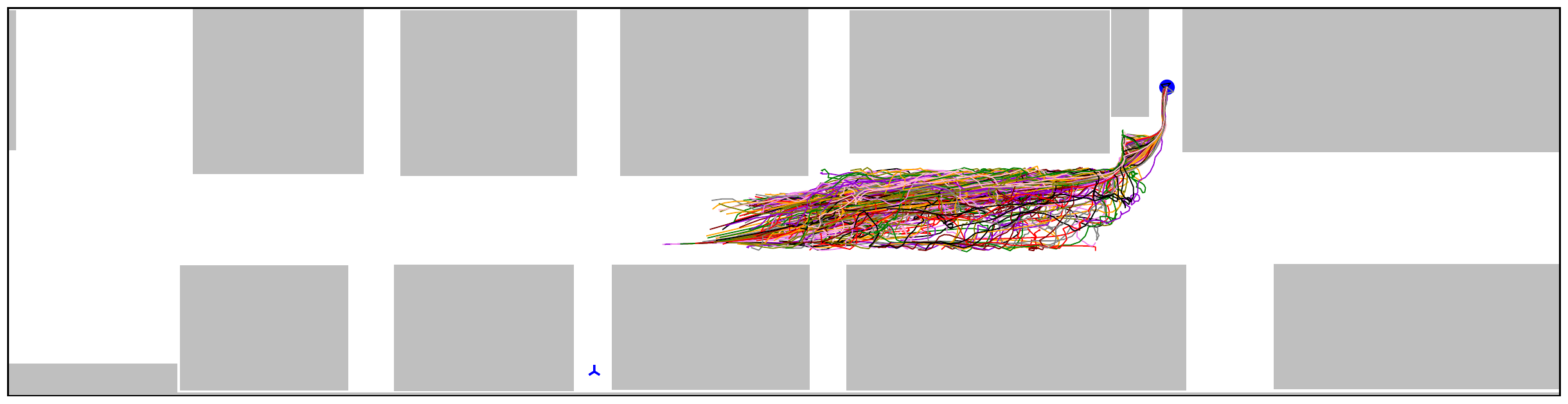}
        \caption{IS = 7.58}
        \label{fig:R_domain_measurement_e}
    \end{subfigure}
    \begin{subfigure}[b]{0.24\textwidth}
        \includegraphics[width=\textwidth]{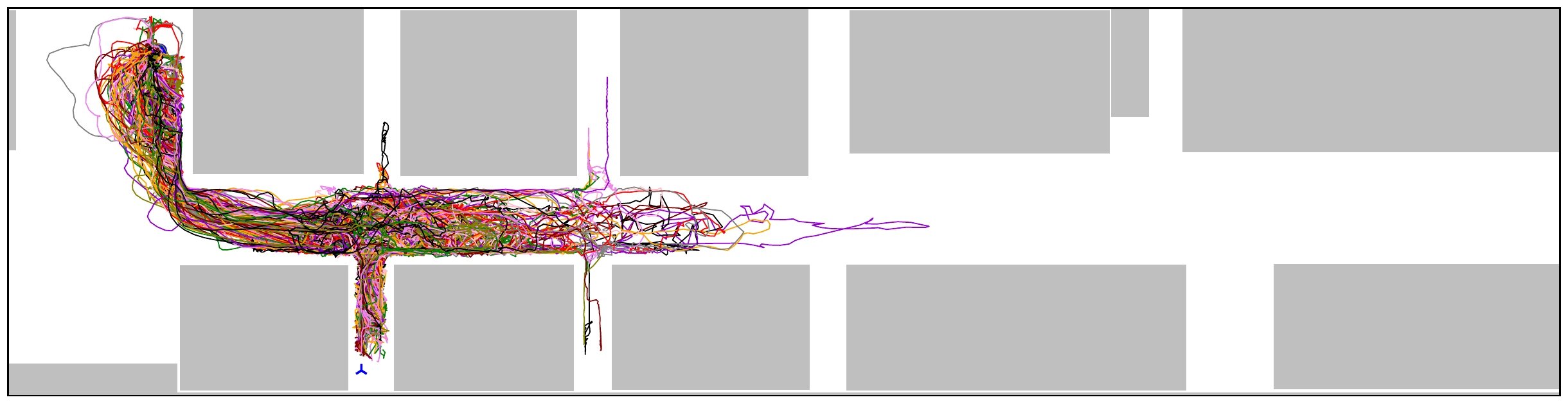}
        \caption{IS = 7.98}
        \label{fig:R_domain_measurement_f}
    \end{subfigure}
    \begin{subfigure}[b]{0.24\textwidth}
        \includegraphics[width=\textwidth]{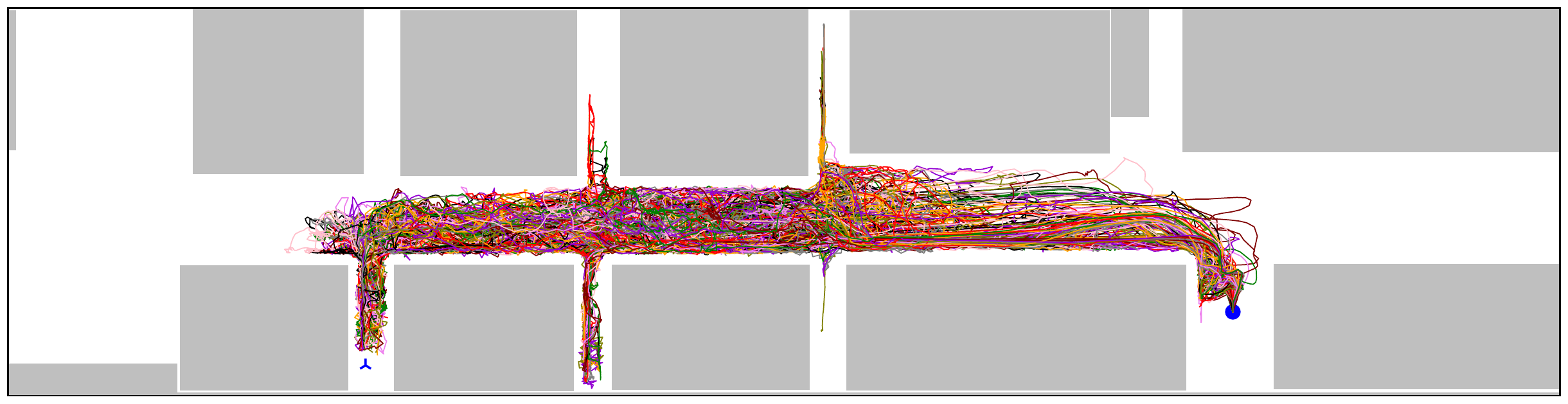}
        \caption{IS = 8.18}
        \label{fig:R_domain_measurement_g}
    \end{subfigure}
    \begin{subfigure}[b]{0.24\textwidth}
        \includegraphics[width=\textwidth]{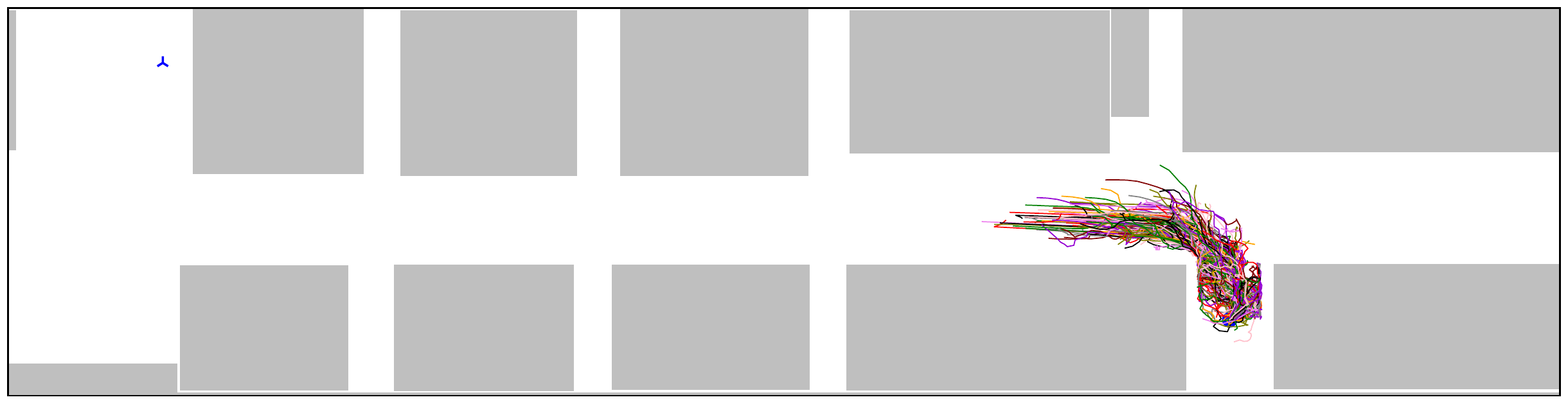}
        \caption{IS = 8.20}
        \label{fig:R_domain_measurement_h}
    \end{subfigure}
\caption{Trajectory sets of eight agents in a scenario in Stanford real domain. A subplot shows a set of trajectories for one agent (task), starting from a blue circle and ending with a blue clover shape. The trajectory set of an agent $X_{i}$ comes from the tuple $\{t_{i}^{1}, t_{i}^{2}, ..., t_{i}^{m}\}$, obtained by simulating all agents in the scenario at $m=300$ parameter points and choosing those trajectories that belong to the agent. The color of a trajectory denotes its abstracted mode index for that task. Gray areas represent the obstacles. The interaction difficulty based on IS measurement is given in a subplot. Zooming in is recommended for a better view.}
\label{fig:R_domain_measurement}
\end{figure*}

Eight trajectory sets of eight agents in one scenario of \texttt{SRD} are shown in \autoref{fig:R_domain_measurement}. In \autoref{fig:R_domain_measurement_a}--\autoref{fig:R_domain_measurement_c}, as the destination becomes further away from the initial position, the IS increases correspondingly. However, the length of a task is not the sole factor that influences the measurement. For example, \autoref{fig:R_domain_measurement_d} has a higher IS than \autoref{fig:R_domain_measurement_c} although the task length in \autoref{fig:R_domain_measurement_d} is shorter than that of \autoref{fig:R_domain_measurement_c}. This is natural as the destination in \autoref{fig:R_domain_measurement_d} is located at a narrow entrance/exit. This is further observed in \autoref{fig:R_domain_measurement_e}, where agents cannot reach the destination at a narrow entrance/exit due to congestion. In \autoref{fig:R_domain_measurement_f} and \autoref{fig:R_domain_measurement_g}, a significant amount of trajectories are repelled at wrong entrances/exits, leading to an even higher interaction. An agent is repelled into a wrong entrance/exit when it confronts a group of agents. Lastly, the IS in \autoref{fig:R_domain_measurement_h} is higher than that of \autoref{fig:R_domain_measurement_e}. This is because the trajectories in \autoref{fig:R_domain_measurement_h} are stuck at an earlier stage of the task, whereas in \autoref{fig:R_domain_measurement_e}, they are not. These visualizations reveal that the IS measurement reflects not only contributory factors but also their superpositions and the dynamic characteristics of the interactivity of tasks. The average of the IS measurement over all agents and over ten scenarios in \texttt{SRD} is shown in~\autoref{tab:approximated_sg_on_target_domains}. 

\begin{figure*}[t!]
 \centering
    \begin{subfigure}[b]{0.24\textwidth}
        \includegraphics[width=\textwidth]{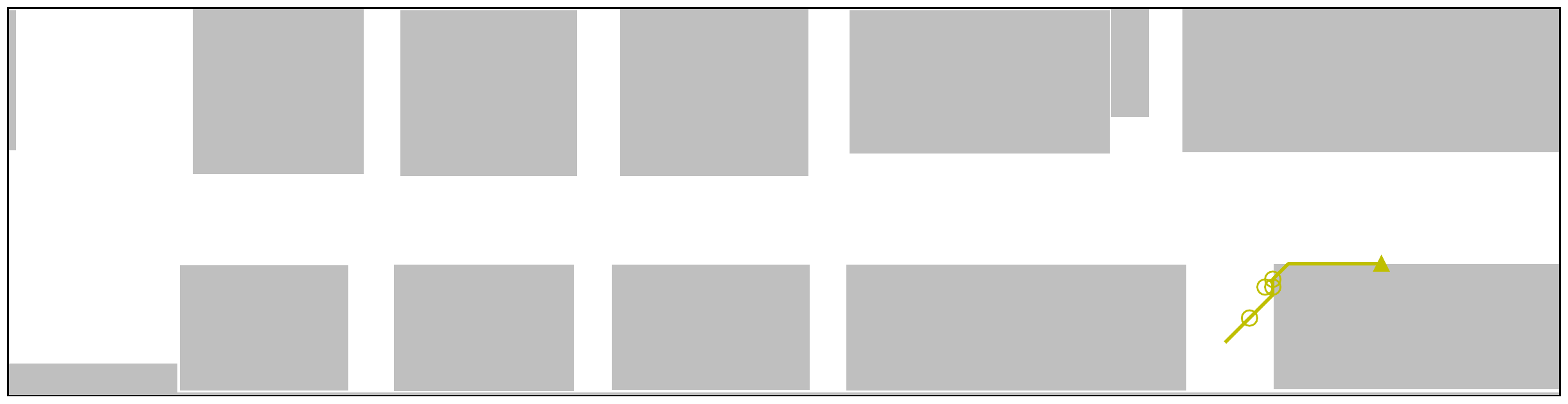}
        \caption{Baseline = 4}
        \label{fig:R_domain_baseline_a}
    \end{subfigure}
    \begin{subfigure}[b]{0.24\textwidth}
        \includegraphics[width=\textwidth]{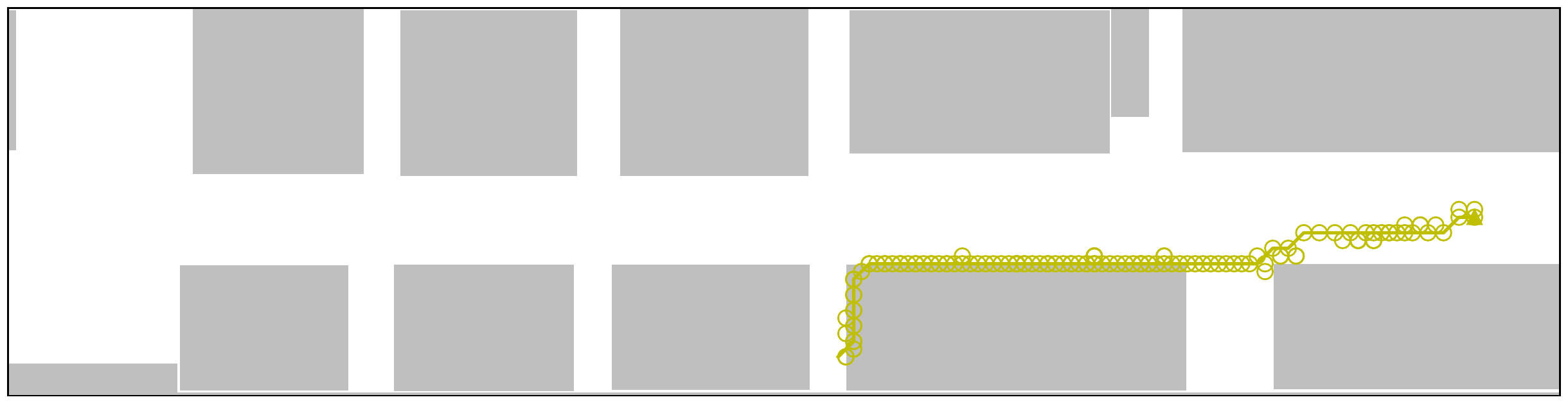}
        \caption{Baseline = 130}
        \label{fig:R_domain_baseline_b}
    \end{subfigure}
    \begin{subfigure}[b]{0.24\textwidth}
        \includegraphics[width=\textwidth]{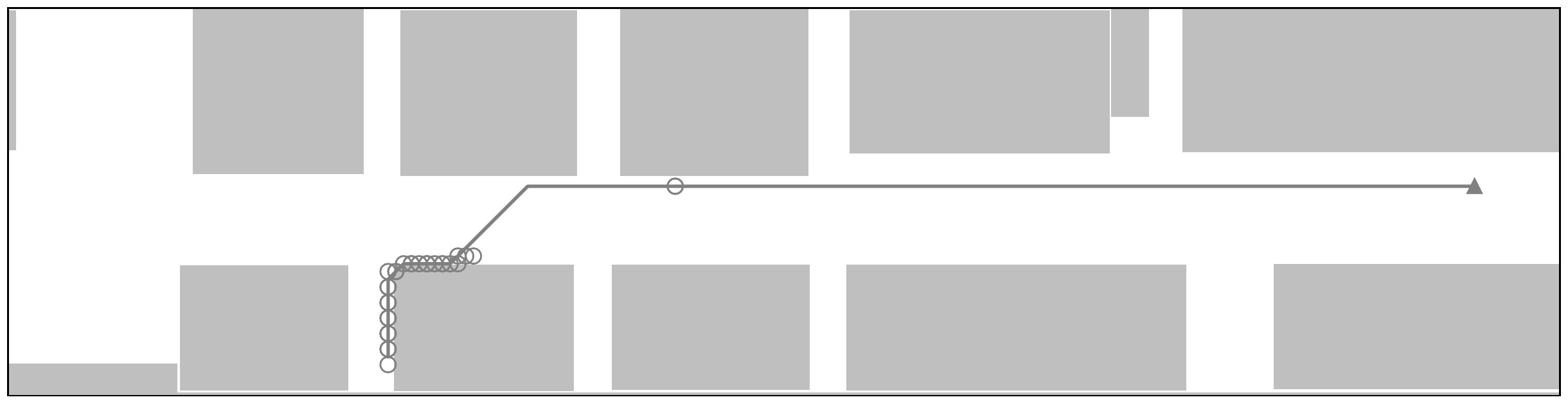}
        \caption{Baseline = 30}
        \label{fig:R_domain_baseline_c}
    \end{subfigure}
    \begin{subfigure}[b]{0.24\textwidth}
        \includegraphics[width=\textwidth]{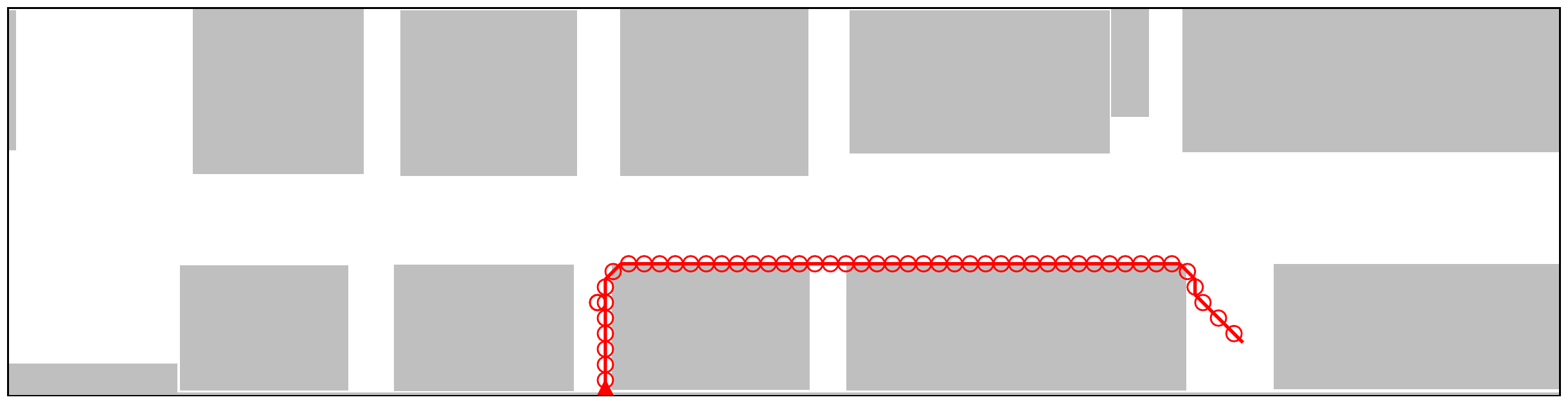}
        \caption{Baseline = 51}
        \label{fig:R_domain_baseline_d}
    \end{subfigure}
    \\
    \begin{subfigure}[b]{0.24\textwidth}
        \includegraphics[width=\textwidth]{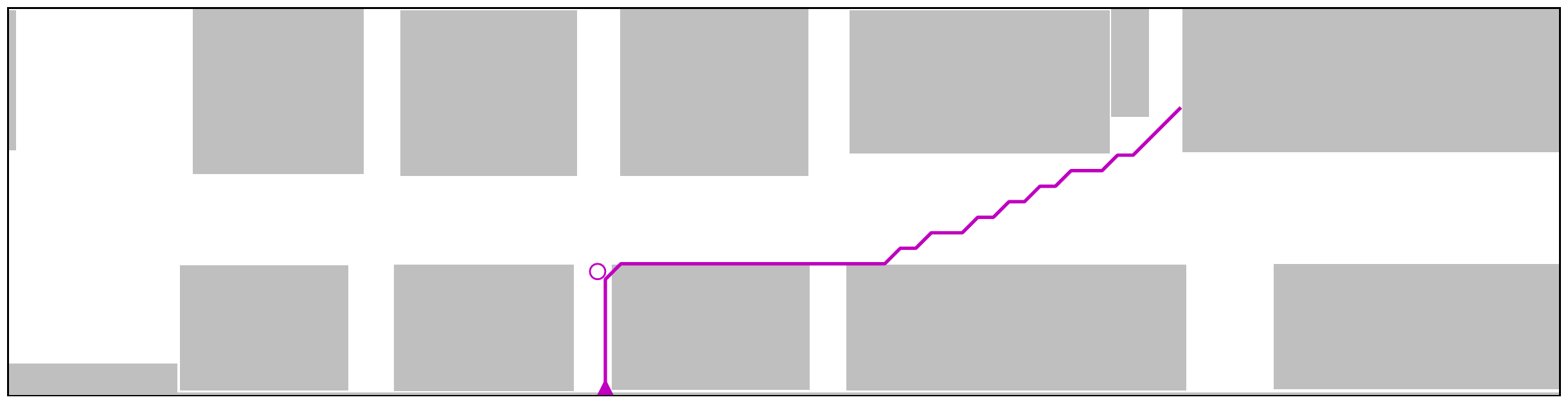}
        \caption{Baseline = 1}
        \label{fig:R_domain_baseline_e}
    \end{subfigure}
    \begin{subfigure}[b]{0.24\textwidth}
        \includegraphics[width=\textwidth]{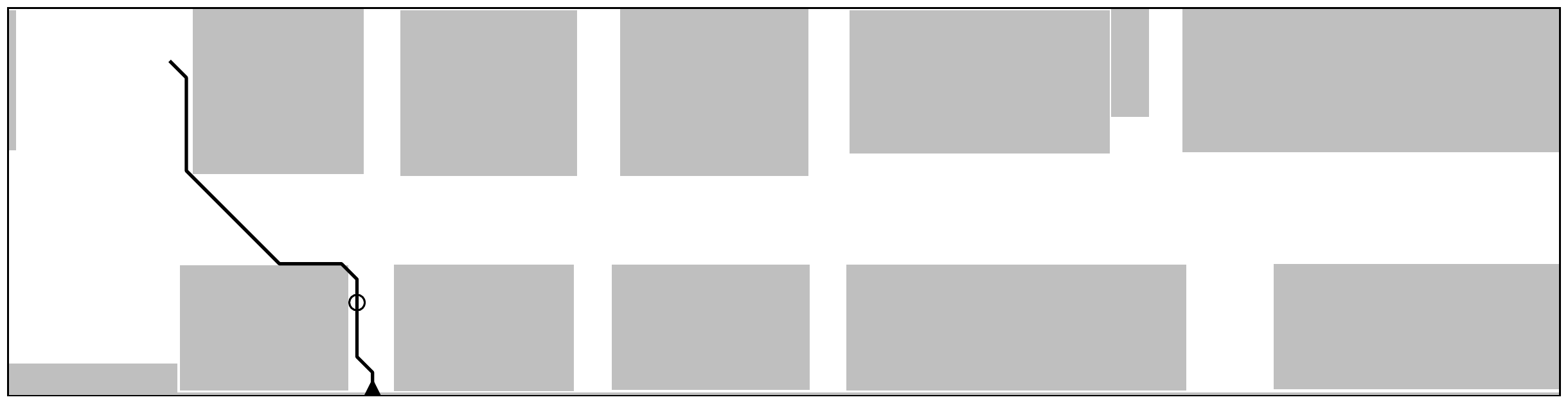}
        \caption{Baseline = 2}
        \label{fig:R_domain_baseline_f}
    \end{subfigure}
    \begin{subfigure}[b]{0.24\textwidth}
        \includegraphics[width=\textwidth]{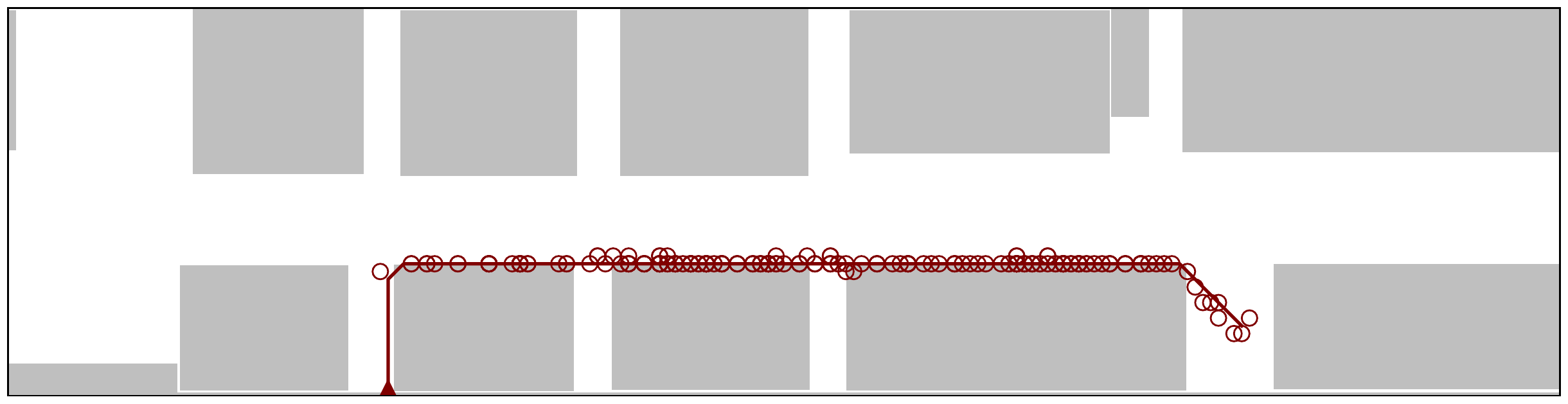}
        \caption{Baseline = 210}
        \label{fig:R_domain_baseline_g}
    \end{subfigure}
    \begin{subfigure}[b]{0.24\textwidth}
        \includegraphics[width=\textwidth]{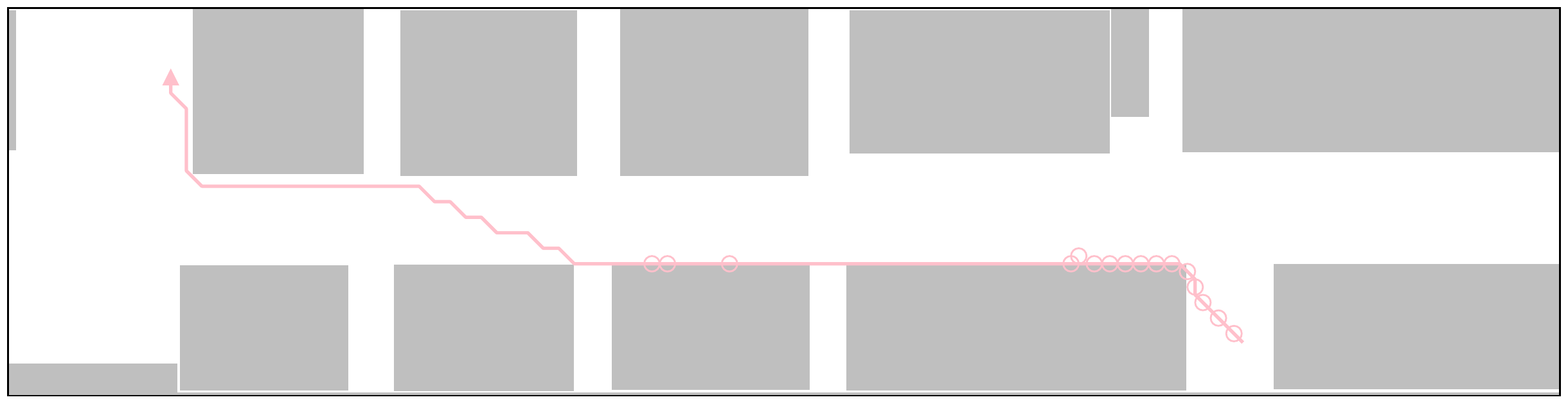}
        \caption{Baseline = 16}
        \label{fig:R_domain_baseline_h}
    \end{subfigure}
\caption{The planned paths and detected interactions according to the baseline for eight agents in a scenario in Stanford real domain. A subplot shows the planned path for one task, ending with a triangle. The small circles along the path denote the detected interactions with other tasks (agents) within the same scenario. The color of a path represents the identity of the agent. Gray areas represent the obstacles. The interaction difficulty based on the baseline is given in a subplot. Zooming in is recommended for a better view.}
\label{fig:R_domain_baseline}
\end{figure*}

The baseline result obtained by using the same experimental setup is also provided in \autoref{tab:approximated_sg_on_target_domains}. Corresponding to the same tasks in \autoref{fig:R_domain_measurement}, the eight planned paths and the detected interactions based on the baseline are depicted in \autoref{fig:R_domain_baseline_a}-\autoref{fig:R_domain_baseline_h} for comparison. Apparently, the baseline results present a large variance and cannot reveal the contributing factors nor their superpositions. 

\section{Verify Estimation of Scenario Generalization} \label{sec:Verify_Estimation_of_Scenario_Generalization}
We verify the consistency between the ranking in true scenario generalizations and that in estimated scenario generalizations. Based on the estimations, we further discuss the selection among candidate domains. 

\subsection{Ground truth Scenario Generalization Ranking} \label{sec:Groundtruth_Scenario_Generalization_Ranking}
The ground truth ranking for scenario generalizations of a model can be (empirically) determined by actually training that model on source, evaluating it on target w.r.t.\@ the desired performance metric and then ranking it across sources/targets on the same metric. In this study, the performance metric is the number of agent--agent collisions along model trajectories, averaged over agents and scenarios. 

The input of a learning model, simulated for each agent at each step, includes a range map (a distance map from the center of the ego agent to the surfaces of its surrounding obstacles and other agents) and a local guidance velocity toward the farthest visible waypoint or the destination (the farthest one that is not occluded along the line of sight of the ego agent by obstacles or other agents), both of which are in the local coordinate system of the agent (each agent is self-centered), and the output of the model is the velocity for the next step of the agent in its local coordinate system. In this work, we consider two kinds of input range map. The first kind has a resolution of 360 covering a 360-degree field of view, which we refer to as the dense feature. The second kind of range map has a resolution of 9 that covers a 90-degree field of view centered at the current heading direction of an agent, which we refer to as the sparse feature. 

We apply three representative training paradigms (BC \cite{long2017deep}, GAIL \cite{ho2016generative}, and PPO \cite{schulman2017proximal}) to train models on \texttt{ExSD} or \texttt{EgRD} source and test the models on the four targets \texttt{ExSD}, \texttt{EgRD}, \texttt{SRD}, and \texttt{SDD}. BC, GAIL, and PPO epitomize three distinct families of models. BC and GAIL are studied in \cite{qiao2019scenario}. They are imitation models that require demonstrated trajectories. In this work, we further cover the reinforcement learning paradigm, realized by proximal policy optimization (PPO) that does not require explicit demonstrations and, instead, gradually learns (by reinforcement) to optimize the local steering policy for solving the task (reach the goal from initial position). These paradigms feature in the following: (i) BC focuses on imitating simple reactive behaviors, encompassing classic regressors like support vector regressor, Random Forest, etc. (ii) GAIL considers the impact of local actions on accumulated outcomes. It does not require a preset reward function, which it learns adversarially. (iii) PPO, a SOTA on-policy RL approach with a proximal policy optimization uses gradient optimization in place of second-order methods, is penalized with the KL divergence between the new policy and the old policy, leading to \begin{equation} \label{equ:PPO_reward_function}
    \text{reward} = w_{l} \cdot r_{l} \cdot p_{t} + r_{d} + p_{c},
\end{equation}
where $w_{l}$ is a weight factor, which is empirically set to be $3$. $r_{l}$ is a location-based reward taking into account the distance to the destination. The smaller the distance, the higher $r_{l}$ is. $p_{t}$ is a predictive collision reward based on time-to-collision: the longer time-to-collision is, the higher this reward is. $r_{d}$, empirically set to be $2$, is an instantaneous reward triggered only when the agent reaches the destination. $p_{c}$, empirically set as $-0.01$, is an instantaneous reward for actual collisions with neighboring agents or with obstacles. 

The true costs are shown in \autoref{tab:integrated_result}, for both imitation models (BC \& GAIL) and reinforcement learning models (PPO). A lower true cost indicates a higher true scenario generalization. According to the table, in terms of the source domain, a model trained on domain \texttt{EgRD} has a higher true scenario generalization than one trained on \texttt{ExSD}, given the two models adopt the same kind of input range map, and are trained with the same paradigm, and then tested on the same target domain. On the other hand, for the same trained model performing on different target domains, due to the variations in input features, source-target domain pairs, and training paradigms, some models yield true scenario generalization rankings, from highest to lowest, in $\texttt{EgRD} \succ \texttt{SDD} \succ \texttt{ExSD} \succ \texttt{SRD}$, while other models yield $\texttt{SDD} \succ \texttt{EgRD} \succ \texttt{ExSD} \succ \texttt{SRD}$, and their agent--agent collision metrics on target domains \texttt{EgRD} and \texttt{SDD} are close. Therefore, we treat either ranking as the ground truth ranking and denote them as $\texttt{EgRD} = \texttt{SDD} \succ \texttt{ExSD} \succ \texttt{SRD}$.  

The results of PPO models are consistent, while for imitation models, there are outliers underlined in \autoref{tab:integrated_result}. In addition, we observe the PPO paradigm generates the highest-quality trajectories, which present consistently successful behaviors in all target domains. On the other hand, both BC and GAIL's performance depends strongly on the model and hyper-parameter tuning; our exhaustive validation during training could not yield reasonable performance. BC and GAIL showed many failing behaviors, which may introduce outliers for them. The underlined outliers in \autoref{tab:integrated_result} may be attributed to the inferior behavior of imitation models, in comparison with PPO. 

\begin{table*}[h!]
\fontsize{8.5}{8.5} \selectfont
\centering
\caption{Comparison of estimated target errors on \texttt{ExSD}, \texttt{EgRD}, \texttt{SRD} and \texttt{SDD} target domains for a trained model, which is averaged over scenarios and agents. The number in the parentheses indicates the ranking among target domains \texttt{ExSD}, \texttt{EgRD}, \texttt{SRD}, and \texttt{SDD}. A lower rank implies a higher scenario generalization.}
\label{tab:approximated_sg_on_target_domains}
\begin{tabular}{crrrr} 
\toprule
& Test on \texttt{ExSD} & Test on \texttt{EgRD}  & Test on \texttt{SRD}    & Test on \texttt{SDD}  \\ \midrule
Baseline & 31.92 $\pm$ 27.99 (2)  & 2.78 $\pm$ 1.03 (1)  & 89.78 $\pm$ 20.79 (3)   & 904.50 $\pm$ 415.84 (4) \\
IS (Ours)  & 4.78 $\pm$ 1.53 (3)   & 1.56 $\pm$ 0.45 (1)    & 7.60 $\pm$ 0.31 (4)    & 2.50 $\pm$ 0.66 (2)  \\ 
\bottomrule
\end{tabular}
\end{table*} 

\begin{sidewaystable}
\sidewaystablefn
\scriptsize
\caption{True target errors (agent--agent collisions) and estimated target errors, averaged over scenarios and agents, with standard deviation. The true target errors are directly measured for models trained with the BC, GAIL, or PPO methods on either \texttt{ExSD} or \texttt{EgRD} source domain and tested on \texttt{ExSD}, \texttt{EgRD}, \texttt{SRD}, and \texttt{SDD} target domains, using either a dense range map or sparse range map as model input. The target errors are also estimated by combining IS or the baseline (BL) term and the DQ$(\mathbb{S})$ term. A number in red within parentheses indicates the source domain's ranking between \texttt{ExSD} and \texttt{EgRD} given the same target domain, with the same training method and the same kind of input range map (comparisons are in the vertical direction). A number in blue presented within parentheses indicates the target domain's ranking among \texttt{ExSD}, \texttt{EgRD}, \texttt{SRD}, and \texttt{SDD} under the same source domain and the same training method (comparisons are in the horizontal direction). A lower rank implies a higher scenario generalization. The underlined numbers are considered outliers in imitation models.}
\label{tab:integrated_result}
\centering
\begin{tabular}{p{0.60cm}rrrrr}
\toprule
\multicolumn{2}{c}{S / Method}  & Test on \texttt{ExSD} & Test on \texttt{EgRD} & Test on \texttt{SRD} & Test on \texttt{SDD}  \\ 
\midrule
\multirow{5}{*}{\texttt{ExSD}} & BC (dense)   & 1.87 $\pm$ 1.70 \textcolor{red}{(2)} \textcolor{blue}{(3)}    & 0.31 $\pm$ 0.16 \textcolor{red}{(2)} \textcolor{blue}{(1)}   & 5.52 $\pm$ 1.72 \textcolor{red}{\underline{(1)}} \textcolor{blue}{(4)}  & 0.32 $\pm$ 0.10 \textcolor{red}{(2)} \textcolor{blue}{(2)} \\

                  & BC (sparse)   & 1.16 $\pm$ 0.87 \textcolor{red}{(2)} \textcolor{blue}{(3)}    & 0.33 $\pm$ 0.16 \textcolor{red}{(2)} \textcolor{blue}{(2)}   & 10.08 $\pm$ 5.34 \textcolor{red}{(2)} \textcolor{blue}{(4)}  & 0.27 $\pm$ 0.12 \textcolor{red}{(2)} \textcolor{blue}{(1)} \\

                  & GAIL (dense)     & 3.43 $\pm$ 2.81 \textcolor{red}{(2)} \textcolor{blue}{(3)}    & 0.37 $\pm$ 0.18 \textcolor{red}{(2)} \textcolor{blue}{(2)}   & 4.45 $\pm$ 1.00 \textcolor{red}{\underline{(1)}} \textcolor{blue}{(4)}  & 0.24 $\pm$ 0.07 \textcolor{red}{(2)} \textcolor{blue}{(1)} \\

                  & GAIL (sparse)     & 2.99 $\pm$ 2.49 \textcolor{red}{(2)} \textcolor{blue}{(3)}    & 0.43 $\pm$ 0.21 \textcolor{red}{(2)} \textcolor{blue}{(1)}   & 11.37 $\pm$ 3.98 \textcolor{red}{(2)} \textcolor{blue}{(4)}  & 0.88 $\pm$ 0.64 \textcolor{red}{(2)} \textcolor{blue}{(2)} \\

                   & PPO (sparse)      & 5.08 $\pm$ 5.81 \textcolor{red}{(2)} \textcolor{blue}{(3)}   & 0.81 $\pm$ 0.41 \textcolor{red}{(2)} \textcolor{blue}{(1)}  & 19.23 $\pm$ 5.02 \textcolor{red}{(2)} \textcolor{blue}{(4)} & 1.36 $\pm$ 0.65 \textcolor{red}{(2)} \textcolor{blue}{(2)} \\

                   & PPO-TDA & 6.49 $\pm$ 0.18 \textcolor{red}{(2)} \textcolor{blue}{(2)}   & 6.49 $\pm$ 0.16 \textcolor{red}{(2)} \textcolor{blue}{(2)}   & 6.75 $\pm$ 0.41 \textcolor{red}{(2)} \textcolor{blue}{(4)}  & 6.43 $\pm$ 0.23 \textcolor{red}{(2)} \textcolor{blue}{(1)}  \\
                
                   & BLDQ & 31.23 $\pm$ 27.99 \textcolor{red}{(2)} \textcolor{blue}{(2)}   & 2.09 $\pm$ 1.03 \textcolor{red}{(2)} \textcolor{blue}{(1)}   & 89.09 $\pm$ 20.79 \textcolor{red}{(2)} \textcolor{blue}{(3)}  & 903.81 $\pm$ 415.84 \textcolor{red}{(2)} \textcolor{blue}{(4)}  \\

                   & ISDQ (ours)       & 4.09 $\pm$ 1.53 \textcolor{red}{(2)} \textcolor{blue}{(3)}   & 0.87 $\pm$ 0.45 \textcolor{red}{(2)} \textcolor{blue}{(1)}   & 6.59 $\pm$ 0.31 \textcolor{red}{(2)} \textcolor{blue}{(4)}    & 1.79 $\pm$ 0.66 \textcolor{red}{(2)} \textcolor{blue}{(2)} \\
\midrule
\multirow{5}{*}{\texttt{EgRD}} & BC (dense)       & 0.16 $\pm$ 0.13 \textcolor{red}{(1)} \textcolor{blue}{\underline{(1)}}    & 0.25 $\pm$ 0.14 \textcolor{red}{(1)} \textcolor{blue}{\underline{(3)}}   & 8.55 $\pm$ 3.66 \textcolor{red}{\underline{(2)}} \textcolor{blue}{(4)}  & 0.22 $\pm$ 0.09 \textcolor{red}{(1)} \textcolor{blue}{(2)}  \\

                    & BC (sparse)    & 1.11 $\pm$ 1.32 \textcolor{red}{(1)} \textcolor{blue}{(3)}    & 0.30 $\pm$ 0.15 \textcolor{red}{(1)} \textcolor{blue}{(2)}   & 2.68 $\pm$ 0.84 \textcolor{red}{(1)} \textcolor{blue}{(4)}  & 0.18 $\pm$ 0.07 \textcolor{red}{(1)} \textcolor{blue}{(1)}  \\

                   & GAIL (dense)    & 0.51 $\pm$ 0.31 \textcolor{red}{(1)} \textcolor{blue}{(3)}    & 0.12 $\pm$ 0.08 \textcolor{red}{(1)} \textcolor{blue}{(1)}   & 6.33 $\pm$ 1.87 \textcolor{red}{\underline{(2)}} \textcolor{blue}{(4)}  & 0.23 $\pm$ 0.08 \textcolor{red}{(1)} \textcolor{blue}{(2)}  \\

                   & GAIL (sparse)    & 0.68 $\pm$ 0.56 \textcolor{red}{(1)} \textcolor{blue}{(3)}    & 0.30 $\pm$ 0.16 \textcolor{red}{(1)} \textcolor{blue}{(2)}   & 1.80 $\pm$ 0.65 \textcolor{red}{(1)} \textcolor{blue}{(4)}  & 0.23 $\pm$ 0.14 \textcolor{red}{(1)} \textcolor{blue}{(1)}  \\

                   & PPO (sparse)     & 3.39 $\pm$ 3.08 \textcolor{red}{(1)} \textcolor{blue}{(3)}   & 0.54 $\pm$ 0.30 \textcolor{red}{(1)} \textcolor{blue}{(1)}  & 15.34 $\pm$ 5.63 \textcolor{red}{(1)} \textcolor{blue}{(4)} & 0.71 $\pm$ 0.28 \textcolor{red}{(1)} \textcolor{blue}{(2)} \\

                   & PPO-TDA & 0.49 $\pm$ 0.03 \textcolor{red}{(1)} \textcolor{blue}{(4)}   & 0.47 $\pm$ 0.02 \textcolor{red}{(1)} \textcolor{blue}{(1)}   & 0.47 $\pm$ 0.02 \textcolor{red}{(1)} \textcolor{blue}{(1)}  & 0.47 $\pm$ 0.03 \textcolor{red}{(1)} \textcolor{blue}{(1)}  \\
                    
                   & BLDQ & 30.80 $\pm$ 27.99 \textcolor{red}{(1)} \textcolor{blue}{(2)}   & 1.66 $\pm$ 1.03 \textcolor{red}{(1)} \textcolor{blue}{(1)}   & 88.65 $\pm$ 20.79 \textcolor{red}{(1)} \textcolor{blue}{(3)}  & 903.38 $\pm$ 415.84 \textcolor{red}{(1)} \textcolor{blue}{(4)} \\

                   & ISDQ (ours)      & 3.66 $\pm$ 1.53 \textcolor{red}{(1)} \textcolor{blue}{(3)}    & 0.44 $\pm$ 0.45 \textcolor{red}{(1)} \textcolor{blue}{(1)}   & 6.16 $\pm$ 0.31 \textcolor{red}{(1)} \textcolor{blue}{(4)}   & 1.36 $\pm$ 0.66 \textcolor{red}{(1)} \textcolor{blue}{(2)} \\
\bottomrule
\end{tabular}
\end{sidewaystable} 

\begin{table}[h!]
\scriptsize
\centering
\caption{DQ$(\mathbb{S})$ terms of \texttt{ExSD} and \texttt{EgRD}. A number within parentheses indicates the rank of the source domain. A lower rank number implies a higher scenario generalization.}
\label{tab:diversity_of_domains}
\begin{tabular}{ccccc} 
\toprule
\multicolumn{1}{l}{} & \multicolumn{1}{l}{H($\mathcal{E})$} & \multicolumn{1}{l}{H(I, D \textbar{} $\mathcal{E}$)} & \multicolumn{1}{l}{H(I, D, $\mathcal{E})$} & \multicolumn{1}{l}{DQ$(\mathbb{S})$}\\ 
\midrule
\texttt{ExSD}       & 2.59           & 4.32             & 6.91     & -6.91 (2)  \\ 
\texttt{EgRD}       & 6.64           & 4.57             & 11.22    & -11.22 (1)  \\ 
\bottomrule
\end{tabular}
\end{table} 

\subsection{Estimated Scenario Generalization} \label{sec:Estimated_Scenario_Generalization}
Scenario generalizations of learning models are estimated according to Eq.\eqref{equ:approximate_SG}. First, DQ$(\mathbb{S})$ (the second term in Eq.\eqref{equ:approximate_SG}) on source \texttt{ExSD} and source \texttt{EgRD} are quantified with the method described in Section \ref{sec:Diversity_Quantification}. The results are shown in \autoref{tab:diversity_of_domains}, indicating that \texttt{EgRD} is quantified with a higher diversity than \texttt{ExSD}, which is in accordance with the subjective assessment. Second, the IS measurements (the first term in Eq.\eqref{equ:approximate_SG}) on target domains \texttt{ExSD}, \texttt{EgRD}, \texttt{SRD}, and \texttt{SDD} are computed. The results are provided in \autoref{tab:approximated_sg_on_target_domains}, compared with the baseline. Third, we combine the two terms with $\lambda=0.1$ to obtain the estimated target errors for various pairs of source domains and target domains. The final results are incorporated in \autoref{tab:integrated_result}. It shows that the ISDQ ranks the estimated scenario generalizations of a trained model on target domains, from highest to lowest: $ \texttt{EgRD} \succ \texttt{SDD} \succ \texttt{ExSD} \succ \texttt{SRD}$, consistent with the ranking in true scenario generalizations. 

For comparison, we substitute the IS term in Eq.\eqref{equ:approximate_SG} with the baseline quantity (BL) to obtain the \textbf{BLDQ} estimated target errors. \autoref{tab:integrated_result} shows that BLDQ yields a ranking over target domains, from highest to lowest: $\texttt{EgRD} \succ \texttt{ExSD} \succ \texttt{SRD} \succ \texttt{SDD}$, inconsistent with the ground true ranking. 

For another comparative scenario generalization estimator PPO-TDA, in terms of the absolute predicted target error, we can see that there is a large discrepancy between the true target error (PPO) and the predicted target error (PPO-TDA), trained under the same source domain. Regarding the ranking of the predicted scenario generalizations on target domains, from highest to lowest, the PPO-TDA trained on \texttt{ExSD} yields $\texttt{SDD} \succ \texttt{ExSD} = \texttt{EgRD} \succ \texttt{SRD}$, while the PPO-TDA trained on \texttt{EgRD} yields $\texttt{EgRD} = \texttt{SRD} = \texttt{SDD} \succ \texttt{ExSD}$, neither of which accords with the ground truth ranking. 

\subsection{Selecting Candidate Domain} \label{sec:Selecting_Candidate_Scenario_Domains}
In the case of a single target domain with multiple source domains, the goal is to select the source domain that has the smallest DQ$(\mathbb{S})$ term. Results in \autoref{tab:integrated_result} and \autoref{tab:diversity_of_domains} reveal that on the same target domain, \texttt{EgRD} domain has a lower DQ$(\mathbb{S})$ and hence a higher estimated scenario generalization than \texttt{ExSD} domain when served as a source domain. This estimated ranking accords with the ranking in the true scenario generalizations. On the other hand, in the case of a single source domain with multiple target domains, the goal is to select the target domain with the smallest IS term for an already trained model. Again, the IS measurement ranks the estimated scenario generalizations in accordance with the ranking in the true scenario generalizations. Therefore, one may exploit the diversity quantification (\autoref{tab:diversity_of_domains}) to choose the source domain and the IS measurement to select the target domain (\autoref{tab:approximated_sg_on_target_domains}) for training and testing a model. Subsequently, when there are multiple source domains and multiple target domains, the combined ISDQ measurement could indicate the appropriate source-target domain pair. 

\section{Further Discussion}  \label{sec:Further_Discussion}
\subsection{Alternative Internal Simulator}
We choose social force as the primary internal steering model in that it is computational efficient, easy to control via parameters, and more suitable than other models. For instance, Reynolds (Boids) simulates the flocking behaviour of birds. Two interaction rules in Boids agents require that the agents form a group and share roughly the same goal position. By contrast, in decentralized crowd systems, each agent has its own task. For RVO and ORCA \cite{van2011reciprocal}, once a feasible space (collision-free velocity space) is found, they choose the velocity closest to the input planned velocity, thus less capable of presenting various interactions among agents (recall the purpose of sampling $m$ points is to cover as many steering policies for all tasks in the scenario as possible). For more recent ones, they are generally more complex, and are even learning-based, which are harder to control and may not be appropriate as an internal steering model.

Although our primary choice is social force, here we conduct experiments using ORCA as an alternative internal steering model, based on which we measure the IS scores over the four domains. The ending parameters of ORCA for sampling trajectories are listed in \autoref{tab:param_configuration_orca}. Other algorithmic settings remain unchanged. 

\begin{table}[h]
\centering
\caption{The ending parameters $\theta^{1}$ and $\theta^{m}$ of the IS measurement for the internal ORCA simulator. All $m=100$ parameters are linearly interpolated between them (we reduce  $m$ from 300 to 100 to speedup sampling of trajectories).}
\begin{tabular}{ccc} 
\toprule
                                        &  $\theta^{1}$     &  $\theta^{m}$  \\
\midrule
neighbor distance (meter)               &  1.0              &  50.0 \\ 
maximum number of neighbors             &  1                &  50 \\ 
time horizon (second)                   &  0.1              &  10.0 \\
time horizon obstacle (second)          &  2.0              &  2.0  \\
\bottomrule
\label{tab:param_configuration_orca}
\end{tabular}
\end{table}

Our experimental results show that the ranking of IS scores over the four domains using ORCA accords with the ranking using social force, provided in \autoref{tab:approximated_sg_on_target_domains_2}. This suggests that the proposed framework is not sensitive to the choice of an internal steering model, in terms of estimating scenario generalizations of source-target domain pairs.

\begin{table*}[h!]
\fontsize{8.5}{8.5} \selectfont
\centering
\caption{Comparison of IS scores with different design choices in estimating task-level inter-agent interaction difficulties on \texttt{ExSD}, \texttt{EgRD}, \texttt{SRD} and \texttt{SDD} target domains, which is averaged over scenarios and agents. The number in the parentheses indicates the ranking among target domains \texttt{ExSD}, \texttt{EgRD}, \texttt{SRD}, and \texttt{SDD}.}
\label{tab:approximated_sg_on_target_domains_2}
\begin{tabular}{crrrr} 
\toprule
& Test on \texttt{ExSD}     & Test on \texttt{EgRD}    & Test on \texttt{SRD}    & Test on \texttt{SDD}  \\ \midrule
IS (SF-percentile)       & 4.78 $\pm$ 1.53 (3)     & 1.56 $\pm$ 0.45 (1)    & 7.60 $\pm$ 0.31 (4)    & 2.50 $\pm$ 0.66 (2)  \\ 
IS (ORCA-percentile)       & 5.06 $\pm$ 1.15 (3)     & 1.66 $\pm$ 0.52 (1)    & 5.92 $\pm$ 0.22 (4)    & 2.50 $\pm$ 0.66 (2)  \\
IS (SF-DBSCAN)  & 0.59 $\pm$ 0.92 (3)     & 0.03 $\pm$ 0.05 (1)    & 1.74 $\pm$ 0.33 (4)    & 0.14 $\pm$ 0.09 (2)  \\
\bottomrule
\end{tabular}
\end{table*}

\begin{figure}[h!]
\centering
\begin{tabular}{cc}
   \includegraphics[width=0.48\linewidth]{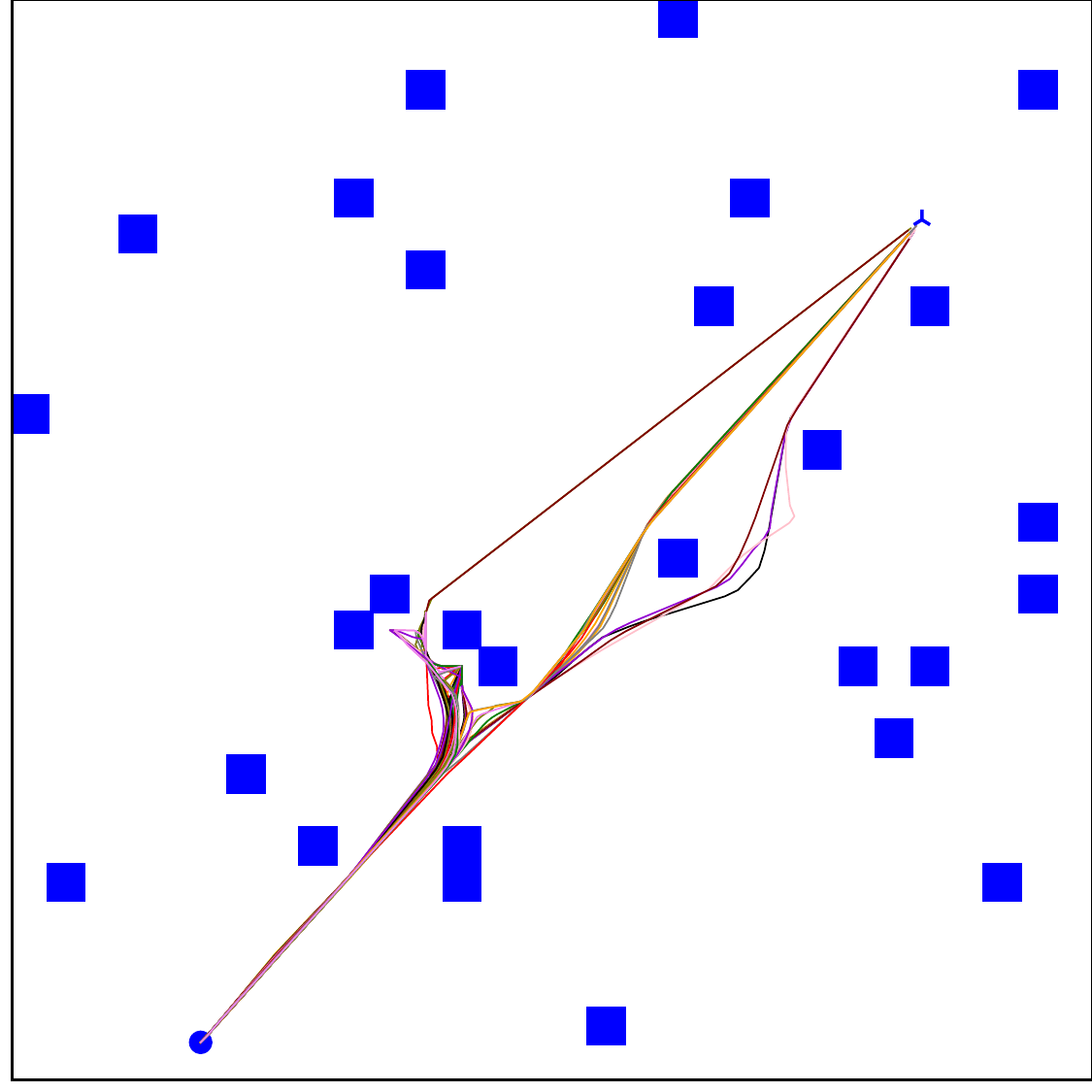}
 & \includegraphics[width=0.48\linewidth]{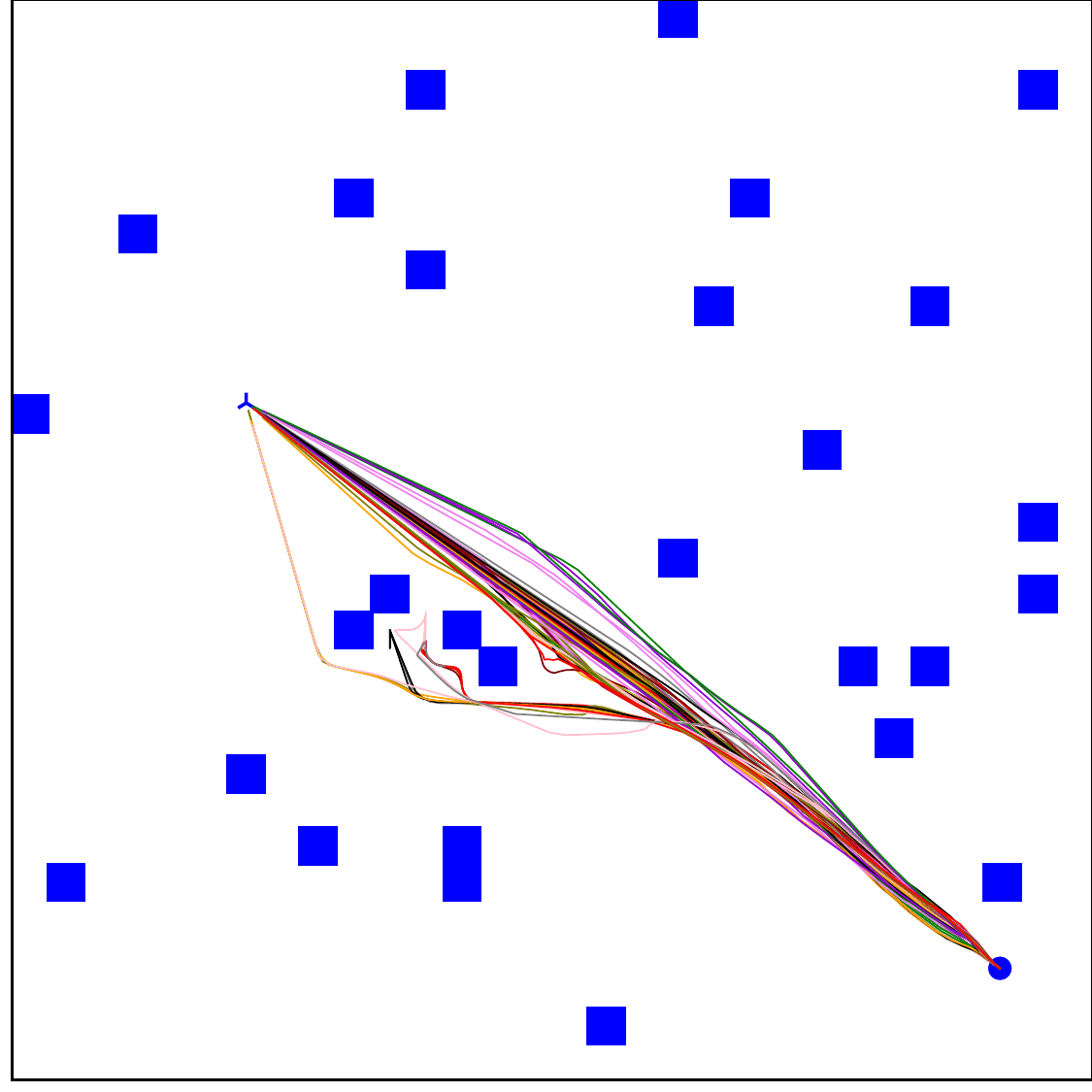}\\
 \includegraphics[width=0.48\linewidth]{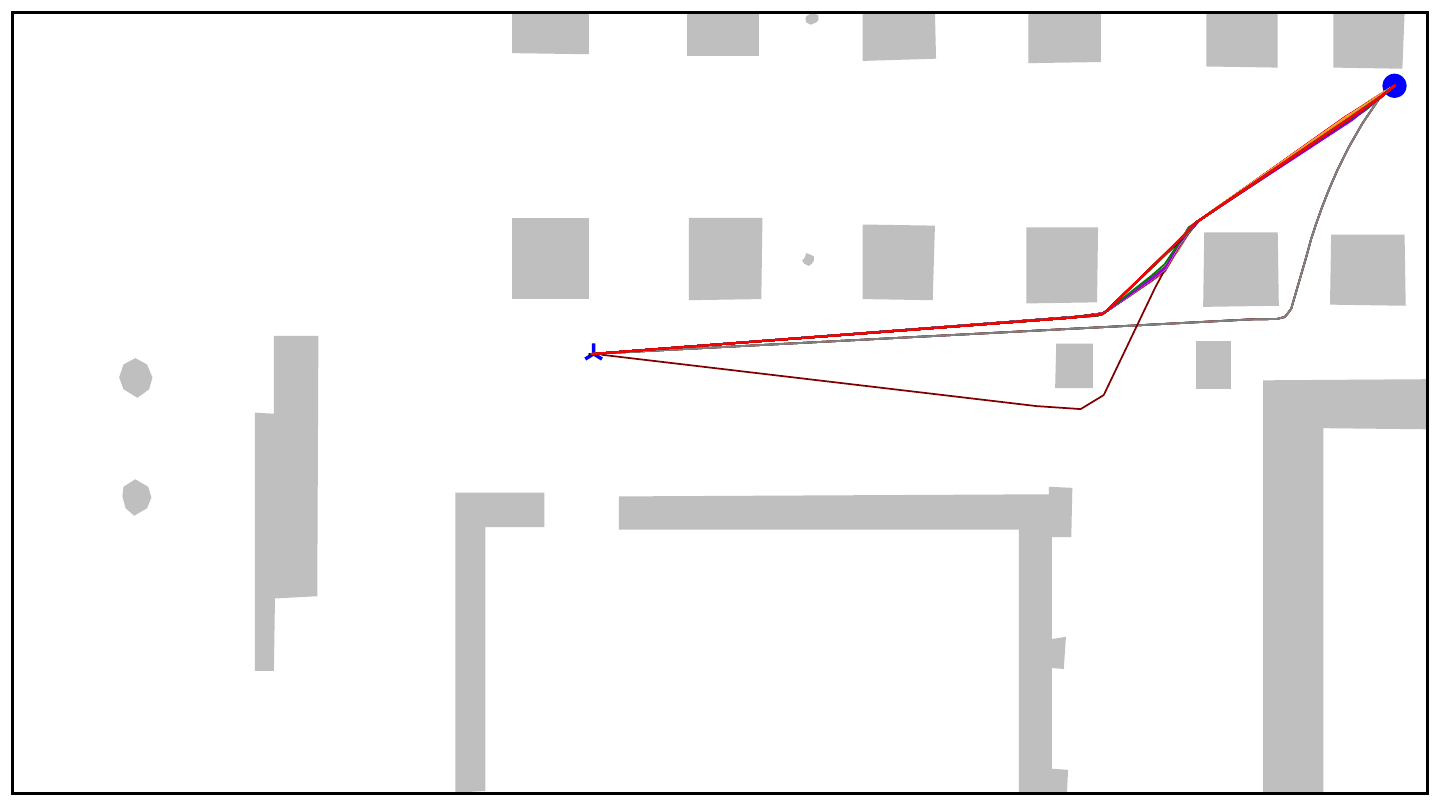}
 & \includegraphics[width=0.48\linewidth]{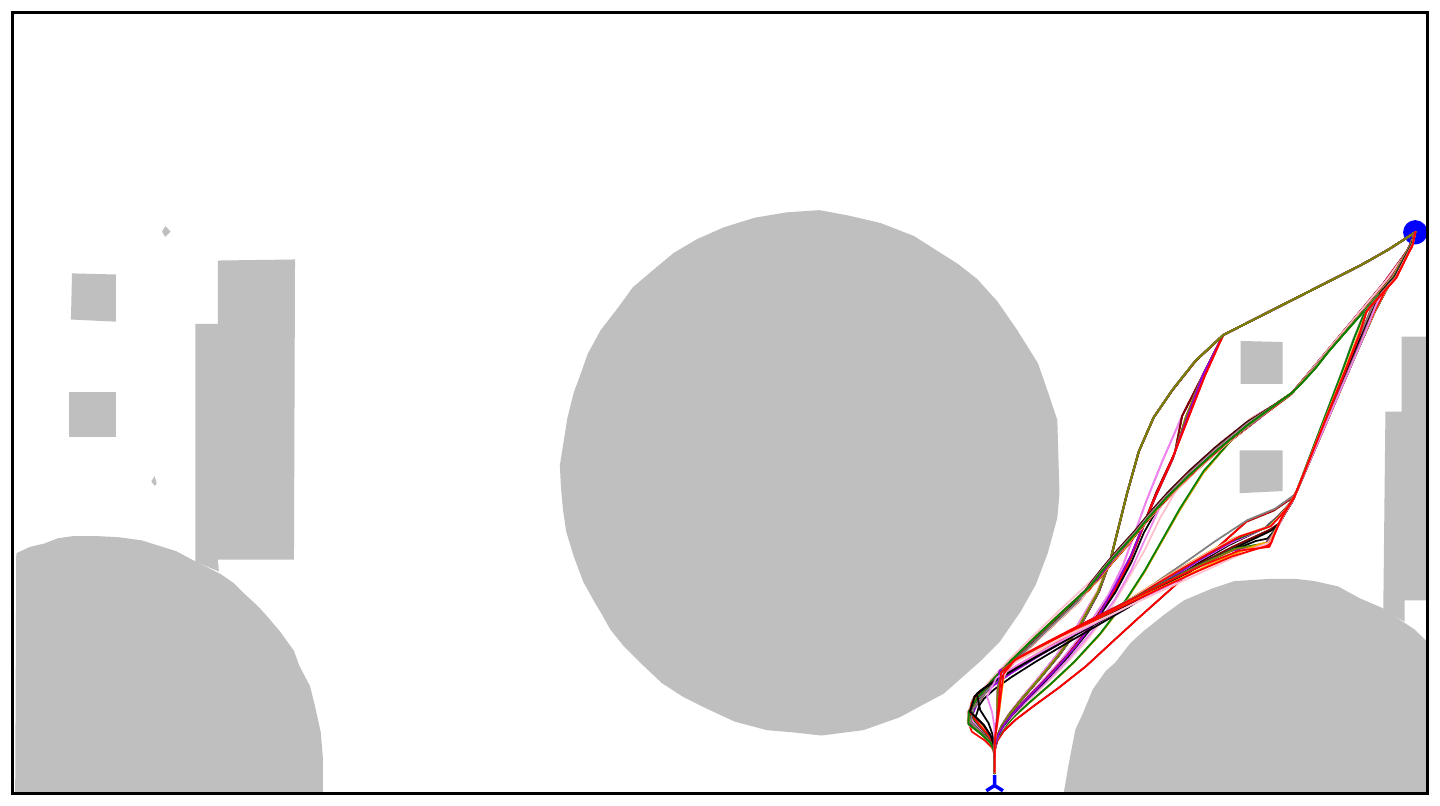}
 
\end{tabular}
\caption{Sampled interactive trajectories of an agent using ORCA from egocentric representative domain (\texttt{EgRD}) and Stanford drone domain (\texttt{SDD}). Each trajectory of the agent interacts with other agents during ORCA simulation.}
\label{fig:vis_ORCA}
\end{figure}

Some visualizations of the sampled interactive trajectories of an agent using ORCA are shown in \autoref{fig:vis_ORCA}.

\subsection{Alternative Trajectory Clustering Method}
In Section \ref{sec:The_Clustering_Procedure}, we suggest a clustering method to group interactive trajectories of an agent into modes, based on DTW labeling and percentile. This method is computational efficient and easy to implement. On the other hand, trajectory clustering is a well-studied area, with many methods specifically for crowds \cite{bian2018survey,wang2006learning,wang2016path,wang2008unsupervised,wang2016globally,emonet2011extracting,wang2016trending,he2020informative,zhou2015learning,wang2011trajectory}.

In order to verify that the proposed framework is robust to the choice of a trajectory clustering method, we conduct an alternative trajectory clustering method using DBSCAN (Density-Based Spatial Clustering of Applications with Noise) \cite{ester1996density}. DBSCAN, a general and widely used clustering algorithm, grows a cluster by adding core points until no more core points can be added, and then adding non-core points from core points until no more non-core points can be extended from core points. A point that can not be added to any cluster is treated as an outlier.

The direct application of DBSCAN to group interactive trajectories of an agent requires the computation of the distance between every pair of the interactive trajectories of the agent, which is computationally expensive (in that trajectories may have different lengths, requiring complex computation like DTW, which is not trivial). To speedup the computation, we still use DTW to label each trajectory, by computing the difference between that trajectory and the solo trajectory of the agent, and then apply DBSCAN to the 1-dimensional DTW data. In short, we keep DTW labeling but replace the percentile-based clustering method with DBSCAN based clustering method, to group DTW labeled trajectories.

We set the parameters of DBSCAN as $\epsilon=10.0$ (the maximum distance between two points for one to be considered as in the neighborhood of the other), $\text{min samples}=5$ (the number of points in a neighborhood for a point to be considered as a core point, including the point itself) across all four domains. The IS scores are shown in the last row of \autoref{tab:approximated_sg_on_target_domains_2}. We can see that although the absolute IS scores reduce dramatically compared with IS scores using the percentile clustering, the ranking of IS scores of the four domains remain unchanged. This suggests that the proposed framework is robust to the trajectory clustering method, in terms of estimating scenario generalizations of source-target domain pairs.

\begin{figure}[h!]
\centering
\begin{tabular}{cc}
   \includegraphics[width=0.48\linewidth]{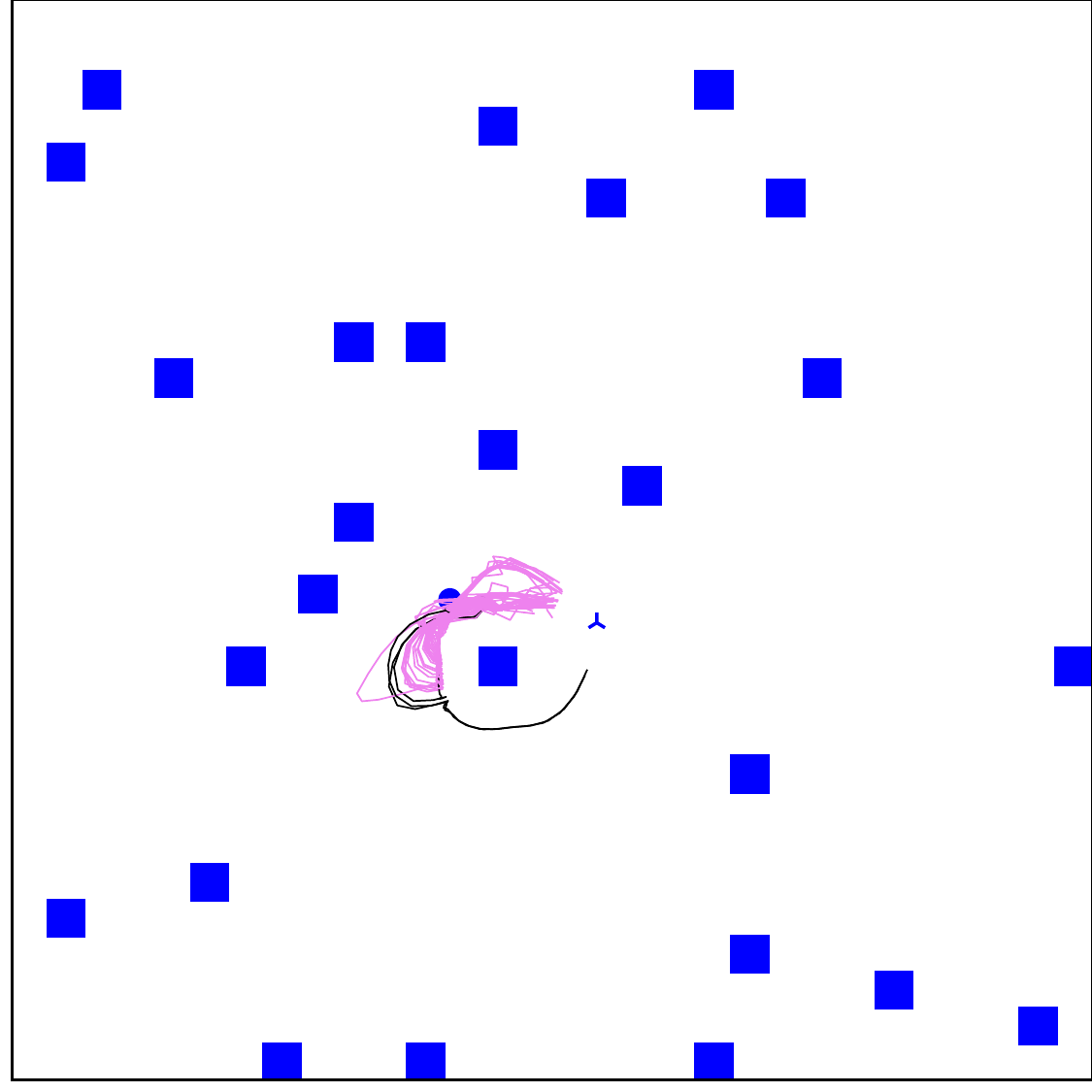}
 & \includegraphics[width=0.48\linewidth]{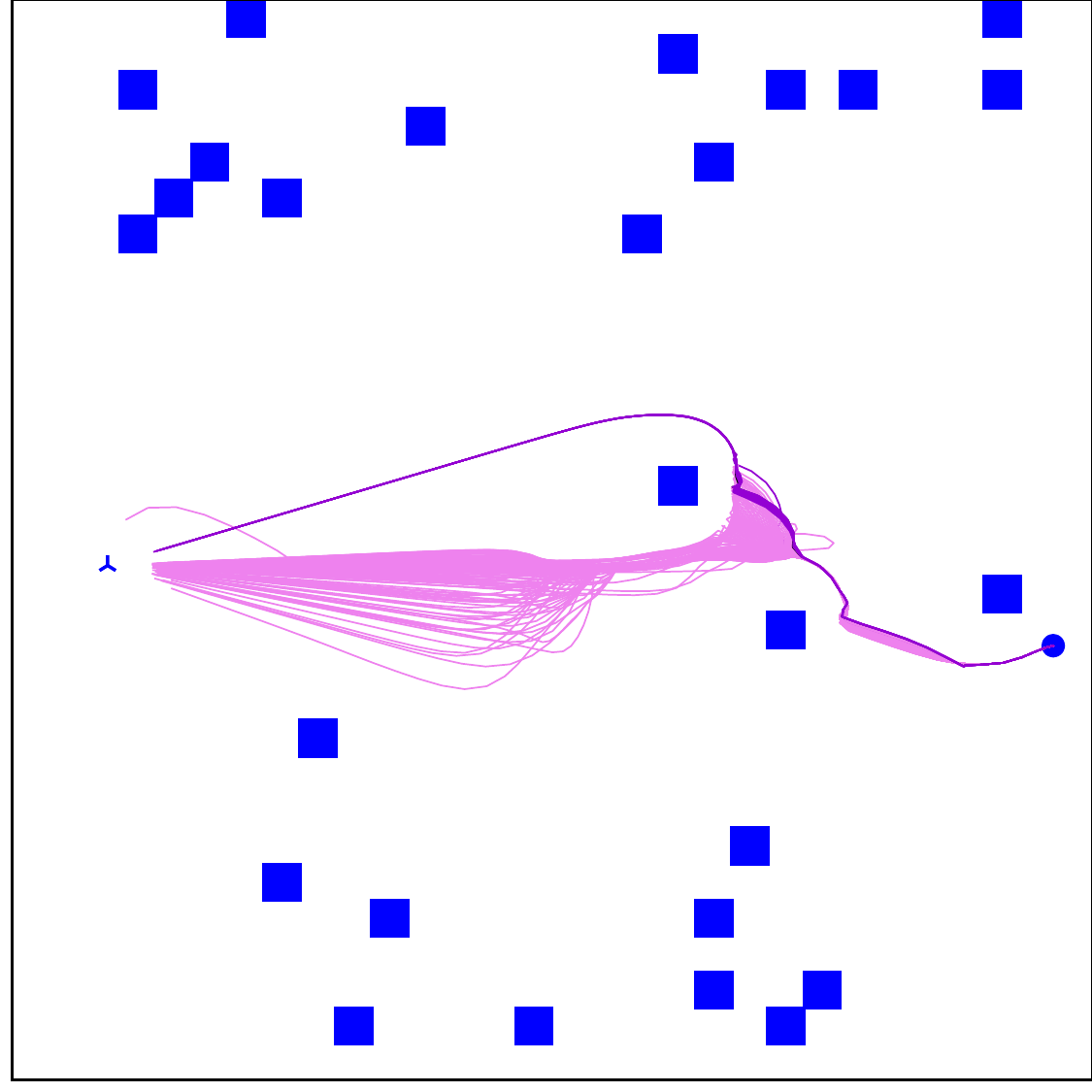}\\
   \includegraphics[width=0.48\linewidth]{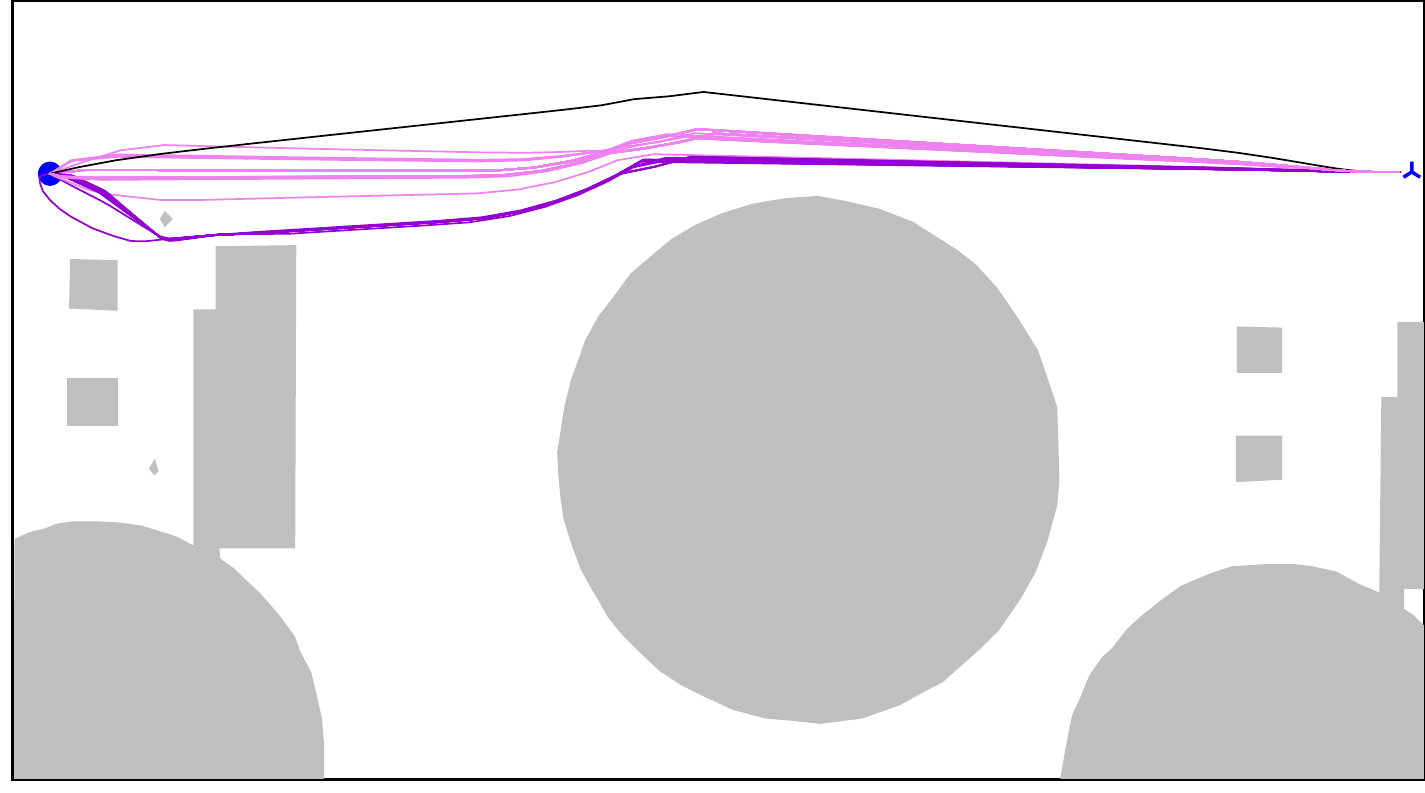}
 & \includegraphics[width=0.48\linewidth]{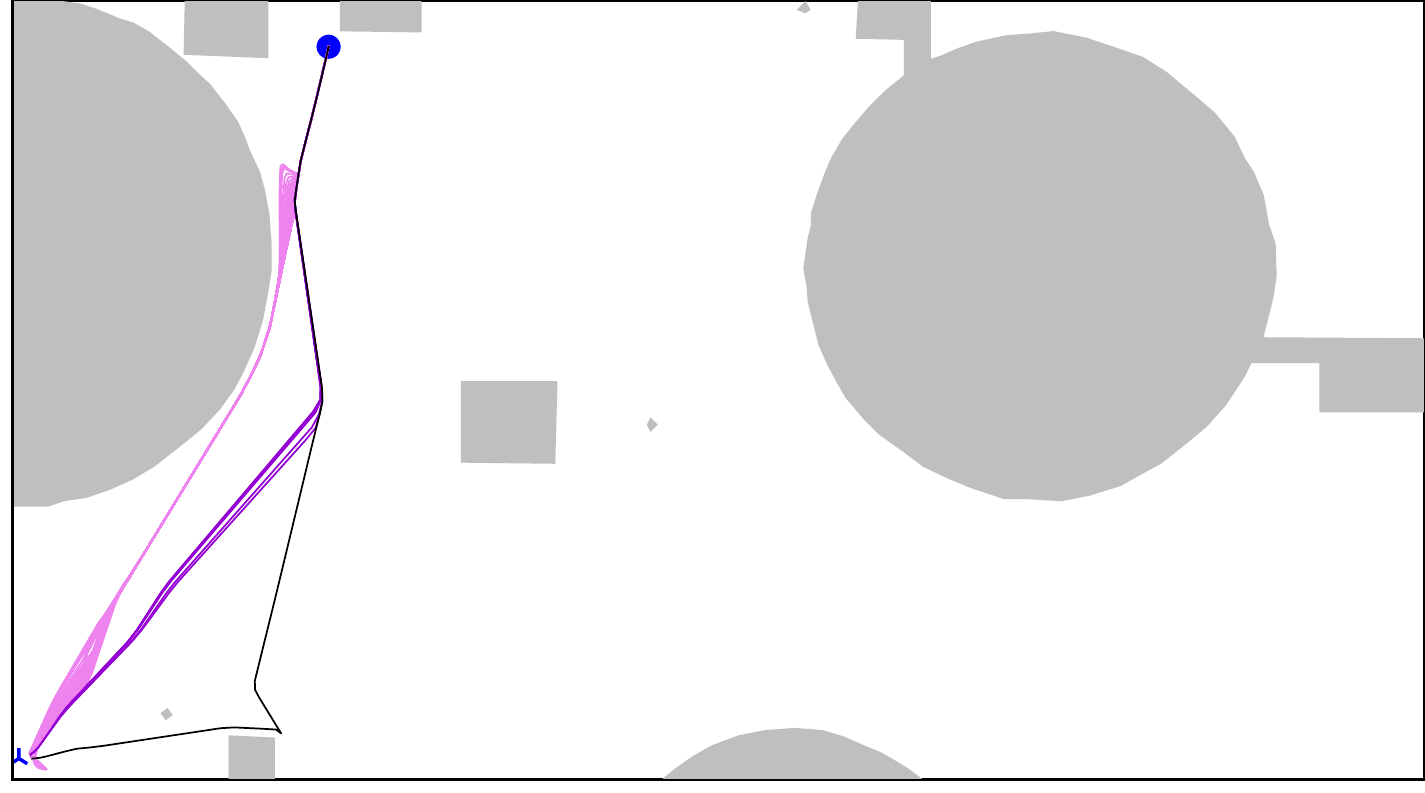}
\end{tabular}
\caption{Clustered interactive trajectories of an agent using DBSCAN in egocentric representative domain (\texttt{EgRD}) and Stanford drone domain (\texttt{SDD}). Each trajectory of the agent interacts with other agents during the social force simulation. Trajectories of the same color are in a group. The trajectories in black color are in the outlier group.}
\label{fig:vis_DBSCAN}
\end{figure}

Some visualizations of the grouped interactive trajectories of an agent using
DBSCAN are shown in \autoref{fig:vis_DBSCAN}.

\subsection{Influence of Obstacle Configuration}
\begin{figure}[h!]
\small
\begin{tabular}{c c c}
\hspace{-0.1cm}\includegraphics[width=0.32\linewidth]{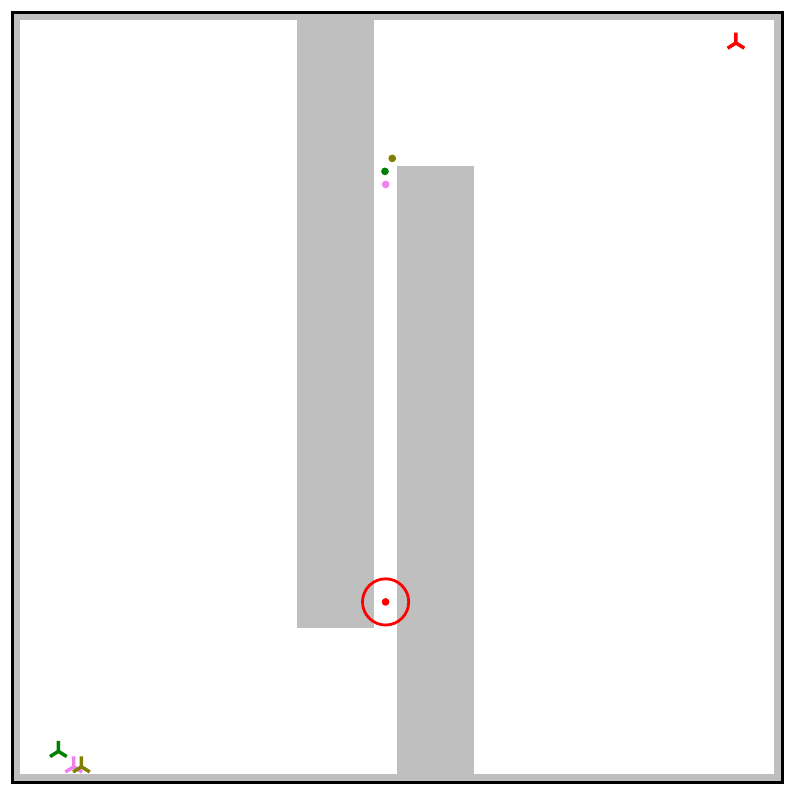}
\includegraphics[width=0.32\linewidth]{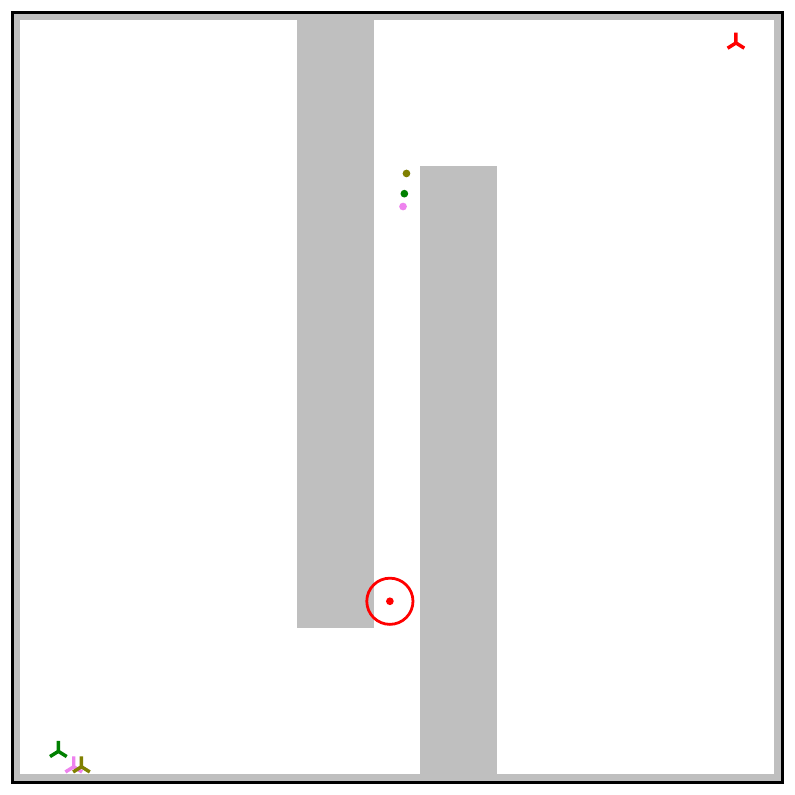}
\includegraphics[width=0.32\linewidth]{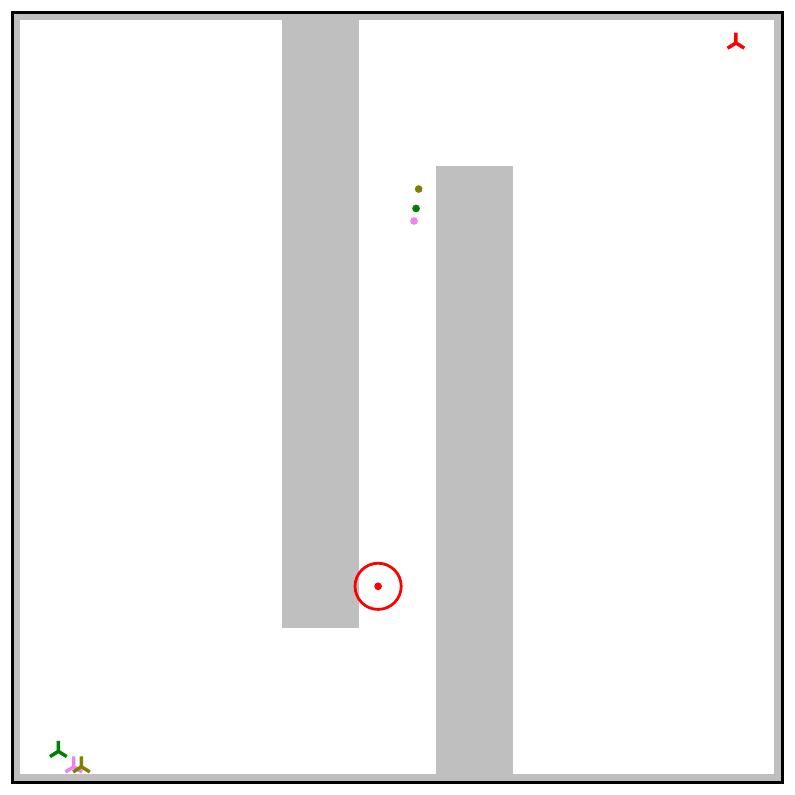}\\
\end{tabular}
\caption{Illustration of gradually enlarged horizontal distance between two obstacles. In the left figure, the distance is 3m, in the middle figure, the distance is 6m, in the right figure, the distance increases to 10m. Note that in each scenario, the tasks do not change: agent $X_{i}$ (the red circled one) starts from the bottom left and targets at the top right position, and agents $X_{-i}$ start from the top right and target at the bottom left positions.}
\label{fig:illustration_vary_env}
\end{figure}

\begin{figure}[h!]
\small
\begin{tabular}{c}
\hspace{-0.2cm}\includegraphics[width=0.98\linewidth]{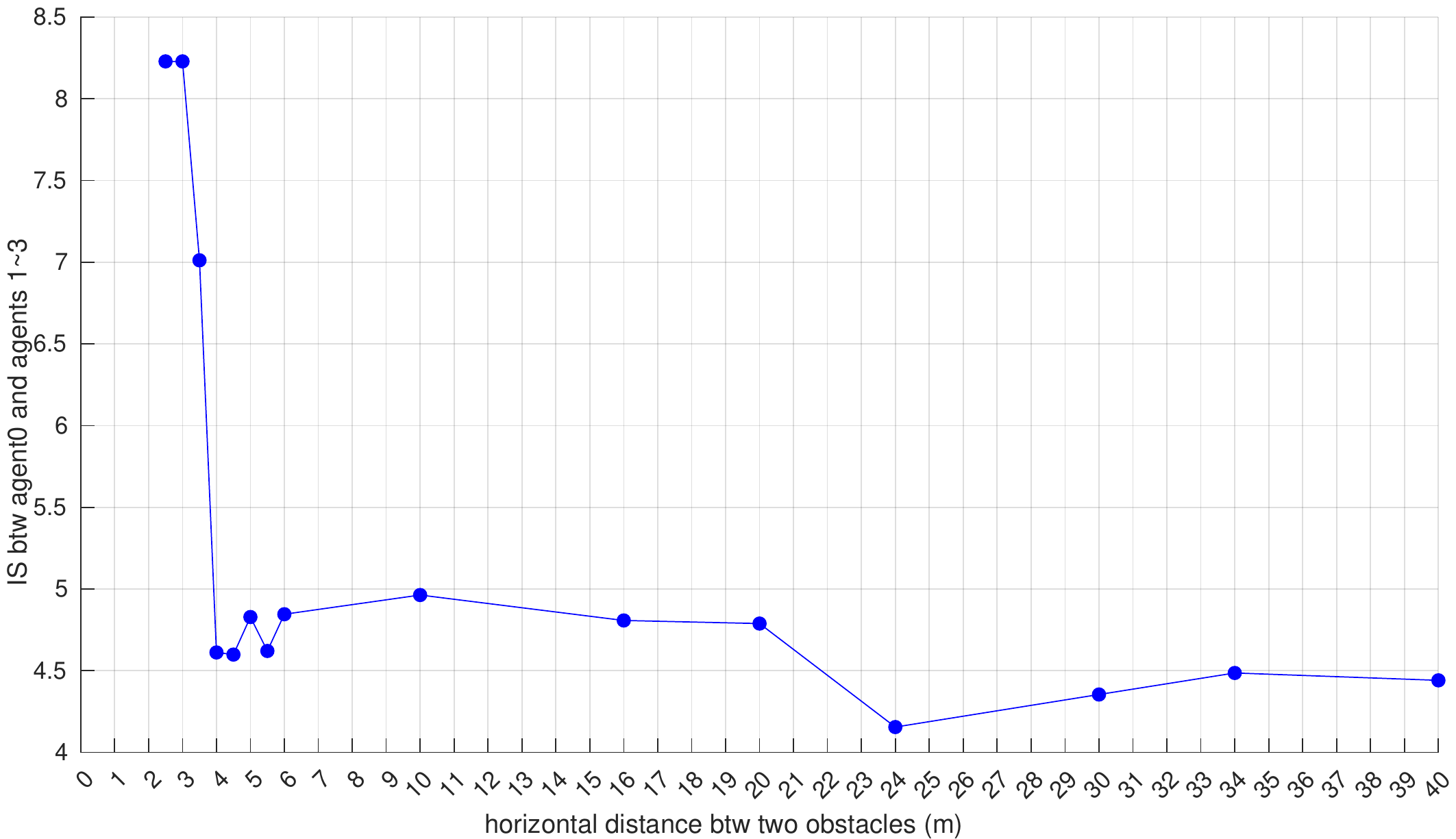}\\ 
\end{tabular}
\caption{IS between agent $X_{i}$ and the other three agents $X_{-i}$ presents a sharp decrease when the horizontal distance between the two obstacles increases from 3m to 4m. Further enlargement of the horizontal distance between the two obstacles does not reduce IS significantly.}
\label{fig:measurement_vary_env}
\end{figure}

In IS, the obstacle configuration is an essential hidden factor on which IS depends, in that obstacle configuration exerts geometric and kinematics constraints on the interactions among agents. For example, agents moving in a small area may need more efforts to avoid each other, than that in a spacious area. To verify the influence of obstacle configuration, we designed a parameterized scenario where obstacle configuration varies while tasks are fixed. \autoref{fig:illustration_vary_env} illustrates how the horizontal distance between two obstacles is gradually enlarged (note that across all scenarios, neither the number of agents nor their tasks change). \autoref{fig:measurement_vary_env} depicts IS w.r.t. the horizontal distance between the two obstacles. The IS decreases dramatically when the distance increases from 3m to 4m. When the distance is further enlarged ($>$4m), the distance becomes less influential on IS, which fluctuates between 4.0 and 5.0. This accords with the observation that agents move diagonally within the two obstacles, and further enlargement of the distance does not exert a sustained impact on the interactions.

\subsection{Influence of Start and Destination Diversities}
Variations in the obstacle configuration $\mathcal{E}$ and tasks $X_{1 \sim n}$ result in the diversity of scenarios within a domain. Diverse source scenarios provide a higher coverage for a crowd learning model to collect feedback and adapt. Our experiments by comparing \text{ExSD} and \text{EgRD} as source domains provide an empirical evidence for the significance of the diversity of a source domain in affecting scenario generalization. According to the red number in the parentheses in \autoref{tab:integrated_result}, in terms of the source domain, a model trained on domain \text{EgRD} has a higher scenario generalization than the one trained on \text{ExSD}, given the two models adopt the same kind of input range map, and are trained with the same paradigm, and then tested on the same target domain, while the DQ scores in \autoref{tab:diversity_of_domains} indicate that \text{EgRD} is more diverse than \text{ExSD}.

\section{Conclusion}  \label{sec:Conclusion}
In this study, we propose ISDQ score, which is composed of a scoring algorithm to measure task-level inter-agent interaction difficulty (IS), and the computation of the diversity of a scenario domain (DQ). We then apply ISDQ score to estimate the scenario generalization of an imitation learning or a model-free reinforcement learning model in decentralized crowd systems. Experimental results validate its efficacy in characterizing interaction difficulty, as well as its potentiality in selecting suitable training and testing domains before actually training and testing the imitation learning or the model-free reinforcement learning model. 

There might exist other applications/extensions of ISDQ score. For instance, we may apply IS score to the problem of multi-agent path finding (MAPF). The root complexity of MAPF lies in task-level inter-agent interaction difficulty. One way to reduce the complexity is to plan the paths for agents sequentially. A path already planned forms a spatial-temporal occupancy, restricting the path to be planned next ~\cite{silver2005cooperative, bnaya2014conflict, stern2019multi}. This is analogous to packing various items in a suitcase, where it is preferable to put large items first, and then put small items, to fully exploit the suitcase. Thus one could prioritize the sequential planning according to interaction difficulties: plan the most difficult task first, and then plan the second difficult task, and so on. On the other hand, regarding DQ score, since it is based on entropy: $\text{DQ}(\mathbb{S}) \doteq -H(\mathbb{S})$, one can extend it to construct a distance/difference metric between two scenario domains, for example $D_{KL}(\mathbb{S}_{1}\mid\mid \mathbb{S}_{2}) = H(\mathbb{S}_{1}, \mathbb{S}_{2}) - H(\mathbb{S}_{1})$, where $\mathbb{S}_{1}$ and $\mathbb{S}_{2}$ denote two scenario domains. 

There are mainly four limitations concerning the proposed ISDQ score in estimating scenario generalization. (i) ISDQ is an information-theoretic quantity, useful for selecting a domain among multiple candidate domains by ranking their scores. However, it does not have a physical unit, leading to a lack of clear physical meaning. (ii) The internal steering model may fail to steer an agent to reach its destination. In this case, there might be a large deviation in computing the IS score for this failed agent. (iii) In certain circumstances, the proposed approximation to scenario generalization might be rough in the sense that it does not reflect other factors of scenario generalization, including the number of scenarios in the source domain, the complexity of the learning model, and the training paradigm. Therefore, to analytically express the impact of various factors on scenario generalization, novel concepts and theoretical methods are needed to upper bound the generalization of such dynamic systems that are measured by application-oriented metrics within the output itself. This would be essentially different from the standard generalization analyses on static systems, based on simple metrics that measure the difference between the output and a ground truth result. (iv) Trajectories are spatial-temporal series, upon which various metrics can be measured. In \cite{daniel2021perceptually}, its final metric is a weighted combination of 21 metrics on crowd trajectories (e.g.,``Flicking in Direction", ``Interaction Strength", ``Interaction Anticipation", ``Difference to Comfort Speed"). If any of them does not preserve the ground truth ranking by the number of agent-agent collisions adopted in our experiment, consequently their weighted combination may not follow our ground truth ranking.

\bibliography{ref.bib}

\begin{thebibliography}{}
\providecommand{\doi}[1]{\url{https://doi.org/#1}}
\bibcommenthead

\bibitem [\protect \citeauthoryear {%
Alahi%
\ \protect \BOthers {.}}{%
Alahi%
\ \protect \BOthers {.}}{%
{\protect \APACyear {2016}}%
}]{%
alahi2016social}
\APACinsertmetastar {%
alahi2016social}%
\begin{APACrefauthors}%
Alahi, A.%
, Goel, K.%
, Ramanathan, V.%
, Robicquet, A.%
, Fei-Fei, L.%
\BCBL {} Savarese, S.%
\end{APACrefauthors}%
\unskip\
\newblock
\APACrefYearMonthDay{2016}{}{}.
\newblock
{\BBOQ}\APACrefatitle {Social lstm: Human trajectory prediction in crowded
  spaces} {Social lstm: Human trajectory prediction in crowded spaces}.{\BBCQ}
\newblock
 \APACrefbtitle {Proceedings of the IEEE conference on computer vision and
  pattern recognition} {Proceedings of the ieee conference on computer vision
  and pattern recognition}\ (\BPGS\ 961--971).
\newblock
\begin{APACrefURL} {\url{https://doi.org/10.1109/CVPR.2016.110}}
  \end{APACrefURL}
\PrintBackRefs{\CurrentBib}

\bibitem [\protect \citeauthoryear {%
Amirian%
\ \protect \BOthers {.}}{%
Amirian%
\ \protect \BOthers {.}}{%
{\protect \APACyear {2020}}%
}]{%
amirian2020opentraj}
\APACinsertmetastar {%
amirian2020opentraj}%
\begin{APACrefauthors}%
Amirian, J.%
, Zhang, B.%
, Castro, F.V.%
, Baldelomar, J.J.%
, Hayet, J\BHBI B.%
\BCBL {} Pettr{\'e}, J.%
\end{APACrefauthors}%
\unskip\
\newblock
\APACrefYearMonthDay{2020}{}{}.
\newblock
{\BBOQ}\APACrefatitle {Opentraj: Assessing prediction complexity in human
  trajectories datasets} {Opentraj: Assessing prediction complexity in human
  trajectories datasets}.{\BBCQ}
\newblock
 \APACrefbtitle {Proceedings of the Asian Conference on Computer Vision.}
  {Proceedings of the asian conference on computer vision.}
\newblock
\begin{APACrefURL} {\url{http://dx.doi.org/10.1007/978-3-030-69544-6_34}}
  \end{APACrefURL}
\PrintBackRefs{\CurrentBib}

\bibitem [\protect \citeauthoryear {%
Berseth%
, Kapadia%
\BCBL {}\ \BBA {} Faloutsos%
}{%
Berseth%
\ \protect \BOthers {.}}{%
{\protect \APACyear {2013}}%
}]{%
berseth2013steerplex}
\APACinsertmetastar {%
berseth2013steerplex}%
\begin{APACrefauthors}%
Berseth, G.%
, Kapadia, M.%
\BCBL {} Faloutsos, P.%
\end{APACrefauthors}%
\unskip\
\newblock
\APACrefYearMonthDay{2013}{}{}.
\newblock
{\BBOQ}\APACrefatitle {Steerplex: Estimating scenario complexity for simulated
  crowds} {Steerplex: Estimating scenario complexity for simulated
  crowds}.{\BBCQ}
\newblock
 \APACrefbtitle {Proceedings of Motion on Games} {Proceedings of motion on
  games}\ (\BPGS\ 67--76).
\newblock
\begin{APACrefURL} {\url{https://doi.org/10.1145/2522628.2522650}}
  \end{APACrefURL}
\PrintBackRefs{\CurrentBib}

\bibitem [\protect \citeauthoryear {%
Bhattacharya%
}{%
Bhattacharya%
}{%
{\protect \APACyear {2010}}%
}]{%
bhattacharya2010search}
\APACinsertmetastar {%
bhattacharya2010search}%
\begin{APACrefauthors}%
Bhattacharya, S.%
\end{APACrefauthors}%
\unskip\
\newblock
\APACrefYearMonthDay{2010}{}{}.
\newblock
{\BBOQ}\APACrefatitle {Search-based path planning with homotopy class
  constraints} {Search-based path planning with homotopy class
  constraints}.{\BBCQ}
\newblock
 \APACrefbtitle {Proceedings of the AAAI Conference on Artificial Intelligence}
  {Proceedings of the aaai conference on artificial intelligence}\ (\BVOL~24).
\PrintBackRefs{\CurrentBib}

\bibitem [\protect \citeauthoryear {%
Bian%
, Tian%
, Tang%
\BCBL {}\ \BBA {} Tao%
}{%
Bian%
\ \protect \BOthers {.}}{%
{\protect \APACyear {2018}}%
}]{%
bian2018survey}
\APACinsertmetastar {%
bian2018survey}%
\begin{APACrefauthors}%
Bian, J.%
, Tian, D.%
, Tang, Y.%
\BCBL {} Tao, D.%
\end{APACrefauthors}%
\unskip\
\newblock
\APACrefYearMonthDay{2018}{}{}.
\newblock
{\BBOQ}\APACrefatitle {A survey on trajectory clustering analysis} {A survey on
  trajectory clustering analysis}.{\BBCQ}
\newblock
\APACjournalVolNumPages{arXiv preprint arXiv:1802.06971}{}{}{}.
\newblock
\begin{APACrefURL} {\url{https://doi.org/10.48550/arXiv.1802.06971}}
  \end{APACrefURL}
\newblock

\newblock

\PrintBackRefs{\CurrentBib}

\bibitem [\protect \citeauthoryear {%
Bnaya%
\ \BBA {} Felner%
}{%
Bnaya%
\ \BBA {} Felner%
}{%
{\protect \APACyear {2014}}%
}]{%
bnaya2014conflict}
\APACinsertmetastar {%
bnaya2014conflict}%
\begin{APACrefauthors}%
Bnaya, Z.%
\BCBT {}\ \BBA {} Felner, A.%
\end{APACrefauthors}%
\unskip\
\newblock
\APACrefYearMonthDay{2014}{}{}.
\newblock
{\BBOQ}\APACrefatitle {Conflict-oriented windowed hierarchical cooperative
  A$*$} {Conflict-oriented windowed hierarchical cooperative a$*$}.{\BBCQ}
\newblock
 \APACrefbtitle {2014 IEEE International Conference on Robotics and Automation
  (ICRA)} {2014 ieee international conference on robotics and automation
  (icra)}\ (\BPGS\ 3743--3748).
\newblock
\begin{APACrefURL} {\url{https://doi.org/10.1109/ICRA.2014.6907401}}
  \end{APACrefURL}
\PrintBackRefs{\CurrentBib}

\bibitem [\protect \citeauthoryear {%
Corneanu%
, Escalera%
\BCBL {}\ \BBA {} Martinez%
}{%
Corneanu%
\ \protect \BOthers {.}}{%
{\protect \APACyear {2020}}%
}]{%
corneanu2020computing}
\APACinsertmetastar {%
corneanu2020computing}%
\begin{APACrefauthors}%
Corneanu, C.A.%
, Escalera, S.%
\BCBL {} Martinez, A.M.%
\end{APACrefauthors}%
\unskip\
\newblock
\APACrefYearMonthDay{2020}{}{}.
\newblock
{\BBOQ}\APACrefatitle {Computing the testing error without a testing set}
  {Computing the testing error without a testing set}.{\BBCQ}
\newblock
 \APACrefbtitle {Proceedings of the IEEE/CVF Conference on Computer Vision and
  Pattern Recognition} {Proceedings of the ieee/cvf conference on computer
  vision and pattern recognition}\ (\BPGS\ 2677--2685).
\newblock
\begin{APACrefURL} {\url{https://doi.org/10.1109/CVPR42600.2020.00275}}
  \end{APACrefURL}
\PrintBackRefs{\CurrentBib}

\bibitem [\protect \citeauthoryear {%
Corneanu%
, Madadi%
, Escalera%
\BCBL {}\ \BBA {} Martinez%
}{%
Corneanu%
\ \protect \BOthers {.}}{%
{\protect \APACyear {2019}}%
}]{%
corneanu2019does}
\APACinsertmetastar {%
corneanu2019does}%
\begin{APACrefauthors}%
Corneanu, C.A.%
, Madadi, M.%
, Escalera, S.%
\BCBL {} Martinez, A.M.%
\end{APACrefauthors}%
\unskip\
\newblock
\APACrefYearMonthDay{2019}{}{}.
\newblock
{\BBOQ}\APACrefatitle {What does it mean to learn in deep networks? And, how
  does one detect adversarial attacks?} {What does it mean to learn in deep
  networks? and, how does one detect adversarial attacks?}{\BBCQ}
\newblock
 \APACrefbtitle {Proceedings of the IEEE/CVF Conference on Computer Vision and
  Pattern Recognition} {Proceedings of the ieee/cvf conference on computer
  vision and pattern recognition}\ (\BPGS\ 4757--4766).
\newblock
\begin{APACrefURL} {\url{https://doi.org/10.1109/CVPR.2019.00489}}
  \end{APACrefURL}
\PrintBackRefs{\CurrentBib}

\bibitem [\protect \citeauthoryear {%
Curtis%
, Zafar%
, Gutub%
\BCBL {}\ \BBA {} Manocha%
}{%
Curtis%
\ \protect \BOthers {.}}{%
{\protect \APACyear {2013}}%
}]{%
curtis2013right}
\APACinsertmetastar {%
curtis2013right}%
\begin{APACrefauthors}%
Curtis, S.%
, Zafar, B.%
, Gutub, A.%
\BCBL {} Manocha, D.%
\end{APACrefauthors}%
\unskip\
\newblock
\APACrefYearMonthDay{2013}{}{}.
\newblock
{\BBOQ}\APACrefatitle {Right of way} {Right of way}.{\BBCQ}
\newblock
\APACjournalVolNumPages{The Visual Computer}{29}{12}{1277--1292}.
\newblock
\begin{APACrefURL} {\url{https://doi.org/10.1007/s00371-012-0769-x}}
  \end{APACrefURL}
\newblock

\newblock

\PrintBackRefs{\CurrentBib}

\bibitem [\protect \citeauthoryear {%
Daniel%
, Marques%
, Hoyet%
, Pettr{\'e}%
\BCBL {}\ \BBA {} Blat%
}{%
Daniel%
\ \protect \BOthers {.}}{%
{\protect \APACyear {2021}}%
}]{%
daniel2021perceptually}
\APACinsertmetastar {%
daniel2021perceptually}%
\begin{APACrefauthors}%
Daniel, B.C.%
, Marques, R.%
, Hoyet, L.%
, Pettr{\'e}, J.%
\BCBL {} Blat, J.%
\end{APACrefauthors}%
\unskip\
\newblock
\APACrefYearMonthDay{2021}{}{}.
\newblock
{\BBOQ}\APACrefatitle {A Perceptually-Validated Metric for Crowd Trajectory
  Quality Evaluation} {A perceptually-validated metric for crowd trajectory
  quality evaluation}.{\BBCQ}
\newblock
\APACjournalVolNumPages{Proceedings of the ACM on Computer Graphics and
  Interactive Techniques}{4}{3}{1--18}.
\newblock
\begin{APACrefURL} {\url{https://doi.org/10.1145/3480136}} \end{APACrefURL}
\newblock

\newblock

\PrintBackRefs{\CurrentBib}

\bibitem [\protect \citeauthoryear {%
Emonet%
, Varadarajan%
\BCBL {}\ \BBA {} Odobez%
}{%
Emonet%
\ \protect \BOthers {.}}{%
{\protect \APACyear {2011}}%
}]{%
emonet2011extracting}
\APACinsertmetastar {%
emonet2011extracting}%
\begin{APACrefauthors}%
Emonet, R.%
, Varadarajan, J.%
\BCBL {} Odobez, J\BHBI M.%
\end{APACrefauthors}%
\unskip\
\newblock
\APACrefYearMonthDay{2011}{}{}.
\newblock
{\BBOQ}\APACrefatitle {Extracting and locating temporal motifs in video scenes
  using a hierarchical non parametric bayesian model} {Extracting and locating
  temporal motifs in video scenes using a hierarchical non parametric bayesian
  model}.{\BBCQ}
\newblock
 \APACrefbtitle {CVPR 2011} {Cvpr 2011}\ (\BPGS\ 3233--3240).
\newblock
\begin{APACrefURL} {\url{https://doi.org/10.1109/CVPR.2011.5995572}}
  \end{APACrefURL}
\PrintBackRefs{\CurrentBib}

\bibitem [\protect \citeauthoryear {%
Ester%
, Kriegel%
, Sander%
, Xu%
\BCBL {}\ \protect \BOthers {.}}{%
Ester%
\ \protect \BOthers {.}}{%
{\protect \APACyear {1996}}%
}]{%
ester1996density}
\APACinsertmetastar {%
ester1996density}%
\begin{APACrefauthors}%
Ester, M.%
, Kriegel, H\BHBI P.%
, Sander, J.%
, Xu, X.%
\BCBL {}\ \BOthersPeriod {.}\end{APACrefauthors}%
\unskip\
\newblock
\APACrefYearMonthDay{1996}{}{}.
\newblock
{\BBOQ}\APACrefatitle {A density-based algorithm for discovering clusters in
  large spatial databases with noise.} {A density-based algorithm for
  discovering clusters in large spatial databases with noise.}{\BBCQ}
\newblock
 \APACrefbtitle {kdd} {kdd}\ (\BVOL~96, \BPGS\ 226--231).
\PrintBackRefs{\CurrentBib}

\bibitem [\protect \citeauthoryear {%
Hansen%
\ \BBA {} Larsen%
}{%
Hansen%
\ \BBA {} Larsen%
}{%
{\protect \APACyear {1996}}%
}]{%
hansen1996unsupervised}
\APACinsertmetastar {%
hansen1996unsupervised}%
\begin{APACrefauthors}%
Hansen, L.K.%
\BCBT {}\ \BBA {} Larsen, J.%
\end{APACrefauthors}%
\unskip\
\newblock
\APACrefYearMonthDay{1996}{}{}.
\newblock
{\BBOQ}\APACrefatitle {Unsupervised learning and generalization} {Unsupervised
  learning and generalization}.{\BBCQ}
\newblock
 \APACrefbtitle {Proceedings of International Conference on Neural Networks
  (ICNN'96)} {Proceedings of international conference on neural networks
  (icnn'96)}\ (\BVOL~1, \BPGS\ 25--30).
\newblock
\begin{APACrefURL} {\url{https://doi.org/10.1109/ICNN.1996.548861}}
  \end{APACrefURL}
\PrintBackRefs{\CurrentBib}

\bibitem [\protect \citeauthoryear {%
He%
, Xiang%
, Zhao%
\BCBL {}\ \BBA {} Wang%
}{%
He%
\ \protect \BOthers {.}}{%
{\protect \APACyear {2020}}%
}]{%
he2020informative}
\APACinsertmetastar {%
he2020informative}%
\begin{APACrefauthors}%
He, F.%
, Xiang, Y.%
, Zhao, X.%
\BCBL {} Wang, H.%
\end{APACrefauthors}%
\unskip\
\newblock
\APACrefYearMonthDay{2020}{}{}.
\newblock
{\BBOQ}\APACrefatitle {Informative scene decomposition for crowd analysis,
  comparison and simulation guidance} {Informative scene decomposition for
  crowd analysis, comparison and simulation guidance}.{\BBCQ}
\newblock
\APACjournalVolNumPages{ACM Transactions on Graphics (TOG)}{39}{4}{50--1}.
\newblock
\begin{APACrefURL} {\url{https://doi.org/10.48550/arXiv.2004.14107}}
  \end{APACrefURL}
\newblock

\newblock

\PrintBackRefs{\CurrentBib}

\bibitem [\protect \citeauthoryear {%
Helbing%
, Farkas%
\BCBL {}\ \BBA {} Vicsek%
}{%
Helbing%
\ \protect \BOthers {.}}{%
{\protect \APACyear {2000}}%
}]{%
helbing2000simulating}
\APACinsertmetastar {%
helbing2000simulating}%
\begin{APACrefauthors}%
Helbing, D.%
, Farkas, I.%
\BCBL {} Vicsek, T.%
\end{APACrefauthors}%
\unskip\
\newblock
\APACrefYearMonthDay{2000}{}{}.
\newblock
{\BBOQ}\APACrefatitle {Simulating dynamical features of escape panic}
  {Simulating dynamical features of escape panic}.{\BBCQ}
\newblock
\APACjournalVolNumPages{Nature}{407}{6803}{487--490}.
\newblock
\begin{APACrefURL} {\url{https://doi.org/10.1038/35035023}} \end{APACrefURL}
\newblock

\newblock

\PrintBackRefs{\CurrentBib}

\bibitem [\protect \citeauthoryear {%
Ho%
\ \BBA {} Ermon%
}{%
Ho%
\ \BBA {} Ermon%
}{%
{\protect \APACyear {2016}}%
}]{%
ho2016generative}
\APACinsertmetastar {%
ho2016generative}%
\begin{APACrefauthors}%
Ho, J.%
\BCBT {}\ \BBA {} Ermon, S.%
\end{APACrefauthors}%
\unskip\
\newblock
\APACrefYearMonthDay{2016}{}{}.
\newblock
{\BBOQ}\APACrefatitle {Generative adversarial imitation learning} {Generative
  adversarial imitation learning}.{\BBCQ}
\newblock
\APACjournalVolNumPages{Advances in neural information processing
  systems}{29}{}{4565--4573}.
\newblock

\newblock

\PrintBackRefs{\CurrentBib}

\bibitem [\protect \citeauthoryear {%
Jaques%
\ \protect \BOthers {.}}{%
Jaques%
\ \protect \BOthers {.}}{%
{\protect \APACyear {2019}}%
}]{%
jaques2019social}
\APACinsertmetastar {%
jaques2019social}%
\begin{APACrefauthors}%
Jaques, N.%
, Lazaridou, A.%
, Hughes, E.%
, Gulcehre, C.%
, Ortega, P.%
, Strouse, D.%
\BDBL {}De~Freitas, N.%
\end{APACrefauthors}%
\unskip\
\newblock
\APACrefYearMonthDay{2019}{}{}.
\newblock
{\BBOQ}\APACrefatitle {Social influence as intrinsic motivation for multi-agent
  deep reinforcement learning} {Social influence as intrinsic motivation for
  multi-agent deep reinforcement learning}.{\BBCQ}
\newblock
 \APACrefbtitle {International Conference on Machine Learning} {International
  conference on machine learning}\ (\BPGS\ 3040--3049).
\PrintBackRefs{\CurrentBib}

\bibitem [\protect \citeauthoryear {%
Kapadia%
, Berseth%
, Singh%
, Reinman%
\BCBL {}\ \BBA {} Faloutsos%
}{%
Kapadia%
\ \protect \BOthers {.}}{%
{\protect \APACyear {2016}}%
}]{%
kapadia2016scenario}
\APACinsertmetastar {%
kapadia2016scenario}%
\begin{APACrefauthors}%
Kapadia, M.%
, Berseth, G.%
, Singh, S.%
, Reinman, G.%
\BCBL {} Faloutsos, P.%
\end{APACrefauthors}%
\unskip\
\newblock
\APACrefYearMonthDay{2016}{}{}.
\newblock
{\BBOQ}\APACrefatitle {Scenario space: characterizing coverage, quality, and
  failure of steering algorithms} {Scenario space: characterizing coverage,
  quality, and failure of steering algorithms}.{\BBCQ}
\newblock
 \APACrefbtitle {Simulating Heterogeneous Crowds with Interactive Behaviors}
  {Simulating heterogeneous crowds with interactive behaviors}\ (\BPGS\
  193--210).
\newblock
\APACaddressPublisher{}{AK Peters/CRC Press}.
\newblock
\begin{APACrefURL} {\url{https://doi.org/10.1145/2019406.2019414}}
  \end{APACrefURL}
\PrintBackRefs{\CurrentBib}

\bibitem [\protect \citeauthoryear {%
Karamouzas%
, Skinner%
\BCBL {}\ \BBA {} Guy%
}{%
Karamouzas%
\ \protect \BOthers {.}}{%
{\protect \APACyear {2014}}%
}]{%
karamouzas2014universal}
\APACinsertmetastar {%
karamouzas2014universal}%
\begin{APACrefauthors}%
Karamouzas, I.%
, Skinner, B.%
\BCBL {} Guy, S.J.%
\end{APACrefauthors}%
\unskip\
\newblock
\APACrefYearMonthDay{2014}{}{}.
\newblock
{\BBOQ}\APACrefatitle {Universal power law governing pedestrian interactions}
  {Universal power law governing pedestrian interactions}.{\BBCQ}
\newblock
\APACjournalVolNumPages{Physical review letters}{113}{23}{238701}.
\newblock
\begin{APACrefURL} {\url{https://doi.org/10.1103/PhysRevLett.113.238701}}
  \end{APACrefURL}
\newblock

\newblock

\PrintBackRefs{\CurrentBib}

\bibitem [\protect \citeauthoryear {%
Karamouzas%
, Sohre%
, Hu%
\BCBL {}\ \BBA {} Guy%
}{%
Karamouzas%
\ \protect \BOthers {.}}{%
{\protect \APACyear {2018}}%
}]{%
karamouzas2018crowd}
\APACinsertmetastar {%
karamouzas2018crowd}%
\begin{APACrefauthors}%
Karamouzas, I.%
, Sohre, N.%
, Hu, R.%
\BCBL {} Guy, S.J.%
\end{APACrefauthors}%
\unskip\
\newblock
\APACrefYearMonthDay{2018}{}{}.
\newblock
{\BBOQ}\APACrefatitle {Crowd space: a predictive crowd analysis technique}
  {Crowd space: a predictive crowd analysis technique}.{\BBCQ}
\newblock
\APACjournalVolNumPages{ACM Transactions on Graphics (TOG)}{37}{6}{1--14}.
\newblock
\begin{APACrefURL} {\url{https://doi.org/10.1145/3272127.3275079}}
  \end{APACrefURL}
\newblock

\newblock

\PrintBackRefs{\CurrentBib}

\bibitem [\protect \citeauthoryear {%
Knob%
, de Andrade~Araujo%
, Favaretto%
\BCBL {}\ \BBA {} Musse%
}{%
Knob%
\ \protect \BOthers {.}}{%
{\protect \APACyear {2018}}%
}]{%
knob2018visualization}
\APACinsertmetastar {%
knob2018visualization}%
\begin{APACrefauthors}%
Knob, P.%
, de Andrade~Araujo, V.F.%
, Favaretto, R.M.%
\BCBL {} Musse, S.R.%
\end{APACrefauthors}%
\unskip\
\newblock
\APACrefYearMonthDay{2018}{}{}.
\newblock
{\BBOQ}\APACrefatitle {Visualization of interactions in crowd simulation and
  video sequences} {Visualization of interactions in crowd simulation and video
  sequences}.{\BBCQ}
\newblock
 \APACrefbtitle {2018 17th Brazilian Symposium on Computer Games and Digital
  Entertainment (SBGames)} {2018 17th brazilian symposium on computer games and
  digital entertainment (sbgames)}\ (\BPGS\ 250--25009).
\newblock
\begin{APACrefURL} {\url{https://doi.org/10.1109/SBGAMES.2018.00037}}
  \end{APACrefURL}
\PrintBackRefs{\CurrentBib}

\bibitem [\protect \citeauthoryear {%
Lee%
, Won%
\BCBL {}\ \BBA {} Lee%
}{%
Lee%
\ \protect \BOthers {.}}{%
{\protect \APACyear {2018}}%
}]{%
lee2018crowd}
\APACinsertmetastar {%
lee2018crowd}%
\begin{APACrefauthors}%
Lee, J.%
, Won, J.%
\BCBL {} Lee, J.%
\end{APACrefauthors}%
\unskip\
\newblock
\APACrefYearMonthDay{2018}{}{}.
\newblock
{\BBOQ}\APACrefatitle {Crowd simulation by deep reinforcement learning} {Crowd
  simulation by deep reinforcement learning}.{\BBCQ}
\newblock
 \APACrefbtitle {Proceedings of the 11th Annual International Conference on
  Motion, Interaction, and Games} {Proceedings of the 11th annual international
  conference on motion, interaction, and games}\ (\BPGS\ 1--7).
\newblock
\begin{APACrefURL} {\url{https://doi.org/10.1145/3274247.3274510}}
  \end{APACrefURL}
\PrintBackRefs{\CurrentBib}

\bibitem [\protect \citeauthoryear {%
Li%
, Ma%
\BCBL {}\ \BBA {} Tomizuka%
}{%
Li%
\ \protect \BOthers {.}}{%
{\protect \APACyear {2019}}%
}]{%
li2019interaction}
\APACinsertmetastar {%
li2019interaction}%
\begin{APACrefauthors}%
Li, J.%
, Ma, H.%
\BCBL {} Tomizuka, M.%
\end{APACrefauthors}%
\unskip\
\newblock
\APACrefYearMonthDay{2019}{}{}.
\newblock
{\BBOQ}\APACrefatitle {Interaction-aware multi-agent tracking and probabilistic
  behavior prediction via adversarial learning} {Interaction-aware multi-agent
  tracking and probabilistic behavior prediction via adversarial
  learning}.{\BBCQ}
\newblock
 \APACrefbtitle {2019 international conference on robotics and automation
  (ICRA)} {2019 international conference on robotics and automation (icra)}\
  (\BPGS\ 6658--6664).
\newblock
\begin{APACrefURL} {\url{https://doi.org/10.1109/ICRA.2019.8793661}}
  \end{APACrefURL}
\PrintBackRefs{\CurrentBib}

\bibitem [\protect \citeauthoryear {%
Long%
, Liu%
\BCBL {}\ \BBA {} Pan%
}{%
Long%
\ \protect \BOthers {.}}{%
{\protect \APACyear {2017}}%
}]{%
long2017deep}
\APACinsertmetastar {%
long2017deep}%
\begin{APACrefauthors}%
Long, P.%
, Liu, W.%
\BCBL {} Pan, J.%
\end{APACrefauthors}%
\unskip\
\newblock
\APACrefYearMonthDay{2017}{}{}.
\newblock
{\BBOQ}\APACrefatitle {Deep-learned collision avoidance policy for distributed
  multiagent navigation} {Deep-learned collision avoidance policy for
  distributed multiagent navigation}.{\BBCQ}
\newblock
\APACjournalVolNumPages{IEEE Robotics and Automation Letters}{2}{2}{656--663}.
\newblock
\begin{APACrefURL} {\url{https://doi.org/10.1109/LRA.2017.2651371}}
  \end{APACrefURL}
\newblock

\newblock

\PrintBackRefs{\CurrentBib}

\bibitem [\protect \citeauthoryear {%
Mohamed%
, Qian%
, Elhoseiny%
\BCBL {}\ \BBA {} Claudel%
}{%
Mohamed%
\ \protect \BOthers {.}}{%
{\protect \APACyear {2020}}%
}]{%
mohamed2020social}
\APACinsertmetastar {%
mohamed2020social}%
\begin{APACrefauthors}%
Mohamed, A.%
, Qian, K.%
, Elhoseiny, M.%
\BCBL {} Claudel, C.%
\end{APACrefauthors}%
\unskip\
\newblock
\APACrefYearMonthDay{2020}{}{}.
\newblock
{\BBOQ}\APACrefatitle {Social-stgcnn: A social spatio-temporal graph
  convolutional neural network for human trajectory prediction} {Social-stgcnn:
  A social spatio-temporal graph convolutional neural network for human
  trajectory prediction}.{\BBCQ}
\newblock
 \APACrefbtitle {Proceedings of the IEEE/CVF Conference on Computer Vision and
  Pattern Recognition} {Proceedings of the ieee/cvf conference on computer
  vision and pattern recognition}\ (\BPGS\ 14424--14432).
\newblock
\begin{APACrefURL} {\url{https://doi.org/10.1109/CVPR42600.2020.01443}}
  \end{APACrefURL}
\PrintBackRefs{\CurrentBib}

\bibitem [\protect \citeauthoryear {%
Munch%
}{%
Munch%
}{%
{\protect \APACyear {2017}}%
}]{%
munch2017user}
\APACinsertmetastar {%
munch2017user}%
\begin{APACrefauthors}%
Munch, E.%
\end{APACrefauthors}%
\unskip\
\newblock
\APACrefYearMonthDay{2017}{}{}.
\newblock
{\BBOQ}\APACrefatitle {A user’s guide to topological data analysis} {A
  user’s guide to topological data analysis}.{\BBCQ}
\newblock
\APACjournalVolNumPages{Journal of Learning Analytics}{4}{2}{47--61}.
\newblock
\begin{APACrefURL} {\url{http://dx.doi.org/10.18608/jla.2017.42.6}}
  \end{APACrefURL}
\newblock

\newblock

\PrintBackRefs{\CurrentBib}

\bibitem [\protect \citeauthoryear {%
Olivier%
, Marin%
, Cr{\'e}tual%
, Berthoz%
\BCBL {}\ \BBA {} Pettr{\'e}%
}{%
Olivier%
\ \protect \BOthers {.}}{%
{\protect \APACyear {2013}}%
}]{%
olivier2013collision}
\APACinsertmetastar {%
olivier2013collision}%
\begin{APACrefauthors}%
Olivier, A\BHBI H.%
, Marin, A.%
, Cr{\'e}tual, A.%
, Berthoz, A.%
\BCBL {} Pettr{\'e}, J.%
\end{APACrefauthors}%
\unskip\
\newblock
\APACrefYearMonthDay{2013}{}{}.
\newblock
{\BBOQ}\APACrefatitle {Collision avoidance between two walkers: Role-dependent
  strategies} {Collision avoidance between two walkers: Role-dependent
  strategies}.{\BBCQ}
\newblock
\APACjournalVolNumPages{Gait \& posture}{38}{4}{751--756}.
\newblock
\begin{APACrefURL} {\url{https://doi.org/10.1016/j.gaitpost.2013.03.017}}
  \end{APACrefURL}
\newblock

\newblock

\PrintBackRefs{\CurrentBib}

\bibitem [\protect \citeauthoryear {%
Qiao%
, Yoon%
, Kapadia%
\BCBL {}\ \BBA {} Pavlovic%
}{%
Qiao%
\ \protect \BOthers {.}}{%
{\protect \APACyear {2018}}%
}]{%
qiao2018role}
\APACinsertmetastar {%
qiao2018role}%
\begin{APACrefauthors}%
Qiao, G.%
, Yoon, S.%
, Kapadia, M.%
\BCBL {} Pavlovic, V.%
\end{APACrefauthors}%
\unskip\
\newblock
\APACrefYearMonthDay{2018}{}{}.
\newblock
{\BBOQ}\APACrefatitle {The role of data-driven priors in multi-agent crowd
  trajectory estimation} {The role of data-driven priors in multi-agent crowd
  trajectory estimation}.{\BBCQ}
\newblock
 \APACrefbtitle {Thirty-Second AAAI Conference on Artificial Intelligence.}
  {Thirty-second aaai conference on artificial intelligence.}
\PrintBackRefs{\CurrentBib}

\bibitem [\protect \citeauthoryear {%
Qiao%
, Zhou%
, Kapadia%
, Yoon%
\BCBL {}\ \BBA {} Pavlovic%
}{%
Qiao%
\ \protect \BOthers {.}}{%
{\protect \APACyear {2019}}%
}]{%
qiao2019scenario}
\APACinsertmetastar {%
qiao2019scenario}%
\begin{APACrefauthors}%
Qiao, G.%
, Zhou, H.%
, Kapadia, M.%
, Yoon, S.%
\BCBL {} Pavlovic, V.%
\end{APACrefauthors}%
\unskip\
\newblock
\APACrefYearMonthDay{2019}{}{}.
\newblock
{\BBOQ}\APACrefatitle {Scenario generalization of data-driven imitation models
  in crowd simulation} {Scenario generalization of data-driven imitation models
  in crowd simulation}.{\BBCQ}
\newblock
 \APACrefbtitle {Motion, Interaction and Games} {Motion, interaction and
  games}\ (\BPGS\ 1--11).
\newblock
\begin{APACrefURL} {\url{https://doi.org/10.1145/3359566.3360087}}
  \end{APACrefURL}
\PrintBackRefs{\CurrentBib}

\bibitem [\protect \citeauthoryear {%
Robicquet%
, Sadeghian%
, Alahi%
\BCBL {}\ \BBA {} Savarese%
}{%
Robicquet%
\ \protect \BOthers {.}}{%
{\protect \APACyear {2016}}%
}]{%
robicquet2016learning}
\APACinsertmetastar {%
robicquet2016learning}%
\begin{APACrefauthors}%
Robicquet, A.%
, Sadeghian, A.%
, Alahi, A.%
\BCBL {} Savarese, S.%
\end{APACrefauthors}%
\unskip\
\newblock
\APACrefYearMonthDay{2016}{}{}.
\newblock
{\BBOQ}\APACrefatitle {Learning social etiquette: Human trajectory
  understanding in crowded scenes} {Learning social etiquette: Human trajectory
  understanding in crowded scenes}.{\BBCQ}
\newblock
 \APACrefbtitle {European conference on computer vision} {European conference
  on computer vision}\ (\BPGS\ 549--565).
\newblock
\begin{APACrefURL} {\url{http://dx.doi.org/10.1007/978-3-319-46484-8_33}}
  \end{APACrefURL}
\PrintBackRefs{\CurrentBib}

\bibitem [\protect \citeauthoryear {%
Salvador%
\ \BBA {} Chan%
}{%
Salvador%
\ \BBA {} Chan%
}{%
{\protect \APACyear {2007}}%
}]{%
salvador2007toward}
\APACinsertmetastar {%
salvador2007toward}%
\begin{APACrefauthors}%
Salvador, S.%
\BCBT {}\ \BBA {} Chan, P.%
\end{APACrefauthors}%
\unskip\
\newblock
\APACrefYearMonthDay{2007}{}{}.
\newblock
{\BBOQ}\APACrefatitle {Toward accurate dynamic time warping in linear time and
  space} {Toward accurate dynamic time warping in linear time and
  space}.{\BBCQ}
\newblock
\APACjournalVolNumPages{Intelligent Data Analysis}{11}{5}{561--580}.
\newblock
\begin{APACrefURL} {\url{https://dl.acm.org/doi/10.5555/1367985.1367993}}
  \end{APACrefURL}
\newblock

\newblock

\PrintBackRefs{\CurrentBib}

\bibitem [\protect \citeauthoryear {%
Schulman%
, Wolski%
, Dhariwal%
, Radford%
\BCBL {}\ \BBA {} Klimov%
}{%
Schulman%
\ \protect \BOthers {.}}{%
{\protect \APACyear {2017}}%
}]{%
schulman2017proximal}
\APACinsertmetastar {%
schulman2017proximal}%
\begin{APACrefauthors}%
Schulman, J.%
, Wolski, F.%
, Dhariwal, P.%
, Radford, A.%
\BCBL {} Klimov, O.%
\end{APACrefauthors}%
\unskip\
\newblock
\APACrefYearMonthDay{2017}{}{}.
\newblock
{\BBOQ}\APACrefatitle {Proximal policy optimization algorithms} {Proximal
  policy optimization algorithms}.{\BBCQ}
\newblock
\APACjournalVolNumPages{arXiv preprint arXiv:1707.06347}{}{}{}.
\newblock

\newblock

\PrintBackRefs{\CurrentBib}

\bibitem [\protect \citeauthoryear {%
Silver%
}{%
Silver%
}{%
{\protect \APACyear {2005}}%
}]{%
silver2005cooperative}
\APACinsertmetastar {%
silver2005cooperative}%
\begin{APACrefauthors}%
Silver, D.%
\end{APACrefauthors}%
\unskip\
\newblock
\APACrefYearMonthDay{2005}{}{}.
\newblock
{\BBOQ}\APACrefatitle {Cooperative Pathfinding} {Cooperative
  pathfinding}.{\BBCQ}
\newblock
\APACjournalVolNumPages{Artificial Intelligence and Interactive Digital
  Entertainment (AIIDE)}{1}{}{117--122}.
\newblock

\newblock

\PrintBackRefs{\CurrentBib}

\bibitem [\protect \citeauthoryear {%
Sohn%
\ \protect \BOthers {.}}{%
Sohn%
\ \protect \BOthers {.}}{%
{\protect \APACyear {2021}}%
}]{%
DBLP:conf/mig/SohnLMQ0YPK21}
\APACinsertmetastar {%
DBLP:conf/mig/SohnLMQ0YPK21}%
\begin{APACrefauthors}%
Sohn, S.S.%
, Lee, M.%
, Moon, S.%
, Qiao, G.%
, Usman, M.%
, Yoon, S.%
\BDBL {}Kapadia, M.%
\end{APACrefauthors}%
\unskip\
\newblock
\APACrefYearMonthDay{2021}{}{}.
\newblock
{\BBOQ}\APACrefatitle {{A2X:} An Agent and Environment Interaction Benchmark
  for Multimodal Human Trajectory Prediction} {{A2X:} an agent and environment
  interaction benchmark for multimodal human trajectory prediction}.{\BBCQ}
\newblock
 R.~Boulic, L.~Hoyet, K.~Singh\BCBL {}\ \BBA {} D.~Rohmer\ (\BEDS),
  \APACrefbtitle {{MIG} '21: Motion, Interaction and Games, Virtual Event,
  Switzerland, November 10-12, 2021} {{MIG} '21: Motion, interaction and games,
  virtual event, switzerland, november 10-12, 2021}\ (\BPGS\ 19:1--19:9).
\newblock
\APACaddressPublisher{}{{ACM}}.
\newblock
\begin{APACrefURL} {\url{https://doi.org/10.1145/3487983.3488302}}
  \end{APACrefURL}
\PrintBackRefs{\CurrentBib}

\bibitem [\protect \citeauthoryear {%
Stern%
\ \protect \BOthers {.}}{%
Stern%
\ \protect \BOthers {.}}{%
{\protect \APACyear {2019}}%
}]{%
stern2019multi}
\APACinsertmetastar {%
stern2019multi}%
\begin{APACrefauthors}%
Stern, R.%
, Sturtevant, N.R.%
, Felner, A.%
, Koenig, S.%
, Ma, H.%
, Walker, T.T.%
\BDBL {}others%
\end{APACrefauthors}%
\unskip\
\newblock
\APACrefYearMonthDay{2019}{}{}.
\newblock
{\BBOQ}\APACrefatitle {Multi-agent pathfinding: Definitions, variants, and
  benchmarks} {Multi-agent pathfinding: Definitions, variants, and
  benchmarks}.{\BBCQ}
\newblock
 \APACrefbtitle {Twelfth Annual Symposium on Combinatorial Search.} {Twelfth
  annual symposium on combinatorial search.}
\PrintBackRefs{\CurrentBib}

\bibitem [\protect \citeauthoryear {%
Van Den~Berg%
, Guy%
, Lin%
\BCBL {}\ \BBA {} Manocha%
}{%
Van Den~Berg%
\ \protect \BOthers {.}}{%
{\protect \APACyear {2011}}%
}]{%
van2011reciprocal}
\APACinsertmetastar {%
van2011reciprocal}%
\begin{APACrefauthors}%
Van Den~Berg, J.%
, Guy, S.J.%
, Lin, M.%
\BCBL {} Manocha, D.%
\end{APACrefauthors}%
\unskip\
\newblock
\APACrefYearMonthDay{2011}{}{}.
\newblock
{\BBOQ}\APACrefatitle {Reciprocal n-body collision avoidance} {Reciprocal
  n-body collision avoidance}.{\BBCQ}
\newblock
 \APACrefbtitle {Robotics research} {Robotics research}\ (\BPGS\ 3--19).
\newblock
\APACaddressPublisher{}{Springer}.
\newblock
\begin{APACrefURL} {\url{https://doi.org/10.1007/978-3-642-19457-3_1}}
  \end{APACrefURL}
\PrintBackRefs{\CurrentBib}

\bibitem [\protect \citeauthoryear {%
H.~Wang%
, Ond{\v{r}}ej%
\BCBL {}\ \BBA {} O'Sullivan%
}{%
H.~Wang%
, Ond{\v{r}}ej%
\BCBL {}\ \BBA {} O'Sullivan%
}{%
{\protect \APACyear {2016}}%
}]{%
wang2016path}
\APACinsertmetastar {%
wang2016path}%
\begin{APACrefauthors}%
Wang, H.%
, Ond{\v{r}}ej, J.%
\BCBL {} O'Sullivan, C.%
\end{APACrefauthors}%
\unskip\
\newblock
\APACrefYearMonthDay{2016}{}{}.
\newblock
{\BBOQ}\APACrefatitle {Path patterns: Analyzing and comparing real and
  simulated crowds} {Path patterns: Analyzing and comparing real and simulated
  crowds}.{\BBCQ}
\newblock
 \APACrefbtitle {Proceedings of the 20th acm siggraph symposium on interactive
  3d graphics and games} {Proceedings of the 20th acm siggraph symposium on
  interactive 3d graphics and games}\ (\BPGS\ 49--57).
\newblock
\begin{APACrefURL} {\url{https://doi.org/10.1145/2856400.2856410}}
  \end{APACrefURL}
\PrintBackRefs{\CurrentBib}

\bibitem [\protect \citeauthoryear {%
H.~Wang%
, Ond{\v{r}}ej%
\BCBL {}\ \BBA {} O’Sullivan%
}{%
H.~Wang%
, Ond{\v{r}}ej%
\BCBL {}\ \BBA {} O’Sullivan%
}{%
{\protect \APACyear {2016}}%
}]{%
wang2016trending}
\APACinsertmetastar {%
wang2016trending}%
\begin{APACrefauthors}%
Wang, H.%
, Ond{\v{r}}ej, J.%
\BCBL {} O’Sullivan, C.%
\end{APACrefauthors}%
\unskip\
\newblock
\APACrefYearMonthDay{2016}{}{}.
\newblock
{\BBOQ}\APACrefatitle {Trending paths: A new semantic-level metric for
  comparing simulated and real crowd data} {Trending paths: A new
  semantic-level metric for comparing simulated and real crowd data}.{\BBCQ}
\newblock
\APACjournalVolNumPages{IEEE transactions on visualization and computer
  graphics}{23}{5}{1454--1464}.
\newblock
\begin{APACrefURL} {\url{https://doi.org/10.1109/TVCG.2016.2642963}}
  \end{APACrefURL}
\newblock

\newblock

\PrintBackRefs{\CurrentBib}

\bibitem [\protect \citeauthoryear {%
H.~Wang%
\ \BBA {} O’Sullivan%
}{%
H.~Wang%
\ \BBA {} O’Sullivan%
}{%
{\protect \APACyear {2016}}%
}]{%
wang2016globally}
\APACinsertmetastar {%
wang2016globally}%
\begin{APACrefauthors}%
Wang, H.%
\BCBT {}\ \BBA {} O’Sullivan, C.%
\end{APACrefauthors}%
\unskip\
\newblock
\APACrefYearMonthDay{2016}{}{}.
\newblock
{\BBOQ}\APACrefatitle {Globally continuous and non-Markovian activity analysis
  from videos} {Globally continuous and non-markovian activity analysis from
  videos}.{\BBCQ}
\newblock
 \APACrefbtitle {European conference on computer vision} {European conference
  on computer vision}\ (\BPGS\ 527--544).
\newblock
\begin{APACrefURL} {\url{https://doi.org/10.1007/978-3-319-46454-1_32}}
  \end{APACrefURL}
\PrintBackRefs{\CurrentBib}

\bibitem [\protect \citeauthoryear {%
X.~Wang%
, Ma%
, Ng%
\BCBL {}\ \BBA {} Grimson%
}{%
X.~Wang%
\ \protect \BOthers {.}}{%
{\protect \APACyear {2011}}%
}]{%
wang2011trajectory}
\APACinsertmetastar {%
wang2011trajectory}%
\begin{APACrefauthors}%
Wang, X.%
, Ma, K.T.%
, Ng, G\BHBI W.%
\BCBL {} Grimson, W.E.L.%
\end{APACrefauthors}%
\unskip\
\newblock
\APACrefYearMonthDay{2011}{}{}.
\newblock
{\BBOQ}\APACrefatitle {Trajectory analysis and semantic region modeling using
  nonparametric hierarchical bayesian models} {Trajectory analysis and semantic
  region modeling using nonparametric hierarchical bayesian models}.{\BBCQ}
\newblock
\APACjournalVolNumPages{International journal of computer
  vision}{95}{3}{287--312}.
\newblock
\begin{APACrefURL} {\url{https://doi.org/10.1109/CVPR.2008.4587718}}
  \end{APACrefURL}
\newblock

\newblock

\PrintBackRefs{\CurrentBib}

\bibitem [\protect \citeauthoryear {%
X.~Wang%
, Ma%
\BCBL {}\ \BBA {} Grimson%
}{%
X.~Wang%
\ \protect \BOthers {.}}{%
{\protect \APACyear {2008}}%
}]{%
wang2008unsupervised}
\APACinsertmetastar {%
wang2008unsupervised}%
\begin{APACrefauthors}%
Wang, X.%
, Ma, X.%
\BCBL {} Grimson, W.E.L.%
\end{APACrefauthors}%
\unskip\
\newblock
\APACrefYearMonthDay{2008}{}{}.
\newblock
{\BBOQ}\APACrefatitle {Unsupervised activity perception in crowded and
  complicated scenes using hierarchical bayesian models} {Unsupervised activity
  perception in crowded and complicated scenes using hierarchical bayesian
  models}.{\BBCQ}
\newblock
\APACjournalVolNumPages{IEEE Transactions on pattern analysis and machine
  intelligence}{31}{3}{539--555}.
\newblock
\begin{APACrefURL} {\url{https://doi.org/10.1109/TPAMI.2008.87}}
  \end{APACrefURL}
\newblock

\newblock

\PrintBackRefs{\CurrentBib}

\bibitem [\protect \citeauthoryear {%
X.~Wang%
, Tieu%
\BCBL {}\ \BBA {} Grimson%
}{%
X.~Wang%
\ \protect \BOthers {.}}{%
{\protect \APACyear {2006}}%
}]{%
wang2006learning}
\APACinsertmetastar {%
wang2006learning}%
\begin{APACrefauthors}%
Wang, X.%
, Tieu, K.%
\BCBL {} Grimson, E.%
\end{APACrefauthors}%
\unskip\
\newblock
\APACrefYearMonthDay{2006}{}{}.
\newblock
{\BBOQ}\APACrefatitle {Learning semantic scene models by trajectory analysis}
  {Learning semantic scene models by trajectory analysis}.{\BBCQ}
\newblock
 \APACrefbtitle {European conference on computer vision} {European conference
  on computer vision}\ (\BPGS\ 110--123).
\newblock
\begin{APACrefURL} {\url{https://doi.org/10.1007/11744078_9}} \end{APACrefURL}
\PrintBackRefs{\CurrentBib}

\bibitem [\protect \citeauthoryear {%
Xu%
, Piao%
\BCBL {}\ \BBA {} Gao%
}{%
Xu%
\ \protect \BOthers {.}}{%
{\protect \APACyear {2018}}%
}]{%
xu2018encoding}
\APACinsertmetastar {%
xu2018encoding}%
\begin{APACrefauthors}%
Xu, Y.%
, Piao, Z.%
\BCBL {} Gao, S.%
\end{APACrefauthors}%
\unskip\
\newblock
\APACrefYearMonthDay{2018}{}{}.
\newblock
{\BBOQ}\APACrefatitle {Encoding crowd interaction with deep neural network for
  pedestrian trajectory prediction} {Encoding crowd interaction with deep
  neural network for pedestrian trajectory prediction}.{\BBCQ}
\newblock
 \APACrefbtitle {Proceedings of the IEEE Conference on Computer Vision and
  Pattern Recognition} {Proceedings of the ieee conference on computer vision
  and pattern recognition}\ (\BPGS\ 5275--5284).
\newblock
\begin{APACrefURL} {\url{https://doi.org/10.1109/CVPR.2018.00553}}
  \end{APACrefURL}
\PrintBackRefs{\CurrentBib}

\bibitem [\protect \citeauthoryear {%
Zhou%
, Tang%
\BCBL {}\ \BBA {} Wang%
}{%
Zhou%
\ \protect \BOthers {.}}{%
{\protect \APACyear {2015}}%
}]{%
zhou2015learning}
\APACinsertmetastar {%
zhou2015learning}%
\begin{APACrefauthors}%
Zhou, B.%
, Tang, X.%
\BCBL {} Wang, X.%
\end{APACrefauthors}%
\unskip\
\newblock
\APACrefYearMonthDay{2015}{}{}.
\newblock
{\BBOQ}\APACrefatitle {Learning collective crowd behaviors with dynamic
  pedestrian-agents} {Learning collective crowd behaviors with dynamic
  pedestrian-agents}.{\BBCQ}
\newblock
\APACjournalVolNumPages{International Journal of Computer
  Vision}{111}{1}{50--68}.
\newblock
\begin{APACrefURL} {\url{https://doi.org/10.1007/s11263-014-0735-3}}
  \end{APACrefURL}
\newblock

\newblock

\PrintBackRefs{\CurrentBib}

\end{thebibliography}

\clearpage

\end{document}